# Benchmarking decision models against generative large language models on medical multiple-choice questions

Oscar Freyer[1], Daniel Glauert[2], Moritz Hofmann[3], Max Ostermann[1], Stephen Gilbert[1,4,*], Jakob Nikolas Kather[1,5,*]

[1] Else Kröner Fresenius Center for Digital Health, Faculty of Medicine and University Hospital Carl Gustav Carus, TUD Dresden University of Technology, Dresden, Germany

[2] Independent researcher, Berlin, Germany

[3] Denkmal.ai, Berlin, Germany

[4] Faculty of Business and Economics, TUD Dresden University of Technology, Dresden, Germany

[5] Institute for Medical AI (IMAI), Heidelberg University Hospital and Medical Faculty of Heidelberg University, Heidelberg, Germany

[*] Stephen Gilbert and Jakob Nikolas Kather contributed equally as senior authors.

**Author contact details**

Oscar Freyer: oscar.freyer@tu-dresden.de

Daniel Glauert: danielglauert@gmail.com

Moritz Hofmann: moritz@denkmal.ai

Max Ostermann: max.ostermann@tu-dresden.de

Stephen Gilbert: stephen.gilbert@tu-dresden.de

Jakob Nikolas Kather: contact@kather.ai

**Corresponding author:**

Dr. med. Oscar Freyer, TUD Dresden University of Technology, Else Kröner Fresenius Center for Digital Health, Faculty of Medicine Carl Gustav Carus, Fetscherstr. 74, 01307 Dresden, Germany, oscar.freyer@tu-dresden.de

# Abstract

Medical workflows often require selecting among predefined options. We compared 13 decision models with 15 generative large language model configurations, including different reasoning settings, using 1,273 four-option MedQA questions under text-only conditions. Accuracy ranged from 67.71% to 89.08% for decision models and from 78.32% to 96.86% for generative configurations. Among the tested configurations, Mercury Decide had the lowest mean final-attempt cost (USD 0.0069 per 1,000 questions), while Clef Omni had the shortest median request duration (0.291 seconds). Decision-model probabilities distinguished correct from incorrect answers, but error rates among answers assigned probabilities of at least 0.90 ranged from 1.71% to 12.77%. Answer agreement across three runs ranged from 92.30% to 100%, while probability repeatability varied. Decision models offered resource advantages alongside lower peak accuracy; selecting models for medical workflows requires considering probability quality and repeatability together with accuracy and the resources needed to detect and correct errors.

# Introduction

Generative foundation models, including large language models (LLMs), are being investigated for a range of medical applications, including summarizing radiology reports and clinical notes[1], answering medical questions[2], and supporting physicians in diagnostic reasoning[3]. Autoregressive LLMs generate text by predicting successive tokens from the preceding context, allowing a pre-trained model to respond to instructions and examples across different tasks without updating its parameters for each task[4]. Improvements in medical question answering, including those achieved by reasoning-optimized models[5], have supported the development of systems in which LLMs also select tools, request information and act on intermediate findings, also referred to as agentic systems[6, 7].

As AI tools are increasingly used in clinical practice[8], errors in their outputs can have consequences for patient care. In clinical documentation, for example, an LLM may add statements that are not supported by the patient record or omit relevant information, which can affect subsequent diagnosis or management[9]. These risks also depend on how clinicians use the output. Misleading reasoning and clinician overreliance on the model can influence clinical decisions, while the disclosure of sensitive patient information introduces further safety and security concerns[10].

One approach to reducing these risks is to provide the model with information from curated clinical sources through retrieval-augmented generation[11]. Generated content can also be checked against the patient record and reviewed by a clinician before it is used in care[9, 10]. As LLM-based systems gain greater autonomy[12, 13], these checks need to extend to the actions that a system can take.

In current research projects, LLMs are themselves being used to perform these checks. An LLM-as-a-judge evaluates another model's output against specified criteria and can support clinical safety assessment[14, 15]. This approach can be used to identify clinical hazards, harmful conversations, and to assess the overall quality of the output of generative AI systems[15–17]. Repeated assessments of this kind could support the evaluation and monitoring of clinical AI systems. However, the model performing the assessment can also make errors, while each additional check takes time and adds to the cost of the application. In agentic systems, several checks may be needed for each response or proposed action. The accuracy and consistency of the evaluator, as well as its cost and response time, therefore affect the operation of the wider system.

Decision-native models offer a new approach to these bounded tasks. One provider, TypeSafe, introduced their decision model JEV as a model built for structured probabilistic decisions, describing a new architecture, parallel output computation and a training method termed Reinforcement Learning for Calibrated Decisions[18]. Its interface receives the information to be assessed and typed questions, and returns a choice and option probabilities for each question[19]. The returned choice could be used, for example, to select a permitted action or to judge whether a response meets a specified evaluation criterion. For JEV, the developer claims low cost, short response times and freedom from hallucinations[18]. Here, freedom from hallucinations refers to guaranteed matching of the output to the specified schema.

The medical evaluation of decision-native models is still at an early stage, with no peer-reviewed studies available to our knowledge. An early evaluation compared JEV 1.13 with GPT-6 Sol[20]. However, this comparison included only one decision model and one generative model family and did not assess repeatability.

We therefore compared 13 decision models with 15 generative LLM configurations (i.e., models with a specific reasoning setting) on all 1,273 English US four-option MedQA[21] test questions under closed-book, text-only conditions. Our primary objective was to compare the accuracy of decision models with one another and with generative LLM configurations. We also assessed observed cost and request duration and examined the effects of native reasoning effort. For decision models, we examined how the returned probabilities related to answer correctness and assessed the repeatability of answers and probabilities across three runs per model.

# Methods

## Study design and dataset

We conducted a closed-book, text-only benchmark of 13 decision models and 15 generative LLM configurations on all 1,273 official English US four-option MedQA test questions. Run 1 supplied primary accuracy, resource and confidence results. Two further runs per decision model supplied repeatability results. The study comprised 35,644 primary and 33,098 repeat question/configuration/run outcomes, totaling 68,742 (**Fig. 1a**).

MedQA provided a common medical knowledge and reasoning task with four predefined options and reference answers, allowing both model types to be evaluated on the same inputs and scoring rule. It was selected for this controlled comparison rather than as a comprehensive measure of clinical reasoning. MedQA contained 10,178 training, 1,272 development and 1,273 test items. Development items were used to develop the benchmark pipeline; the test set supplied the evaluation questions[21]. The full dataset was downloaded from the MedQA repository at https://github.com/jind11/MedQA. A Python preparation pipeline preserved question wording, option labels and text, option order and source-row order. Stable item identifiers linked each question to its source hash and row number. Validation checked four options labelled A-D and agreement between each reference label and the supplied correct-answer text. Reference keys were stored separately from the inference inputs, which OF checked for solution leakage. Primary scoring retained all test questions and the original reference keys.

## Models and inference

The roster was a selected comparison of 13 native decision models and 15 generative LLM configurations across nine generative LLM families (**Extended Data Table 2**). A configuration comprised a model and its inference settings, including native reasoning effort where applicable. Three matched native effort pairs were evaluated: GPT-6 Luna Low/High, DeepSeek V4.1 Flash Low/High and Qwen3.8-27B Low/Xhigh. Decision-model eligibility required a native selected-label interface with support for four label-indexed option probabilities. All configurations were accessed through OpenRouter, an API gateway used to access the selected model providers, with a fixed serving provider and disabled fallbacks. Within each effort pair, the model, provider, prompt, parser and requested output allowance were identical apart from effort. These pairs allowed native effort settings to be compared while holding the other requested controls constant. Selection was restricted to hosted routes supporting the required response contract and full inputs; local inference, free variants and automatic model selection were excluded. Development checks verified route identities and supported controls before collection.

Every request contained the same question stem and four options in their original order, a bounded-choice instruction adapted to the API schema and empty conversation history. Reference keys, explanations, images, retrieval, browsing, tools, and native search were excluded. All models therefore received text and answer options without images; image-processing capability was not evaluated. Generative models were instructed to return a final JSON object containing only an 'answer' field with one label A-D; only the final response channel was parsed. Decision models were requested through their native Decisions or System One interface. Their selected label, four-option values and separate provider scalars were retained without replacing the selected label with the largest vector value. Generative requests specified max_completion_tokens=16384. Kimi K2.5 Instant requested reasoning.enabled=false, while its Native configuration retained native reasoning. Generative effort and token controls were not imposed on decision models. Development inputs were used to verify route identities, supported controls, full-input handling and response fields. Evaluation inputs were neither shortened nor truncated to fit a route. Requested and returned model identities, provider names, usage, charges, timestamps, finish states and raw responses were retained.

Collection comprised the first run (G01-G15 and D01), a Luna extension (D02) and a decision model extension (D03-D13). The 11 additional decision models were selected on 11 October 2026 after the initial runs. Original and Luna initial calls were recorded on 6-7 October 2026 (UTC), with failure recovery on 7-8 October. All campaigns used a 600-second timeout and scheduling seed 20261006. Original collection used up to 20 shared workers. InferenceNet segments used up to 20 active requests with two-second spacing, one active request with ten-second spacing, or ten active requests with five-second spacing. Mistral used two-second start spacing. Luna initial collection used one worker and one-second start spacing. Original/Luna failure recovery used ten shared workers, five-second spacing for OpenAI and InferenceNet, and two-second spacing for Mistral. The Decision model extension ran on 11-12 October 2026 with ten shared workers, a ten-call ceiling per provider and five-second minimum start spacing shared across each provider's models. Actual service dates, provider routes, concurrency and pacing were retained for each segment and selected final attempt. Transport retries and provider fallbacks were disabled.

Each question/configuration/run received one initial attempt. Failed or invalid responses permitted up to ten additional persisted attempts with unchanged inputs and parsing, stopping at the first parser-valid response independently of correctness. Valid wrong answers were final. Missing billing or probability fields did not trigger another call after a valid selected answer. The selected outcome was the valid initial response, first valid retry or exhausted failure. Primary initial attempts reached terminal outcomes before repeat initial attempts began within each campaign; primary failure recovery preceded repeat failure recovery. All attempts retained their question, configuration, campaign, run and retry identities. Unstarted or indeterminate outcomes remained missing and precluded complete-study inference.

## Statistics and reproducibility

The study was unregistered. The methods for accuracy estimation, confidence intervals and paired testing were documented before the original collection. The expanded decision-model roster, primary comparison family, additional reporting, visual sensitivity analysis and resource-imputation rules were specified after access to outcomes from the original and Luna extension campaigns.

Analyses were performed in Python 3.14.7 using NumPy 2.5.3 and SciPy 1.18.1. Figures were produced with Matplotlib 3.10.8. The sample comprised the full official test split, with all 1,273 questions evaluated by every configuration. Reference keys were kept separate from inference inputs and used for offline scoring.

***Accuracy and paired inference***

Top-1 accuracy was the number of correct selected answers divided by 1,273. Valid incorrect answers and exhausted refusals, truncations, invalid outputs and HTTP/transport failures counted as incorrect. Accuracy estimates were reported with Wilson 95% confidence intervals[22].

Paired effects were calculated as the first configuration's accuracy minus the second's, in percentage points. Two-sided exact McNemar tests[23] compared the numbers of questions answered correctly by only one configuration; P was set to 1 when there were no discordant answers. The primary analysis comprised 273 pairwise comparisons: each of the 13 decision models versus each of the 15 generative configurations (195 comparisons), plus every unique pair of decision models (78 comparisons). For decision-model pairs, accuracy differences subtracted the later model from the earlier model in the fixed D01-D13 order. We applied Holm adjustments to the 273 primary P values to control the family-wise error rate at 0.05[24].

Three exploratory comparisons assessed native reasoning effort: Luna High-Low, DeepSeek High-Low and Qwen 27B Xhigh-Low. Their P values were adjusted together using a separate Holm procedure. For each primary or effort comparison, an adjusted $P \leq 0.05$ was interpreted as evidence against equal accuracy. Comparisons were specified independently of observed accuracy rankings or significance.

For each accuracy difference, we calculated a paired 95% confidence interval by resampling questions with replacement 10,000 times. Each primary-analysis resample contained 1,273 question draws, with all configurations' responses to a sampled question kept together to preserve pairing. The 2.5th and 97.5th percentiles of the resulting accuracy differences defined the interval, using linear interpolation. Resampling used NumPy's PCG64 generator with seed 20261008. In the complete-valid sensitivity analysis, each draw contributed only questions with valid responses from both compared configurations. The visual sensitivity analysis resampled the common retained set of 1,165 questions. Supplementary Newcombe method 10 paired score intervals used the same contrast direction and denominator. Intervals were calculated separately for each comparison without adjustment for multiple comparisons. Question-level inference assumed approximately independent items.

***Resource analysis***

Cost and latency used the same selected final attempt for each primary question outcome: the valid initial response, first valid retry or last terminal attempt for an exhausted failure. Earlier unsuccessful-attempt charges and durations were excluded from primary resource summaries and frontiers. Mean cost in USD per question divided the total analysis cost by 1,273.

Request duration was the elapsed time from starting the selected final API call until it returned a response or failed. Each question contributed one duration, including exhausted failures. We reported the sum, median, interquartile range (IQR) and 95th percentile of these durations. Queueing, backoff, provider pauses and collection windows were recorded separately from request duration. Resource summaries retained the selected attempt's campaign, segment, provider and service conditions, including failure-recovery provenance when applicable.

Each missing final-attempt charge or duration in the primary run was imputed with the unrounded arithmetic mean of observed values for that resource among selected final attempts in the same configuration and run. Donors included observed zeros and exhausted final failures, and excluded earlier attempts, other configurations, repeat runs and imputed values.

We identified descriptive Pareto frontiers across all 28 primary configurations. Coordinates were run-1 accuracy, and mean final-attempt cost in USD per question and median final-attempt request duration in seconds respectively. Primary frontiers were identified using the observed-plus-imputed resource basis. In each accuracy-resource comparison, a configuration was dominated if another had at least equal accuracy and no greater resource use, with a strict improvement in at least one measure. Configurations with identical coordinates remained jointly nondominated. For the main probability figures, we selected decision models from the joint frontier across all 28 configurations: a model was dominated if another had at least equal accuracy, no greater mean cost and no greater median duration, with a strict improvement in at least one measure.

***Confidence and repeatability***

We evaluated each decision model separately using its four returned option probabilities. Provider documentation and development responses were used to verify the probability fields, their mapping to labels A-D and their returned numerical precision.

Confidence analyses used the native probability assigned to the model's selected option in run 1. We checked the complete vector to ensure that the selected-option score came from a usable four-option distribution. Eligible vectors contained the four supplied labels exactly once, finite probabilities between 0 and 1, and a total between 0.98 and 1.02 inclusive, allowing a small deviation from 1 to account for rounding. Native selected-option probabilities were used without rescaling. For comparisons of whole distributions across runs, each vector was divided by its sum so that its four probabilities summed to 1.

We assessed whether higher selected-option probabilities distinguished correct from incorrect answers using the area under the receiver operating characteristic curve (AUROC). For each model, the 2.5th and 97.5th percentiles of 10,000 bootstrap AUROC estimates defined the 95% interval, using seed 20261008. Each resample drew correct and incorrect answers separately, preserving their observed counts. The binary Brier score measured the mean squared difference between the selected-option probability and correctness, coded as 1 for a correct answer and 0 otherwise. Lower Brier scores indicated smaller probability errors. Calibration was assessed in ten fixed probability bins of width 0.1 by comparing each bin's mean selected-option probability with its observed proportion of correct answers and Wilson 95% interval. Bins included their lower boundary; only the final bin included its upper boundary of 1. We also reported retained answers, errors and error rates at probability ≥0.90.

Selective prediction evaluated the error rate when only answers meeting a probability cutoff were retained. At each cutoff, coverage was the number retained divided by all eligible answers for that model, and error rate was the number incorrect divided by the number retained. Answers with identical probabilities were retained or excluded as a group at each cutoff. Eligible denominators were reported separately for each model. Analyses used the returned probabilities without recalibration or selection of a cutoff from test-set performance.

Answer repeatability was assessed separately for each decision model using its three run accuracies and agreement between selected answer labels. For each run pair (1-2, 1-3 and 2-3), agreement was the proportion of questions with the same selected label among questions answered validly in both runs. Three-run agreement required the same label in all three runs and used only questions with valid answers in every run as its denominator.

Probability repeatability was assessed for each run pair in three ways. First, we compared the probabilities assigned to each run's own selected answer. Second, we measured the absolute change in probability assigned to the earlier run's selected option, keeping that option fixed even if the later run selected another answer. Third, we measured change across all four normalized probabilities using total variation, half the sum of their absolute differences. We report eligible counts and the mean, median and 95th percentile of each change measure, with exact unchanged counts where available.

Generative configurations had no repeat runs. The same repeatability measures were applied separately to all 13 decision models. Correctness/failure transition tables retained all 1,273 questions, including failed answers; probability comparisons used only questions with eligible vectors in the compared runs.

***Sensitivity analyses***

Additional sensitivity analyses assessed how retry recovery, exclusion of unusable responses and exclusion of questions flagged for potentially missing visual material affected accuracy estimates and paired comparisons.

The original first-attempt sensitivity used all 1,273 questions and scored original-call failures as incorrect, regardless of subsequent retry recovery. The complete-valid paired sensitivity analysis retained only questions with valid selected responses from both configurations. A post hoc visual sensitivity analysis excluded 108 questions flagged after outcome review for potentially missing visual material, leaving 1,165 questions for every configuration. The nomination set included explicit visual references and one examination reference; image necessity was not reviewed.

The visual exclusion was applied identically to each configuration and each separate repeat run, preserving original keys and first-valid scoring. All sensitivity comparisons are descriptive.

## AI-assisted code development

The authors also used GPT-6 Astra (OpenAI, Inc.) in order to support the writing of code, including the benchmark pipeline and the analysis notebook. After using these tools and services, the authors reviewed, tested, and edited the code to assure correctness.

## Ethics

The study used existing examination questions and involved no prospective patient intervention or the collection of primary patient data. Therefore, ethics approval was not applicable.

# Results

## Evaluation coverage and primary accuracy

All 68,742 question/configuration/run outcomes were resolved across 54 runs, with 69,318 preserved calls including 576 retries. Of 240 originally failed primary outputs, 225 yielded valid responses after retries. Fifteen primary queries remained without valid answers: 11 GLM-5 truncations, one Qwen3.8-27B Xhigh truncation, one Sonnet refusal and two Luna Decisions refusals returned as HTTP 502 responses. All were scored as incorrect in the primary analysis. The evaluation calls were recorded from 6 to 12 October 2026. The original and Luna initial collection took place on 6-7 October, failure recovery on 7-8 October, and the additional decision-model collection on 11-12 October.

Primary accuracy ranged from 67.71% for Kev 4B to 96.86% for GPT-6.1 Sol Low (**Table 1; Figure 1b**). Decision-model accuracy ranged from 67.71% to 89.08%, with Luna Decisions highest at 1,134/1,273 correct (89.08%; Wilson 95% CI, 87.25%-90.68%). Liquid d1 answered 1,125 questions correctly (88.37%; 86.50%-90.02%) and JEV answered 1,124 correctly (88.30%; 86.41%-89.95%). Of the 195 decision-versus-generative comparisons, generative configurations had higher accuracy in 168 and decision models in 14 at Holm-adjusted $P \leq 0.05$; the remaining 13 did not meet this threshold. Among the 78 decision-model pairs, 61 differed at the adjusted threshold and 17 did not. Overall, 243 of 273 primary comparisons met the threshold (Supplementary Table S1).

JEV's accuracy was 0.79 percentage points lower than Luna Decisions' (JEV-Luna difference, -0.79; bootstrap 95% CI, -2.51 to 0.94; adjusted $P=1.000$), with 60 questions correct only for JEV and 70 correct only for Luna Decisions. **Supplementary Table S1** reports all 273 primary effects and P values; **Supplementary Table S2** reports paired contingency counts and Newcombe intervals.

## Resource use and native effort

Mercury Decide had the lowest mean final-attempt charge at USD 0.0069 per 1,000 questions, followed by Decider V1.1 27B at USD 0.0071. JEV cost USD 0.0261 and Luna Decisions USD 0.0365 per 1,000 questions. Llama Maverick was the least expensive generative configuration at USD 0.0524 per 1,000 questions. Clef Omni had the shortest median final-attempt duration at 0.291 seconds, followed by JEV at 0.330 seconds and Luna Decisions at 0.335 seconds. The shortest generative median was 0.416 seconds for Mistral Medium (**Table 2**).

The accuracy-cost frontier contained Mercury Decide, Decider V1.1 27B, Liquid d1, Luna Decisions, Luna Low/High, DeepSeek Low and GPT-6.1 Sol Low (**Figure 1c**). Liquid d1 had both higher observed accuracy and lower cost than JEV. The accuracy-duration frontier contained Clef Omni, JEV, Luna Decisions, Sonnet Low and GPT-6.1 Sol Low (**Figure 1d**). The joint accuracy-cost-duration frontier is shown in **Supplementary Interactive Fig. 1**.

Higher native effort changed accuracy by 1.02 percentage points for Luna, -0.94 for DeepSeek and 1.34 for Qwen3.8-27B. None met the separate three-comparison Holm threshold (adjusted $P=.068$ for each; **Supplementary Table S3**). Higher effort increased mean final-attempt cost in all three pairs; median final-attempt duration increased for Luna and DeepSeek and decreased for Qwen.

## Confidence and selective prediction

Every valid run-1 decision-model answer had an eligible probability vector: 1,271 for Luna Decisions and 1,273 for each other model. Correctness AUROC ranged from 0.766 for Clef Flash to 0.901 for Decider V1.1 27B, while binary Brier scores ranged from 0.071 for Luna Decisions to 0.178 for Solar Decide Flash (**Table 3; Extended Data Fig. 1**). At probability ≥0.90, error rates ranged from 1.71% for JEV (15/877) to 12.77% for Solar Decide Flash (96/752). JEV retained 68.89% of eligible answers and Luna Decisions retained 77.03%, with 26 errors among 979 retained answers (2.66%). **Fig. 2** shows selective error rates, and **Fig. 3** shows discrimination and calibration for the six decision models on the joint accuracy-cost-duration frontier among all 28 configurations. Corresponding displays for all 13 decision models are provided in **Extended Data Fig. 1; Extended Data Fig. 2; Extended Data Fig. 3; Extended Data Fig. 4**. Reliability bins and score support are reported in **Supplementary Table S4** and **Supplementary Table S5**.

Calibration differed across models within the highest probability bin (**Supplementary Table S4; Extended Data Fig. 2**). For JEV, the 877 answers assigned probabilities of at least 0.90 had a mean probability of 0.982 and an observed accuracy of 98.29% (95% CI, 97.20-98.96%). For Solar Decide, the corresponding 1,017 answers had a mean probability of 0.979 and an observed accuracy of 89.18% (95% CI, 87.13-90.95%). Thus, similar mean assigned probabilities accompanied different observed accuracies.

## Answer and probability repeatability

The three-run selected-label agreement ranged from 92.30% for Solar Decide to 100% for Luna Decisions, Clef, Clef Flash and Decider V1.1 27B (**Extended Data Table 1**). JEV accuracy was 88.30%, 87.98% and 88.14% across runs 1-3, with identical selected labels in 1,249/1,273 questions (98.11%). Luna Decisions had 89.08% accuracy in each run, identical labels in all 1,271 valid triples and the same two failed questions in every run. Correctness/failure transitions are reported in **Supplementary Table S6**.

The proportion of questions for which all four option probabilities were exactly identical across the three runs ranged from 0% for Solar Decide Flash to 100% for Luna Decisions, Clef, Clef Flash and Decider V1.1 27B, at the precision returned by each service (**Supplementary Table S7**). JEV's native selected-answer probability was unchanged for 574/1,273 questions (45.09%), while its full native vector was unchanged for 568/1,273 (44.62%). JEV's mean pairwise selected-answer probability change ranged from 0.0144 to 0.0147 and normalized-vector total variation from 0.0163 to 0.0169. All eligible pairs were unchanged on both measures for Luna Decisions, Clef, Clef Flash and Decider V1.1 27B. **Supplementary Table S8** reports the probability changes across all 13 models.

## Sensitivity analyses

First-valid scoring increased primary accuracy over the original first attempt for 15 configurations by 0.08-5.26 percentage points. The largest change was GLM-5, from 89.71% to 94.97%; JEV remained at 88.30%, while Luna Decisions increased from 88.92% to 89.08% (**Supplementary Table S9**). Excluding the 108 visual-nominated questions left 1,165 questions per configuration. JEV increased to 89.10%, Luna Decisions to 89.36%, and GPT-6.1 Sol Low to 96.91%. Visual filtering changed accuracy by -0.04 percentage points for Mistral Medium to 1.08 points for Clef Flash. The JEV-Luna Decisions difference was -0.26 points in the filtered set (95% CI, -2.06 to 1.55). Complete-valid comparisons retained 1,260-1,273 matched questions. JEV versus Luna Decisions retained 1,271 pairs, with a difference of -0.87 points (95% CI, -2.59 to 0.87) (**Supplementary Table S10**).

# Discussion

Accuracy ranged from 67.71% to 89.08% across the 13 decision models, with generative configurations performing better in most paired comparisons. Some decision models nevertheless outperformed individual generative configurations, and several combined low cost and short request duration with accuracy near the upper end of the decision-model range. Native probabilities provided information about correctness, although the strength of this relationship and the error rate among high-probability answers differed across models. Answers were generally repeatable over three runs, while the consistency of the probabilities depended strongly on the model.

The highest accuracy among decision models was 89.08% for GPT-6 Luna Decisions, compared with 96.86% for the leading generative configuration, GPT-6.1 Sol Low. Several decision models returned answers at lower cost and with shorter request duration. In the point-estimate comparisons, Liquid d1 had slightly higher observed accuracy and lower cost than JEV, while JEV remained on the accuracy-duration frontier. The least costly way to obtain a given level of accuracy may therefore not be the fastest.

The cost and time benefit demonstrated in this study could make decision models particularly interesting for healthcare scenarios where a high number of time sensitive decisions has to be made. This could include triage decisions, regular checking of proposed treatments against contraindications documented in patient records, or supervision of AI systems with a higher autonomy. In these settings, absolutely small savings per call could accumulate across many records or recommendations. Whether lower charges or shorter request durations would justify lower accuracy, however, depends on the consequences of errors and how they are identified and corrected.

One potential way of separating truly correct from incorrect decisions could be the probability value that decision models provide on top of their given answer. The native probabilities of all 13 decision models distinguished correct from incorrect answers to differing degrees and could therefore help prioritise answers for review. However, some answers assigned high probabilities were wrong. JEV returned a valid answer for every primary question, yet 15 of the 877 answers assigned probabilities of at least 0.90 were incorrect. Solar Decide Flash had the highest error rate at this threshold, with 96 errors among 752 retained answers (12.77%). Kev 4B, which had the lowest overall accuracy, had 6 errors among 199 retained answers (3.02%). Decision models can still produce confident yet wrong answers. Therefore, probabilities could help to determine which results require further review, but a high value does not rule out the possibility of an error. In a clinical scenario, accepting an uncertain recommendation (by setting the acceptance threshold low) could result in a recommendation being accepted without further review, whilst rejecting a certain recommendation (by setting the acceptance threshold very high) could lead to unnecessary work. Setting a threshold therefore requires an examination of the errors among the accepted results, their consequences for the intended task, and the consistency with which the model assigns probabilities to the same decision.

Our study also showed, that receiving the same answer on repeated requests does not necessarily mean receiving the same probability for some models, which could have an effect on implementation scenarios.

For example, in a hypothetical workflow with an acceptance threshold of 0.90, an unchanged answer assigned 0.89 in one run and 0.91 in another would be sent for review in one case and accepted in the other. JEV's mean probability changes were small, but their effect on review decisions would depend on how close individual values were to the chosen threshold.

Future studies should use new, previously unseen benchmarks and cases that more closely reflect clinical practice, including billing code assignment, diagnosis and triage. The use of decision models to judge other models' outputs should also be evaluated against hazard-specific criteria, such as those proposed in Meczner et al.'s preprint[16]. They should also be tested as safety layers that check outputs or proposed actions against defined clinical rules and regulatory requirements. Repeating generative-model runs under comparable conditions would allow their answer repeatability to be compared with that of decision models. Robustness to changes in question wording should also be evaluated.

This study has several limitations. First, the evaluation used public English examination questions. Prior exposure to MedQA during training could have affected accuracy and comparisons between models, but this study did not measure the extent of exposure or differences in exposure across models. Transferability to other languages is limited. The selected roster and collection period also limit representativeness beyond the tested configurations. Second, some questions contained visual references, although excluding the set of 108 nominated visual questions changed accuracy only modestly. Two pairs of questions shared clinical context, and this dependence was not modelled by the question-level bootstrap, Wilson intervals or McNemar tests. Third, it is currently unknown whether Luna's decision and generative configurations shared weights, computation or serving behavior. Fourth, response times were measured through different providers, on different dates and with different numbers of simultaneous requests. They therefore describe performance under those collection conditions, since a shorter duration could reflect provider infrastructure or load as well as the computation required by the model. Unreported provider updates between repeat runs collected on different dates could also contribute to changes in answers or probabilities. Three missing primary charges were mean-imputed without additional uncertainty. Generative configurations contributed one realised response per question, so their rerun variability was not estimated. Fifth, the four supplied answer choices limited the task to selecting among options that had already been identified, which does not depict clinical workflow realities. The questions also assume that the supplied information is sufficient to choose an answer, whereas patient records or examination results may leave a decision unresolved. This benchmark must therefore be understood as only establishing a baseline of decision model capabilities in medicine. Finally, API charges measured the prices paid by users rather than providers' computational or operating costs, so lower charges do not establish greater computational efficiency.

In this medical question-answering benchmark, decision models differed substantially in accuracy, and their resource advantages were generally accompanied by lower accuracy than the leading generative model configurations. Native probabilities helped distinguish correct from incorrect answers, but a common threshold did not produce a common residual error rate. Decision models still produced confident yet wrong answers.

Model selection therefore requires assessing accuracy, probability reliability and the repeatability of both answers and probabilities alongside cost and response time for the intended task. Whether lower resource use per decision benefits a clinical workflow depends on the exact task, the work required to detect and correct errors, and the consequences of failure.

## Data availability

The structured per-question results are stored on Zenodo under CC BY 4.0, including selected answers, correctness, available native probabilities, resource measurements and the run/attempt provenance needed to reproduce the analysis 10.5281/zenodo.23698804. Raw request/response logs remain private. The MedQA dataset is available at the public MedQA repository (https://github.com/jind11/MedQA). Instructions describe how to reconstruct the evaluated questions, options and reference keys from the upstream dataset.

## Code availability

The original benchmarking and analysis code is available under the MIT licence in the GitHub repository oscarfreyer/decision-model-benchmark, with a versioned archive on Zenodo 10.5281/zenodo.23698804. Third-party components retain their existing licences.

## Acknowledgments

During the preparation of this work, the authors used DeepL (DeepL SE), Grammarly (Grammarly, Inc.), and ChatGPT (GPT-6 Astra; OpenAI, Inc.) to improve the manuscript's grammar, spelling, and readability. After using these tools and services, the authors reviewed and edited the content as needed and take full responsibility for the content of the publication.

## Funding statement

This work was supported by the European Commission under the Horizon Europe Program, as part of the project CYMEDSEC (101094218). The views and opinions expressed are those of the authors only and do not necessarily reflect those of the European Union. Neither the European Union nor the granting authorities can be held responsible for them. Responsibility for the information and views expressed therein lies entirely with the authors. This work has been supported by the BRIDGE project, funded by the Innovative Health Initiative Joint Undertaking (IHI JU) under grant agreement No 101165912. The IHI JU receives support from the European Union's Horizon Europe research and innovation programme and from COCIR, EFPIA, EuropaBio, MedTech Europe, and Vaccines Europe. This publication reflects only the views of the authors and does not necessarily represent the views of the BRIDGE consortium. This work was supported by the German Federal Ministry of Education, Research and Space (BMFTR) as part of the Zukunftscluster SEMECO (03ZU2210BA). The views and opinions expressed are those of the authors only and do not necessarily reflect those of their employer. The funders had no role in study design, data collection, analysis, interpretation of the results, or writing the manuscript.

## Author contributions

OF: Conceptualization, Methodology, Software, Validation, Formal Analysis, Investigation, Data Curation, Writing - Original Draft, Writing - Review & Editing, Visualization, Project Administration. DG: Conceptualization, Methodology, Validation, Writing - Review & Editing. MH: Conceptualization,

Resources, Writing - Review & Editing. MO: Conceptualization, Methodology, Writing - Review & Editing. SG: Conceptualization, Writing - Review & Editing, Supervision, Funding Acquisition. JNK: Conceptualization, Methodology, Validation, Resources, Writing - Review & Editing, Visualization, Supervision, Funding Acquisition.

# Competing interests

OF has a leadership role and holds stock in WhalesDontFly GmbH, has had consulting relationships with Prova Health Ltd, and has a subcontractual consulting relationship with the Saudi FDA. S.G. has or has had consulting relationships with Ada Health GmbH, AMBOSS SE, European Commission Directorate General for Research & Innovation, Flo Ltd., FORUM Institut für Management GmbH, High-Tech Gründerfonds Management GmbH, Haleon Plc., ICURA ApS, Lindus Health Ltd., Prova Health Ltd., Rock Health Inc., Saudi Arabia Food and Drug Authority (FDA) funded through the United Nations Development Programme (UNDP), Thymia Ltd., Una Health GmbH and holds share options in Ada Health GmbH. JNK declares consulting services for Bioptimus, France; Panakeia, UK; AstraZeneca, UK; and MultiplexDx, Slovakia. Furthermore, he holds shares in StratifAI, Germany, Synagen, Germany, and Ignition Lab, Germany; has received an institutional research grant by GSK; and has received honoraria by AstraZeneca, Bayer, Daiichi Sankyo, Eisai, Janssen, Merck, MSD, BMS, Roche, Pfizer, and Fresenius. MO has a subcontractual consulting relationship with the Saudi FDA. MH has a leadership role and holds stock in WhalesDontFly GmbH. DG declares no competing interests.

# Tables

**Table 1. Primary accuracy and unanswered queries.** Accuracy was calculated for all 1,273 questions in run 1 for each configuration, using the first valid answer and counting exhausted failures as incorrect. Pointwise Wilson 95% confidence intervals describe the uncertainty in each estimate. After retries, 15 queries remained unanswered: 11 for GLM-5, one each for Qwen3.8-27B Xhigh and Sonnet Low, and two for Luna Decisions.

| Model (ID) | Correct | n | Accuracy (%) | Wilson 95% CI (%) | Unanswered after retries |
|---|---|---|---|---|---|
| GPT-6.1 Sol Low (G01) | 1,233 | 1,273 | 96.86 | 95.75 to 97.68 | 0 |
| Gemini 3.8 Flash Low (G05) | 1,232 | 1,273 | 96.78 | 95.66 to 97.62 | 0 |
| DeepSeek V4.1 Flash Low (G06) | 1,216 | 1,273 | 95.52 | 94.24 to 96.53 | 0 |
| Qwen3.8-27B Xhigh (G10) | 1,215 | 1,273 | 95.44 | 94.16 to 96.46 | 1 |
| GPT-6 Luna High (G03) | 1,212 | 1,273 | 95.21 | 93.89 to 96.25 | 0 |
| Claude Sonnet 5.5 Low (G04) | 1,211 | 1,273 | 95.13 | 93.81 to 96.18 | 1 |
| GLM-5 Native (G12) | 1,209 | 1,273 | 94.97 | 93.63 to 96.04 | 11 |
| DeepSeek V4.1 Flash High (G07) | 1,204 | 1,273 | 94.58 | 93.20 to 95.69 | 0 |
| Kimi K2.5 Native (G13) | 1,203 | 1,273 | 94.50 | 93.11 to 95.62 | 0 |
| GPT-6 Luna Low (G02) | 1,199 | 1,273 | 94.19 | 92.76 to 95.34 | 0 |
| Qwen3.8-27B Low (G09) | 1,198 | 1,273 | 94.11 | 92.68 to 95.27 | 0 |

| Model (ID) | Correct | n | Accuracy (%) | Wilson 95% CI (%) | Unanswered after retries |
|---|---|---|---|---|---|
| Qwen3.8 Max Low (G08) | 1,141 | 1,273 | 89.63 | 87.84 to 91.19 | 0 |
| GPT-6 Luna Decisions (D02) | 1,134 | 1,273 | 89.08 | 87.25 to 90.68 | 2 |
| Liquid d1 (D08) | 1,125 | 1,273 | 88.37 | 86.50 to 90.02 | 0 |
| JEV 1.13 (D01) | 1,124 | 1,273 | 88.30 | 86.41 to 89.95 | 0 |
| Decider V1.1 27B (D09) | 1,118 | 1,273 | 87.82 | 85.91 to 89.51 | 0 |
| Mercury Decide (D05) | 1,085 | 1,273 | 85.23 | 83.18 to 87.07 | 0 |
| Kimi K2.5 Instant (G15) | 1,082 | 1,273 | 85.00 | 82.93 to 86.85 | 0 |
| Clef (D06) | 1,079 | 1,273 | 84.76 | 82.68 to 86.63 | 0 |
| Llama 4 Maverick Standard (G14) | 1,067 | 1,273 | 83.82 | 81.69 to 85.74 | 0 |
| Solar Decide (D04) | 1,052 | 1,273 | 82.64 | 80.46 to 84.62 | 0 |
| Microsoft-Decision-1 (D13) | 1,013 | 1,273 | 79.58 | 77.27 to 81.70 | 0 |
| Clef Omni (D11) | 1,011 | 1,273 | 79.42 | 77.11 to 81.55 | 0 |
| Mistral Medium 3.5 None (G11) | 997 | 1,273 | 78.32 | 75.97 to 80.50 | 0 |
| Clef Flash (D07) | 938 | 1,273 | 73.68 | 71.20 to 76.03 | 0 |
| Drex v1.5 (D12) | 936 | 1,273 | 73.53 | 71.04 to 75.88 | 0 |

| Model (ID) | Correct | n | Accuracy (%) | Wilson 95% CI (%) | Unanswered after retries |
|---|---|---|---|---|---|
| Solar Decide Flash (D10) | 928 | 1,273 | 72.90 | 70.39 to 75.27 | 0 |
| Kev 4B (D03) | 862 | 1,273 | 67.71 | 65.10 to 70.23 | 0 |

**Table 2. Primary final-attempt cost and duration.** Cost and duration were calculated from the same selected final attempt for each of the 1,273 questions, including exhausted failures. Earlier unsuccessful attempts were included only in the retry-call count. Missing charges were imputed using the mean of the observed final charges for the respective model: 1,271 charges for Luna Decisions, which had two missing values, and 1,272 for Sonnet, which had one. The IQR is the difference between the 75th and 25th percentiles. Costs are expressed in USD per 1,000 questions.

| Model (ID) | USD/1,000 questions | Median duration, s | IQR, s | 95th percentile, s | Retry calls |
|---|---|---|---|---|---|
| GPT-6.1 Sol Low (G01) | 0.8554 | 2.506 | 1.281 | 5.346 | 0 |
| Gemini 3.8 Flash Low (G05) | 0.8837 | 3.188 | 1.107 | 5.868 | 1 |
| DeepSeek V4.1 Flash Low (G06) | 0.1894 | 3.037 | 4.905 | 33.771 | 34 |
| Qwen3.8-27B Xhigh (G10) | 3.7595 | 6.198 | 15.881 | 137.670 | 67 |
| GPT-6 Luna High (G03) | 0.1110 | 2.544 | 0.960 | 7.726 | 0 |
| Claude Sonnet 5.5 Low (G04) | 1.3997 | 1.632 | 0.539 | 3.925 | 21 |
| GLM-5 Native (G12) | 7.2455 | 9.458 | 71.857 | 231.853 | 319 |
| DeepSeek V4.1 Flash High (G07) | 0.3113 | 3.836 | 7.488 | 54.061 | 25 |
| Kimi K2.5 Native (G13) | 6.7063 | 44.290 | 87.319 | 508.562 | 2 |
| GPT-6 Luna Low (G02) | 0.0751 | 2.532 | 0.982 | 5.524 | 0 |
| Qwen3.8-27B Low (G09) | 1.6098 | 7.830 | 4.342 | 20.193 | 0 |
| Qwen3.8 Max Low (G08) | 3.8807 | 12.225 | 4.965 | 23.639 | 0 |
| GPT-6 Luna Decisions (D02) | 0.0365 | 0.335 | 0.102 | 0.801 | 22 |
| Liquid d1 (D08) | 0.0115 | 0.377 | 0.088 | 0.634 | 0 |
| JEV 1.13 (D01) | 0.0261 | 0.330 | 0.067 | 0.473 | 0 |

| Model (ID) | USD/1,000 questions | Median duration, s | IQR, s | 95th percentile, s | Retry calls |
|---|---|---|---|---|---|
| Decider V1.1 27B (D09) | 0.0071 | 0.392 | 0.080 | 0.650 | 0 |
| Mercury Decide (D05) | 0.0069 | 0.455 | 0.231 | 4.752 | 0 |
| Kimi K2.5 Instant (G15) | 0.1308 | 1.676 | 0.187 | 1.991 | 0 |
| Clef (D06) | 0.0996 | 0.477 | 0.266 | 1.275 | 2 |
| Llama 4 Maverick Standard (G14) | 0.0524 | 1.529 | 0.796 | 2.925 | 0 |
| Solar Decide (D04) | 0.0298 | 0.876 | 5.317 | 18.935 | 0 |
| Microsoft-Decision-1 (D13) | 0.0126 | 0.402 | 0.102 | 0.821 | 2 |
| Clef Omni (D11) | 0.0613 | 0.291 | 0.177 | 1.119 | 13 |
| Mistral Medium 3.5 None (G11) | 0.4360 | 0.416 | 0.113 | 0.843 | 2 |
| Clef Flash (D07) | 0.0158 | 0.520 | 0.318 | 1.432 | 2 |
| Drex v1.5 (D12) | 0.0114 | 0.637 | 0.124 | 1.163 | 1 |
| Solar Decide Flash (D10) | 0.0298 | 0.803 | 0.181 | 1.335 | 1 |
| Kev 4B (D03) | 0.0120 | 0.636 | 0.147 | 1.678 | 0 |

**Table 3. Run-1 confidence discrimination and high-confidence errors.** The analysis used the native probability assigned to the selected option for each eligible run-1 answer. AUROC describes how well these probabilities distinguished correct from incorrect answers, with pointwise 95% intervals calculated from 10,000 stratified bootstrap resamples. The binary Brier score is the mean squared probability error. High-confidence error rates were calculated among answers assigned a probability ≥0.90. The two failed Luna Decisions responses were excluded; each other decision model had 1,273 eligible responses.

| Model (ID) | Eligible n | AUROC | AUROC 95% CI | Brier | Errors at ≥0.90 | Answers at ≥0.90 | Error rate at ≥0.90 (%) |
|---|---|---|---|---|---|---|---|
| JEV 1.13 (D01) | 1,273 | 0.901 | 0.880 to 0.920 | 0.074 | 15 | 877 | 1.71 |
| GPT-6 Luna Decisions (D02) | 1,271 | 0.895 | 0.869 to 0.918 | 0.071 | 26 | 979 | 2.66 |

| Model (ID) | Eligible n | AUROC | AUROC 95% CI | Brier | Errors at ≥0.90 | Answers at ≥0.90 | Error rate at ≥0.90 (%) |
|---|---|---|---|---|---|---|---|
| Kev 4B (D03) | 1,273 | 0.784 | 0.759 to 0.808 | 0.173 | 6 | 199 | 3.02 |
| Solar Decide (D04) | 1,273 | 0.811 | 0.782 to 0.840 | 0.134 | 110 | 1,017 | 10.82 |
| Mercury Decide (D05) | 1,273 | 0.865 | 0.839 to 0.889 | 0.107 | 77 | 1,042 | 7.39 |
| Clef (D06) | 1,273 | 0.865 | 0.837 to 0.890 | 0.096 | 28 | 832 | 3.37 |
| Clef Flash (D07) | 1,273 | 0.766 | 0.734 to 0.795 | 0.163 | 47 | 486 | 9.67 |
| Liquid d1 (D08) | 1,273 | 0.892 | 0.868 to 0.914 | 0.075 | 16 | 867 | 1.85 |
| Decider V1.1 27B (D09) | 1,273 | 0.901 | 0.879 to 0.922 | 0.077 | 29 | 963 | 3.01 |
| Solar Decide Flash (D10) | 1,273 | 0.791 | 0.765 to 0.816 | 0.178 | 96 | 752 | 12.77 |
| Clef Omni (D11) | 1,273 | 0.806 | 0.776 to 0.835 | 0.129 | 27 | 504 | 5.36 |
| Drex v1.5 (D12) | 1,273 | 0.777 | 0.749 to 0.803 | 0.175 | 80 | 697 | 11.48 |
| Microsoft-Decision-1 (D13) | 1,273 | 0.825 | 0.799 to 0.850 | 0.130 | 36 | 698 | 5.16 |

# Figures

**Fig. 1. Study design, primary accuracy and resource use.** Panel a shows the study design. 13 decision models were evaluated in three runs and 15 generative configurations in one run each, using 1,273 MedQA questions. Panel b shows accuracy for all 28 configurations in run 1, calculated from the first valid answer, with exhausted failures counted as incorrect and pointwise Wilson 95% intervals. Panels c and d relate accuracy to resource use for the selected final attempts, including exhausted failures. Mean charge in USD per 1,000 questions in c and median duration in seconds in d. Both resource axes use logarithmic scales. Labels in panels c and d refer to the configuration IDs in panel b. Missing charges for two Luna Decisions answers and one Sonnet answer were imputed using the respective mean charges. Orange denotes decision models and teal denotes generative configurations. Diamonds identify the nondominated configurations in each two-dimensional comparison. For these configurations, no other tested configuration achieved at least the same accuracy with no greater resource use and a strict improvement in at least one measure. Dashed lines connect these points and indicate the pareto frontier. Supplementary Interactive Fig. 1 shows the 3D pareto frontier.

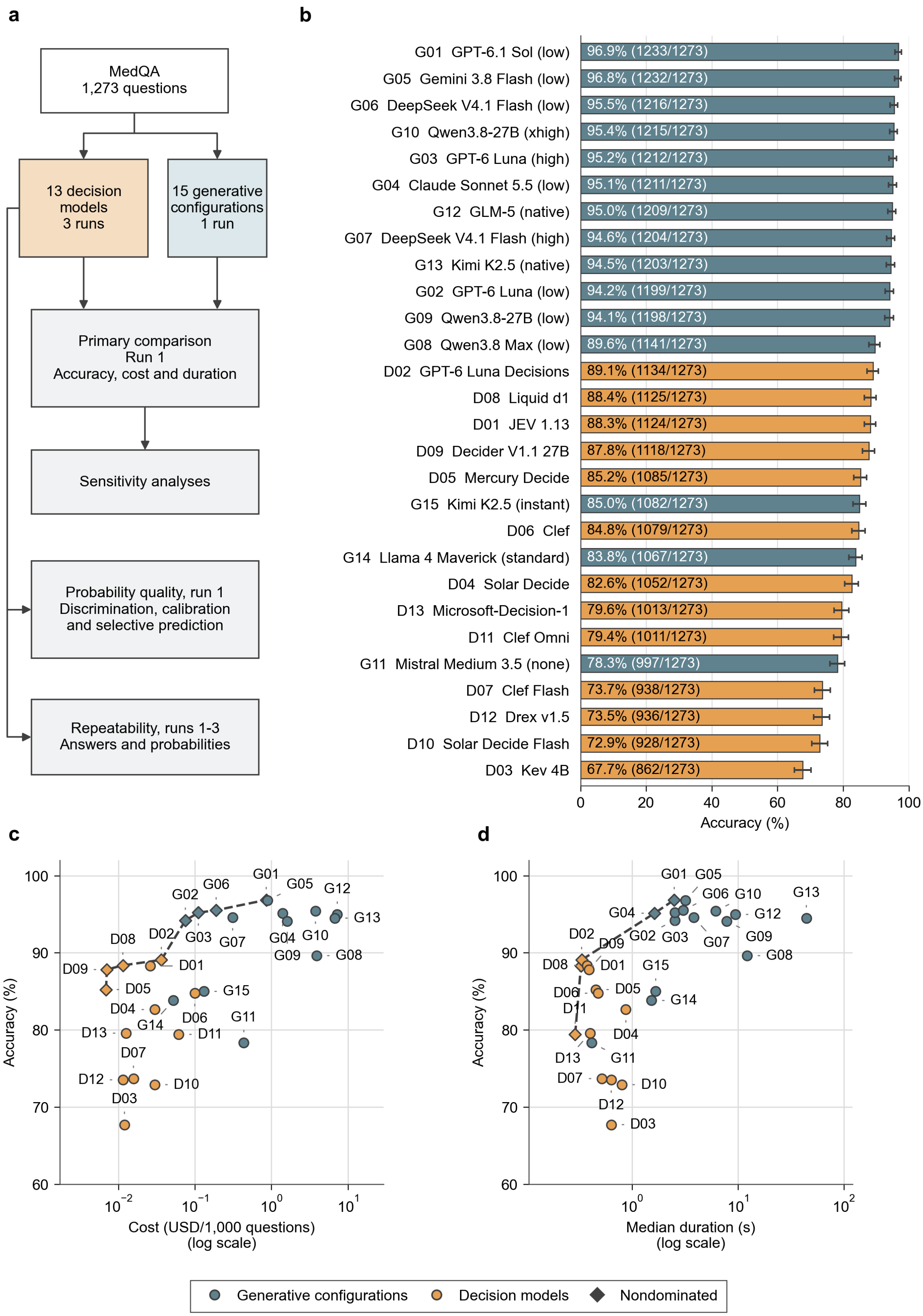
a
MedQA
1,273 questions
13 decision models
3 runs
15 generative configurations
1 run
Primary comparison
Run 1
Accuracy, cost and duration
Sensitivity analyses
Probability quality, run 1
Discrimination, calibration and selective prediction
Repeatability, runs 1-3
Answers and probabilities
b
G01 GPT-6.1 Sol (low) 96.9% (1233/1273)
G05 Gemini 3.8 Flash (low) 96.8% (1232/1273)
G06 DeepSeek V4.1 Flash (low) 95.5% (1216/1273)
G10 Qwen3.8-27B (xhigh) 95.4% (1215/1273)
G03 GPT-6 Luna (high) 95.2% (1212/1273)
G04 Claude Sonnet 5.5 (low) 95.1% (1211/1273)
G12 GLM-5 (native) 95.0% (1209/1273)
G07 DeepSeek V4.1 Flash (high) 94.6% (1204/1273)
G13 Kimi K2.5 (native) 94.5% (1203/1273)
G02 GPT-6 Luna (low) 94.2% (1199/1273)
G09 Qwen3.8-27B (low) 94.1% (1198/1273)
G08 Qwen3.8 Max (low) 89.6% (1141/1273)
D02 GPT-6 Luna Decisions 89.1% (1134/1273)
D08 Liquid d1 88.4% (1125/1273)
D01 JEV 1.13 88.3% (1124/1273)
D09 Decider V1.1 27B 87.8% (1118/1273)
D05 Mercury Decide 85.2% (1085/1273)
G15 Kimi K2.5 (instant) 85.0% (1082/1273)
D06 Clef 84.8% (1079/1273)
G14 Llama 4 Maverick (standard) 83.8% (1067/1273)
D04 Solar Decide 82.6% (1052/1273)
D13 Microsoft-Decision-1 79.6% (1013/1273)
D11 Clef Omni 79.4% (1011/1273)
G11 Mistral Medium 3.5 (none) 78.3% (997/1273)
D07 Clef Flash 73.7% (938/1273)
D12 Drex v1.5 73.5% (936/1273)
D10 Solar Decide Flash 72.9% (928/1273)
D03 Kev 4B 67.7% (862/1273)
0 20 40 60 80 100
Accuracy (%)
c
Accuracy (%)
Cost (USD/1,000 questions) (log scale)
d
Accuracy (%)
Median duration (s) (log scale)
Generative configurations
Decision models
Nondominated

**Fig. 2. Selective error rates for decision models on the joint accuracy-cost-duration frontier.** Within each block, the panels show JEV, GPT-6 Luna Decisions, Mercury Decide, Liquid d1, Decider V1.1 27B and Clef Omni in reading order. These six decision models were on the joint frontier for accuracy, mean selected-final-attempt cost and median duration among all 28 model configurations. The upper panels (a-f) relate error rate to the proportion of eligible answers retained, whereas the lower panels (g-l) relate it to the native selected-answer probability cutoff. Answers at or above each cutoff were retained, with answers assigned identical probabilities kept or excluded together. Error rates are shown from 0-25%, with values above 25% outside the displayed range; no error rate is defined when a cutoff retains no answers. The analysis included 1,271 eligible run-1 answers for Luna Decisions and 1,273 for each other displayed model. Extended Data Figs. 3-4 show all 13 decision models.

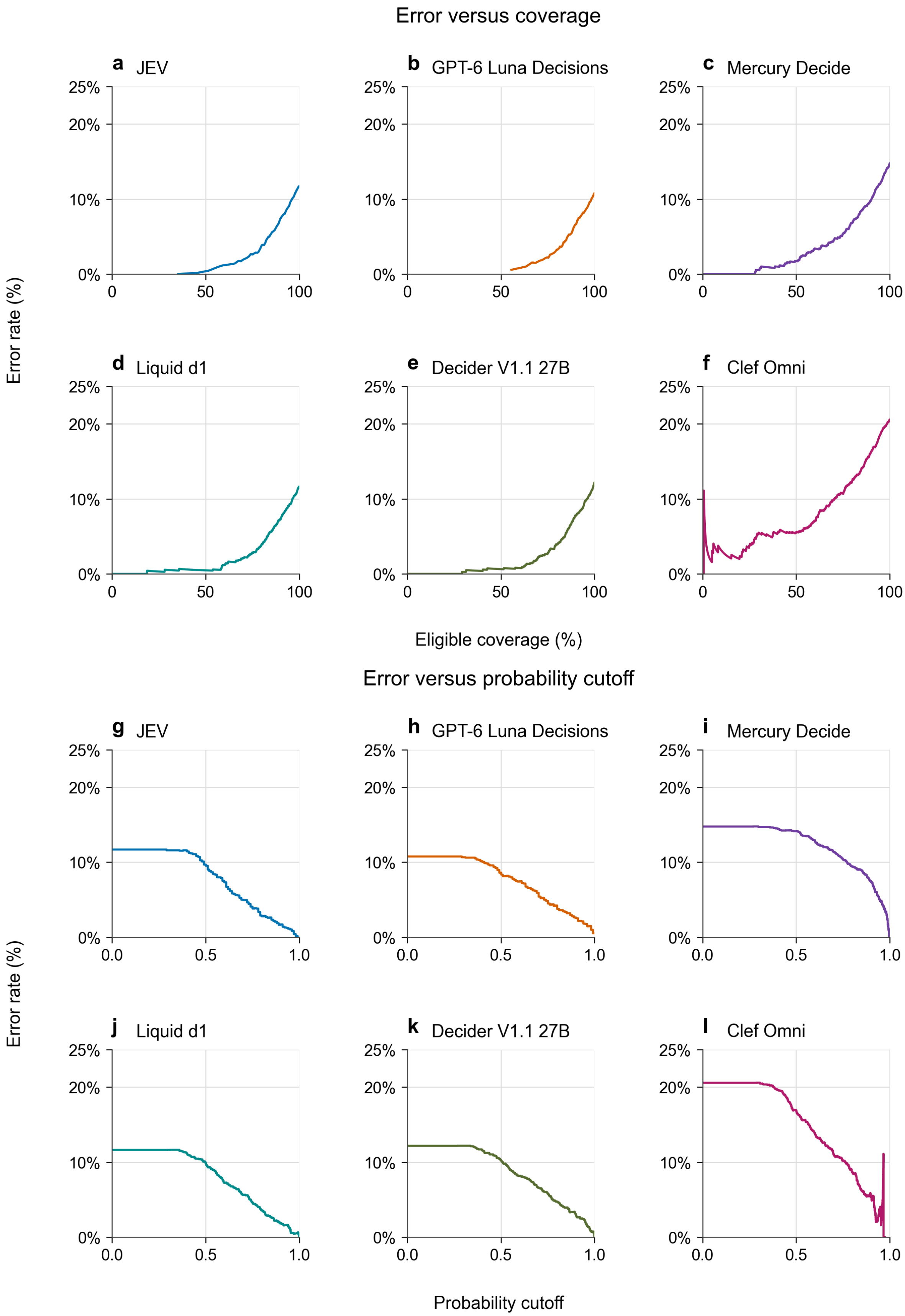
Error versus coverage
a JEV
b GPT-6 Luna Decisions
c Mercury Decide
d Liquid d1
e Decider V1.1 27B
f Clef Omni
Error rate (%)
25%
20%
10%
0%
0
50
100
Eligible coverage (%)
Error versus probability cutoff
g JEV
h GPT-6 Luna Decisions
i Mercury Decide
j Liquid d1
k Decider V1.1 27B
l Clef Omni
Error rate (%)
0.0
0.5
1.0
Probability cutoff

**Fig. 3. Probability discrimination and calibration for decision models on the joint accuracy-cost-duration frontier.** Within each block, the panels show JEV, GPT-6 Luna Decisions, Mercury Decide, Liquid d1, Decider V1.1 27B and Clef Omni in reading order, using the same joint frontier selection as Fig. 2. The upper panels (a-f) show ROC curves for distinguishing correct from incorrect answers, together with AUROC and pointwise 95% stratified-bootstrap intervals. The lower panels (g-l) compare the mean native selected-option probability with observed accuracy in ten fixed probability bins. Each unconnected point represents a nonempty bin, with a pointwise Wilson 95% interval for observed accuracy; empty bins have no plotted estimate. The dotted diagonals represent no discrimination in the ROC panels and perfect calibration in the calibration panels. The analysis included 1,271 eligible run-1 answers for Luna Decisions and 1,273 for each other displayed model. Extended Data Figs. 1-2 show all 13 decision models.

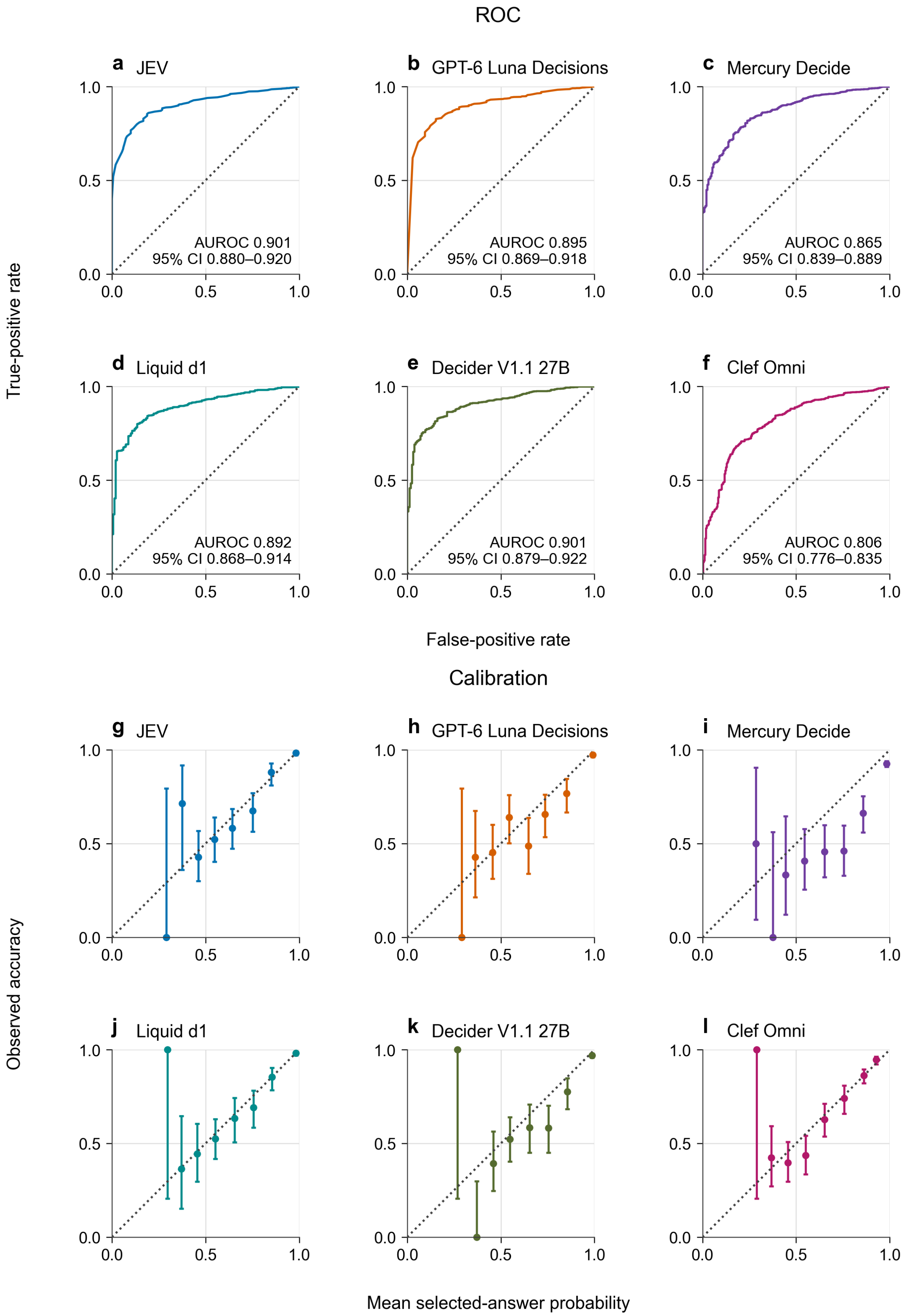
ROC
a JEV
AUROC 0.901
95% CI 0.880–0.920
b GPT-6 Luna Decisions
AUROC 0.895
95% CI 0.869–0.918
c Mercury Decide
AUROC 0.865
95% CI 0.839–0.889
d Liquid d1
AUROC 0.892
95% CI 0.868–0.914
e Decider V1.1 27B
AUROC 0.901
95% CI 0.879–0.922
f Clef Omni
AUROC 0.806
95% CI 0.776–0.835
True-positive rate
False-positive rate
Calibration
g JEV
h GPT-6 Luna Decisions
i Mercury Decide
j Liquid d1
k Decider V1.1 27B
l Clef Omni
Observed accuracy
Mean selected-answer probability

# Extended Data

**Extended Data Fig. 1. Correctness discrimination across all 13 decision models.** Panels a-m show how well the native probability assigned to the selected answer distinguished correct from incorrect answers in run 1. Each panel shows the ROC curve and reports AUROC with a pointwise 95% interval calculated from 10,000 stratified bootstrap resamples. The dotted diagonal represents no discrimination. The analysis included 1,271 eligible answers for Luna Decisions and 1,273 for each other model.

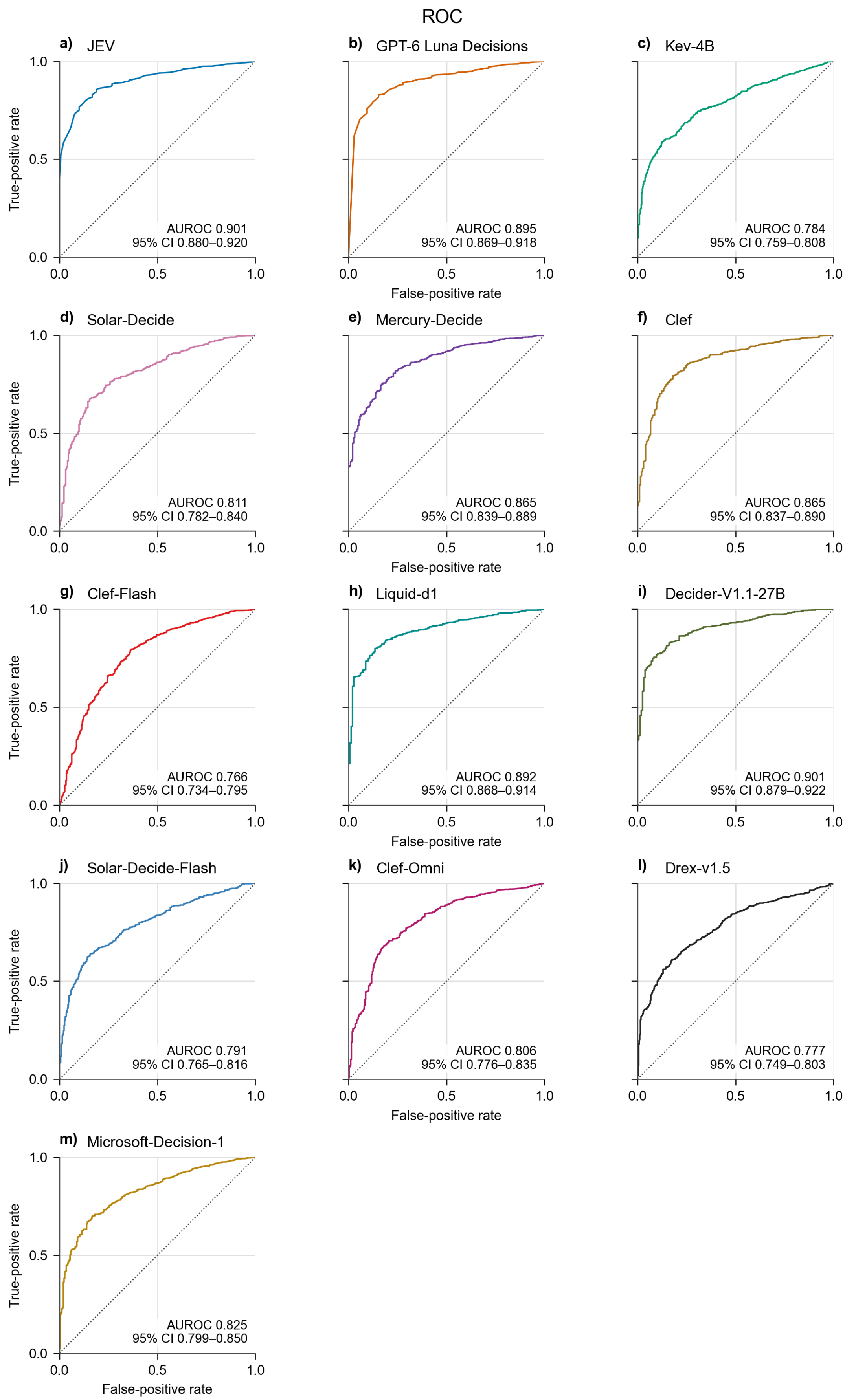
ROC
a) JEV
AUROC 0.901
95% CI 0.880–0.920
b) GPT-6 Luna Decisions
AUROC 0.895
95% CI 0.869–0.918
c) Kev-4B
AUROC 0.784
95% CI 0.759–0.808
d) Solar-Decide
AUROC 0.811
95% CI 0.782–0.840
e) Mercury-Decide
AUROC 0.865
95% CI 0.839–0.889
f) Clef
AUROC 0.865
95% CI 0.837–0.890
g) Clef-Flash
AUROC 0.766
95% CI 0.734–0.795
h) Liquid-d1
AUROC 0.892
95% CI 0.868–0.914
i) Decider-V1.1-27B
AUROC 0.901
95% CI 0.879–0.922
j) Solar-Decide-Flash
AUROC 0.791
95% CI 0.765–0.816
k) Clef-Omni
AUROC 0.806
95% CI 0.776–0.835
l) Drex-v1.5
AUROC 0.777
95% CI 0.749–0.803
m) Microsoft-Decision-1
AUROC 0.825
95% CI 0.799–0.850
True-positive rate
False-positive rate
0.0
0.5
1.0

**Extended Data Fig. 2. Calibration across all 13 decision models.** Panels a-m compare the mean native selected-option probability with observed accuracy in ten fixed bins for run 1. Each point represents a nonempty bin, with a pointwise Wilson 95% interval for observed accuracy; empty bins have no plotted estimate. Each bin included its lower boundary and excluded its upper boundary, except the final bin, which included 1. The dotted diagonal represents perfect calibration. The analysis included 1,271 eligible answers for Luna Decisions and 1,273 for each other model. Supplementary Table S4 provides the counts and estimates for each bin.

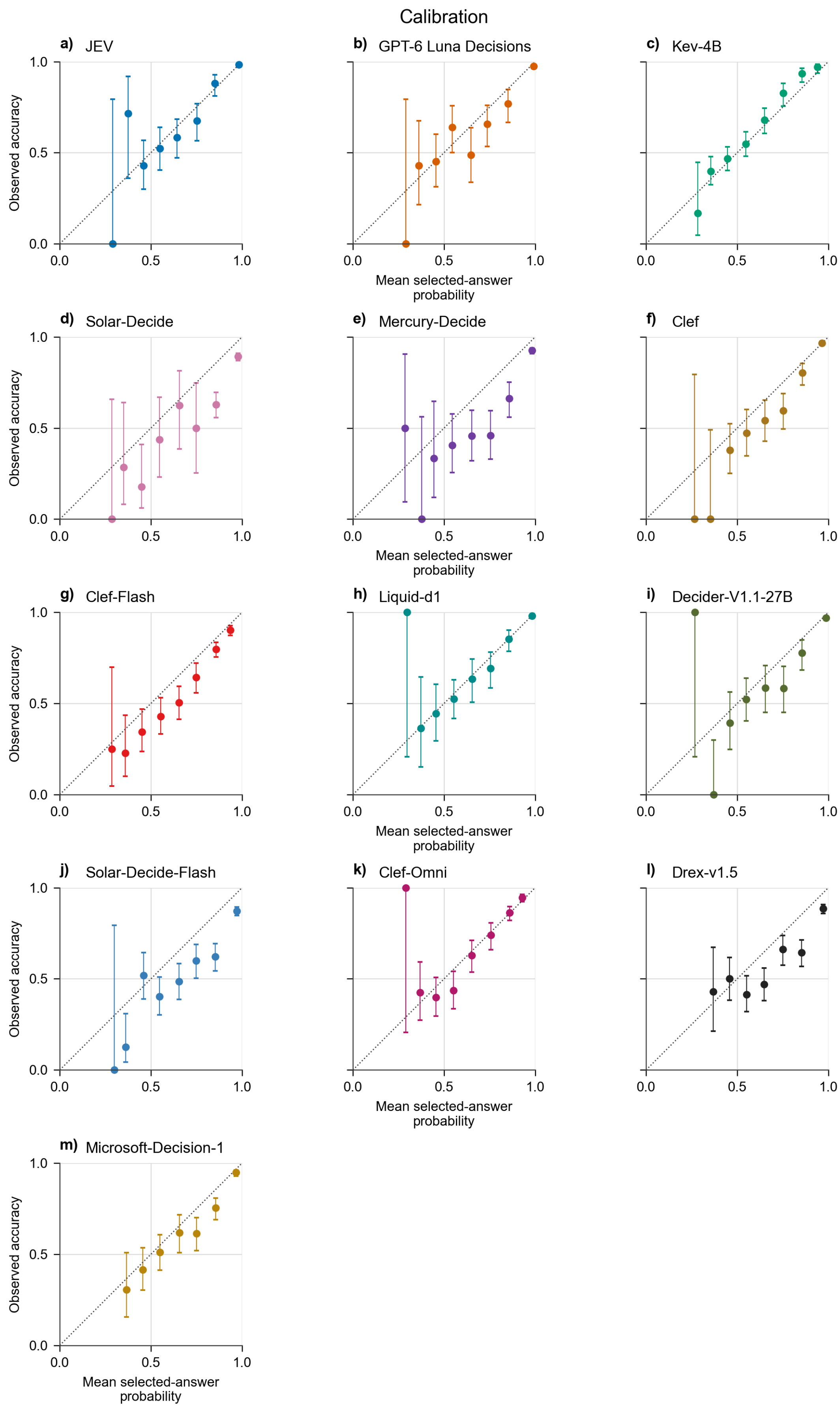
Calibration
a) JEV
b) GPT-6 Luna Decisions
c) Kev-4B
d) Solar-Decide
e) Mercury-Decide
f) Clef
g) Clef-Flash
h) Liquid-d1
i) Decider-V1.1-27B
j) Solar-Decide-Flash
k) Clef-Omni
l) Drex-v1.5
m) Microsoft-Decision-1
Observed accuracy
Mean selected-answer probability
0.0
0.5
1.0

**Extended Data Fig. 3. Error rate versus coverage across all 13 decision models.** Panels a-m show the same analysis as the upper panels of Fig. 2, extended to all 13 decision models. For each model, answers were retained if the native probability assigned to the selected answer was at or above a cutoff. The horizontal axis shows coverage, the percentage of eligible answers retained, and the vertical axis shows the percentage of retained answers that were incorrect. For example, a cutoff of 0.90 retained 877 of 1,273 eligible JEV answers (68.89%), including 15 incorrect answers, giving an error rate of 1.71%. Answers assigned identical probabilities were retained or excluded together. The analysis included 1,271 eligible run-1 answers for Luna Decisions and 1,273 for each other model.

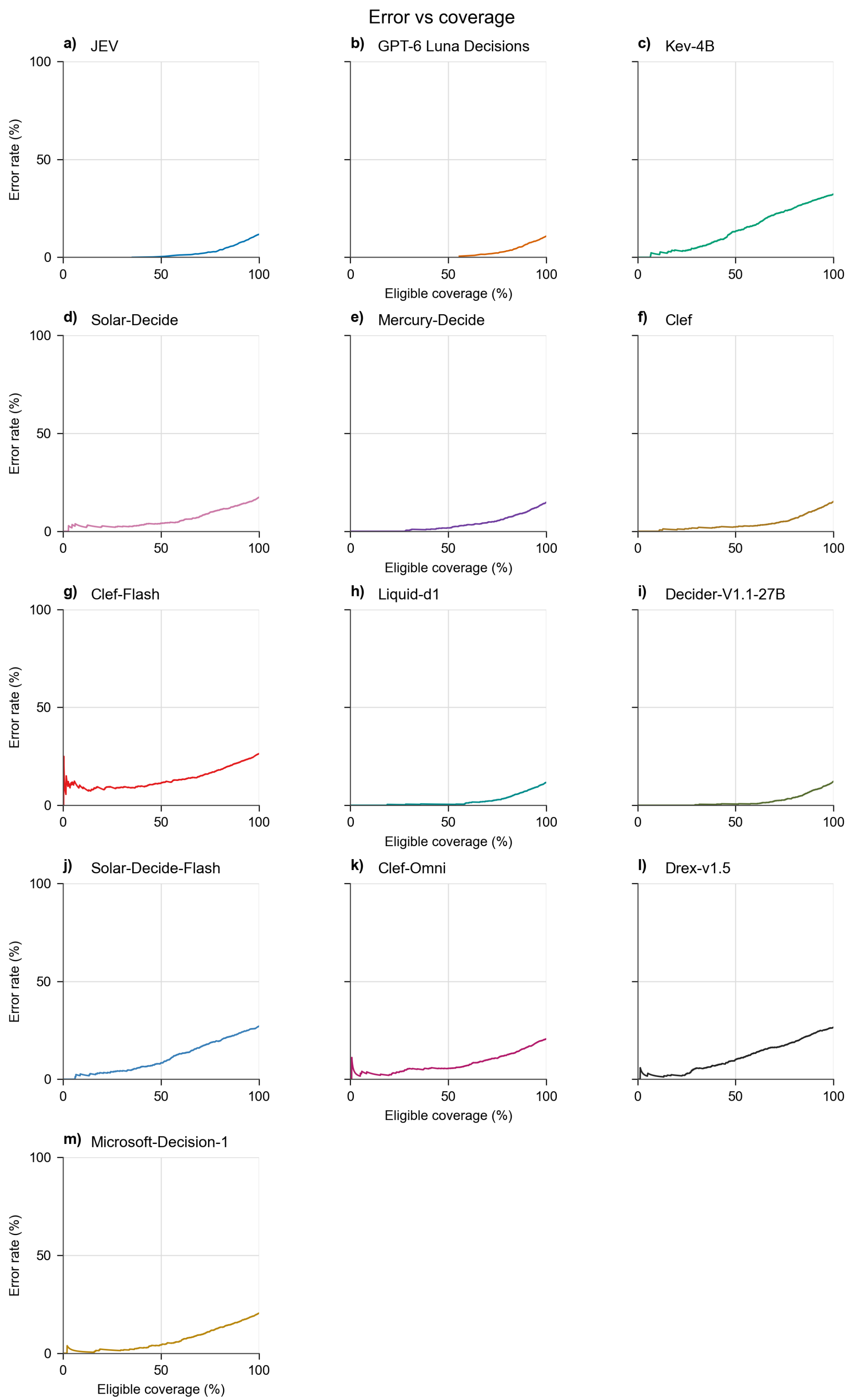
Error vs coverage
a) JEV
b) GPT-6 Luna Decisions
c) Kev-4B
d) Solar-Decide
e) Mercury-Decide
f) Clef
g) Clef-Flash
h) Liquid-d1
i) Decider-V1.1-27B
j) Solar-Decide-Flash
k) Clef-Omni
l) Drex-v1.5
m) Microsoft-Decision-1
Error rate (%)
Eligible coverage (%)
0
50
100

**Extended Data Fig. 4. Error rate versus probability cutoff across all 13 decision models.** Panels a-m show the same analysis as the lower panels of Fig. 2, extended to all 13 decision models. The horizontal axis shows the native selected-answer probability cutoff. At each cutoff, only answers assigned a probability at or above that value were retained; the vertical axis shows the percentage of those answers that were incorrect. For example, at a cutoff of 0.90, JEV retained 877 answers, of which 15 were incorrect, giving an error rate of 1.71%. Extended Data Fig. 3 shows the same error rates against the percentage of eligible answers retained rather than the cutoff. Answers assigned identical probabilities were retained or excluded together. No error rate is defined when a cutoff retains no answers. The analysis included 1,271 eligible run-1 answers for Luna Decisions and 1,273 for each other model.

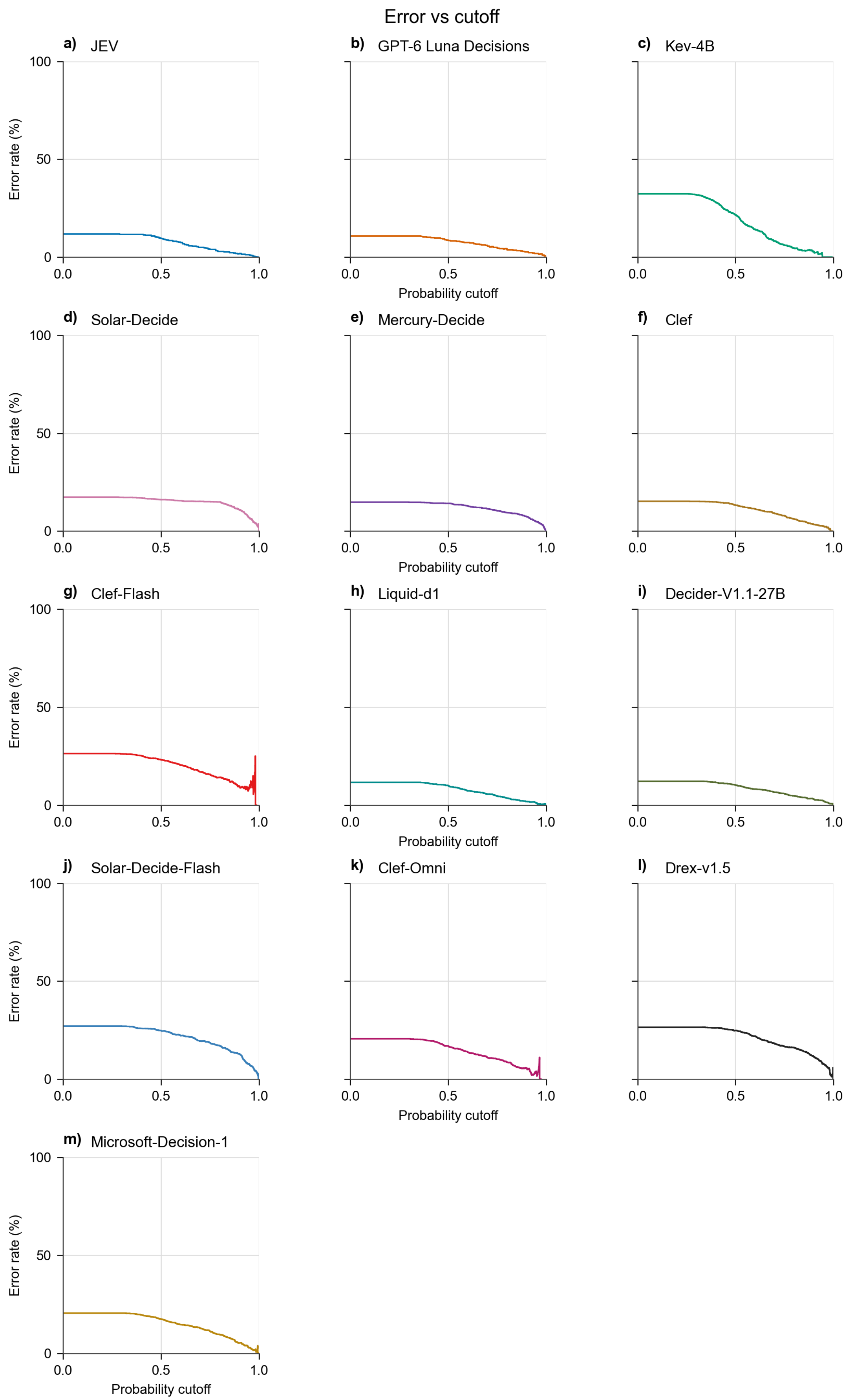

Error vs cutoff
a) JEV
b) GPT-6 Luna Decisions
c) Kev-4B
d) Solar-Decide
e) Mercury-Decide
f) Clef
g) Clef-Flash
h) Liquid-d1
i) Decider-V1.1-27B
j) Solar-Decide-Flash
k) Clef-Omni
l) Drex-v1.5
m) Microsoft-Decision-1
Error rate (%)
Probability cutoff
0
50
100
0.0
0.5
1.0

**Extended Data Table 1. Answer accuracy and agreement across three runs.** Accuracy was calculated using all 1,273 questions in each run, with failed outcomes counted as incorrect. Agreement in the selected answer was calculated only for questions with valid answers in both runs for pairwise comparisons or in all three runs for three-run comparisons. This left 1,271 valid pairs and triples for Luna Decisions and 1,273 for each other model. The pairwise agreement range reports the lowest and highest agreement across runs 1-2, 1-3 and 2-3.

| Model (ID) | Run 1 accuracy (%) | Run 2 accuracy (%) | Run 3 accuracy (%) | Three-run agreement (%) | Valid triples | Pairwise agreement range (%) |
|---|---|---|---|---|---|---|
| JEV 1.13 (D01) | 88.30 | 87.98 | 88.14 | 98.11 | 1,273 | 98.59 to 99.06 |
| GPT-6 Luna Decisions (D02) | 89.08 | 89.08 | 89.08 | 100.00 | 1,271 | 100.00 to 100.00 |
| Kev 4B (D03) | 67.71 | 67.71 | 67.71 | 99.84 | 1,273 | 99.84 to 100.00 |
| Solar Decide (D04) | 82.64 | 82.33 | 82.80 | 92.30 | 1,273 | 94.27 to 94.66 |
| Mercury Decide (D05) | 85.23 | 85.86 | 85.55 | 98.74 | 1,273 | 99.06 to 99.21 |
| Clef (D06) | 84.76 | 84.76 | 84.76 | 100.00 | 1,273 | 100.00 to 100.00 |
| Clef Flash (D07) | 73.68 | 73.68 | 73.68 | 100.00 | 1,273 | 100.00 to 100.00 |
| Liquid d1 (D08) | 88.37 | 88.77 | 88.53 | 98.35 | 1,273 | 98.51 to 99.45 |
| Decider V1.1 27B (D09) | 87.82 | 87.82 | 87.82 | 100.00 | 1,273 | 100.00 to 100.00 |
| Solar Decide Flash (D10) | 72.90 | 72.98 | 72.51 | 96.47 | 1,273 | 97.41 to 97.96 |
| Clef Omni (D11) | 79.42 | 79.50 | 79.50 | 99.92 | 1,273 | 99.92 to 100.00 |
| Drex v1.5 (D12) | 73.53 | 73.13 | 73.13 | 97.88 | 1,273 | 97.88 to 100.00 |
| Microsoft-Decision-1 (D13) | 79.58 | 79.81 | 79.73 | 96.39 | 1,273 | 96.86 to 97.96 |

**Extended Data Table 2. Model configurations and selected serving routes.** Configuration IDs begin with G for generative configurations and D for decision models. Providers refer to the selected serving routes. “API only” indicates documented access through a hosted service for which no public weights were identified, whereas “Open weights” indicates a downloadable model release under the listed licence. All study inference used hosted APIs, including for models with open weights.

| ID | Model | Native effort/mode | Provider | Model availability |
|---|---|---|---|---|
| G01 | GPT-6.1 Sol | Low | OpenAI | API only |
| G02 | GPT-6 Luna | Low | OpenAI | API only |
| G03 | GPT-6 Luna | High | OpenAI | API only |
| G04 | Claude Sonnet 5.5 | Low | Anthropic | API only |
| G05 | Gemini 3.8 Flash | Low | Google AI Studio | API only |
| G06 | DeepSeek V4.1 Flash | Low | InferenceNet | Open weights; MIT |
| G07 | DeepSeek V4.1 Flash | High | InferenceNet | Open weights; MIT |
| G08 | Qwen3.8 Max | Low | Alibaba | API only |
| G09 | Qwen3.8-27B | Low | Mancer 2 | Open weights; Apache-2.0 |
| G10 | Qwen3.8-27B | Xhigh | Mancer 2 | Open weights; Apache-2.0 |
| G11 | Mistral Medium 3.5 | None | Mistral | Open weights; Modified MIT |
| G12 | GLM-5 | Native | Baidu | Open weights; MIT |
| G13 | Kimi K2.5 | Native | SiliconFlow | Open weights; Modified MIT |
| G14 | Llama 4 Maverick | Standard | DigitalOcean | Open weights; Llama 4 Community |
| G15 | Kimi K2.5 | Instant (reasoning disabled) | SiliconFlow | Open weights; Modified MIT |
| D01 | JEV 1.13 | Native | TypeSafe | API only |
| D02 | GPT-6 Luna Decisions | Native Choice | OpenAI | API only |
| D03 | Kev 4B | Native | SiliconFlow | Open weights; Apache-2.0 |
| D04 | Solar Decide | Native | Upstage | API only |
| D05 | Mercury Decide | Native | Inception | API only |
| D06 | Clef | Native | Cloudflare | Open weights; Apache-2.0 |

| ID | Model | Native effort/mode | Provider | Model availability |
|---|---|---|---|---|
| D07 | Clef Flash | Native | Cloudflare | Open weights; Apache-2.0 |
| D08 | Liquid d1 | Native | Liquid | API only |
| D09 | Decider V1.1 27B | Native | Perplexity | Open weights; Apache-2.0 |
| D10 | Solar Decide Flash | Native | Upstage | API only |
| D11 | Clef Omni | Native | Cloudflare | Open weights; Apache-2.0 |
| D12 | Drex v1.5 | Native | DeepInfra | Open weights; Nace.AI Open RAIL-M |
| D13 | Microsoft-Decision-1 | Native | Azure | API only |

# Supplementary material

**Supplementary Interactive Fig. 1**, which shows the joint accuracy, cost, and duration comparison as an interactive 3D plot (3D Pareto frontier), is uploaded at https://oscarfreyer.github.io/decision-model-benchmark/. The file allows readers to rotate and zoom the figure and inspect individual values.

## Supplementary table guide

| Table | Contents | Data rows |
|---|---|---:|
| S1 | Primary paired accuracy comparisons. | 273 |
| S2 | Paired contingency counts and supplementary Newcombe intervals. | 276 |
| S3 | Exploratory native-effort accuracy contrasts. | 3 |
| S4 | Reliability of run-1 native selected-option probabilities. | 130 |
| S5 | Support of native selected-option probabilities in run 1. | 13 |
| S6 | Question-matched correctness and failure transitions. | 39 |

| Table | Contents | Data rows |
|---|---|---|
| S7 | Exact three-run equality of decision-model probabilities. | 26 |
| S8 | Pairwise changes in decision-model probabilities. | 117 |
| S9 | Original-first-attempt and visual-nomination accuracy sensitivities. | 54 |
| S10 | Complete-valid and visual-filtered paired accuracy sensitivities. | 276 |

# Supplementary tables

**Supplementary Table S1. Primary paired accuracy comparisons.** The primary comparison family (S1) included 195 decision-versus-generative comparisons and 78 decision-model pairs, with Holm adjustment for all 273 comparisons. For comparisons between decision models, the model with the later ID was subtracted from the model with the earlier ID. Effects and pointwise 95% question-bootstrap intervals are expressed in percentage points, with negative values favoring Model B. Two-sided exact McNemar P values below .001 are shown as <.001. Of the 273 comparisons, 243 met the adjusted P≤0.05 threshold: 182 decision-versus-generative and 61 decision-model comparisons. Neither JEV nor Luna Decisions differed significantly from Qwen3.8 Max; the comparison between JEV and Luna Decisions had an adjusted P=1.000.

| Model A - Model B | n | Effect, pp | Bootstrap 95% CI, pp | Exact P | Holm P |
|---|---|---|---|---|---|
| JEV 1.13 (D01) - GPT-6.1 Sol Low (G01) | 1,273 | -8.56 | -10.21 to -6.91 | <.001 | <.001 |
| JEV 1.13 (D01) - GPT-6 Luna Low (G02) | 1,273 | -5.89 | -7.62 to -4.16 | <.001 | <.001 |
| JEV 1.13 (D01) - GPT-6 Luna High (G03) | 1,273 | -6.91 | -8.64 to -5.18 | <.001 | <.001 |
| JEV 1.13 (D01) - Claude Sonnet 5.5 Low (G04) | 1,273 | -6.83 | -8.48 to -5.18 | <.001 | <.001 |
| JEV 1.13 (D01) - Gemini 3.8 Flash Low (G05) | 1,273 | -8.48 | -10.21 to -6.83 | <.001 | <.001 |

| Model A - Model B | n | Effect, pp | Bootstrap 95% CI, pp | Exact P | Holm P |
|---|---|---|---|---|---|
| JEV 1.13 (D01) - DeepSeek V4.1 Flash Low (G06) | 1,273 | -7.23 | -8.88 to -5.58 | <.001 | <.001 |
| JEV 1.13 (D01) - DeepSeek V4.1 Flash High (G07) | 1,273 | -6.28 | -7.93 to -4.56 | <.001 | <.001 |
| JEV 1.13 (D01) - Qwen3.8 Max Low (G08) | 1,273 | -1.34 | -3.06 to 0.39 | .159 | 1.000 |
| JEV 1.13 (D01) - Qwen3.8-27B Low (G09) | 1,273 | -5.81 | -7.62 to -4.01 | <.001 | <.001 |
| JEV 1.13 (D01) - Qwen3.8-27B Xhigh (G10) | 1,273 | -7.15 | -8.88 to -5.42 | <.001 | <.001 |
| JEV 1.13 (D01) - Mistral Medium 3.5 None (G11) | 1,273 | 9.98 | 7.86 to 12.10 | <.001 | <.001 |
| JEV 1.13 (D01) - GLM-5 Native (G12) | 1,273 | -6.68 | -8.41 to -4.95 | <.001 | <.001 |
| JEV 1.13 (D01) - Kimi K2.5 Native (G13) | 1,273 | -6.21 | -7.93 to -4.48 | <.001 | <.001 |
| JEV 1.13 (D01) - Llama 4 Maverick Standard (G14) | 1,273 | 4.48 | 2.51 to 6.52 | <.001 | <.001 |
| JEV 1.13 (D01) - Kimi K2.5 Instant (G15) | 1,273 | 3.30 | 1.41 to 5.26 | .001 | .040 |
| GPT-6 Luna Decisions (D02) - GPT-6.1 Sol Low (G01) | 1,273 | -7.78 | -9.43 to -6.21 | <.001 | <.001 |
| GPT-6 Luna Decisions (D02) - | 1,273 | -5.11 | -6.60 to -3.61 | <.001 | <.001 |

| Model A - Model B | n | Effect, pp | Bootstrap 95% CI, pp | Exact P | Holm P |
|---|---|---|---|---|---|
| GPT-6 Luna Low (G02) | | | | | |
| GPT-6 Luna Decisions (D02) - GPT-6 Luna High (G03) | 1,273 | -6.13 | -7.70 to -4.63 | <.001 | <.001 |
| GPT-6 Luna Decisions (D02) - Claude Sonnet 5.5 Low (G04) | 1,273 | -6.05 | -7.70 to -4.48 | <.001 | <.001 |
| GPT-6 Luna Decisions (D02) - Gemini 3.8 Flash Low (G05) | 1,273 | -7.70 | -9.35 to -6.05 | <.001 | <.001 |
| GPT-6 Luna Decisions (D02) - DeepSeek V4.1 Flash Low (G06) | 1,273 | -6.44 | -8.17 to -4.71 | <.001 | <.001 |
| GPT-6 Luna Decisions (D02) - DeepSeek V4.1 Flash High (G07) | 1,273 | -5.50 | -7.15 to -3.85 | <.001 | <.001 |
| GPT-6 Luna Decisions (D02) - Qwen3.8 Max Low (G08) | 1,273 | -0.55 | -2.28 to 1.10 | .592 | 1.000 |
| GPT-6 Luna Decisions (D02) - Qwen3.8-27B Low (G09) | 1,273 | -5.03 | -6.68 to -3.38 | <.001 | <.001 |
| GPT-6 Luna Decisions (D02) - Qwen3.8-27B Xhigh (G10) | 1,273 | -6.36 | -8.01 to -4.71 | <.001 | <.001 |

| Model A - Model B | n | Effect, pp | Bootstrap 95% CI, pp | Exact P | Holm P |
|---|---|---|---|---|---|
| GPT-6 Luna Decisions (D02) - Mistral Medium 3.5 None (G11) | 1,273 | 10.76 | 8.64 to 12.88 | <.001 | <.001 |
| GPT-6 Luna Decisions (D02) - GLM-5 Native (G12) | 1,273 | -5.89 | -7.62 to -4.16 | <.001 | <.001 |
| GPT-6 Luna Decisions (D02) - Kimi K2.5 Native (G13) | 1,273 | -5.42 | -7.15 to -3.69 | <.001 | <.001 |
| GPT-6 Luna Decisions (D02) - Llama 4 Maverick Standard (G14) | 1,273 | 5.26 | 3.30 to 7.31 | <.001 | <.001 |
| GPT-6 Luna Decisions (D02) - Kimi K2.5 Instant (G15) | 1,273 | 4.08 | 2.20 to 6.05 | <.001 | .002 |
| Kev 4B (D03) - GPT-6.1 Sol Low (G01) | 1,273 | -29.14 | -31.74 to -26.55 | <.001 | <.001 |
| Kev 4B (D03) - GPT-6 Luna Low (G02) | 1,273 | -26.47 | -29.14 to -23.88 | <.001 | <.001 |
| Kev 4B (D03) - GPT-6 Luna High (G03) | 1,273 | -27.49 | -30.16 to -24.90 | <.001 | <.001 |
| Kev 4B (D03) - Claude Sonnet 5.5 Low (G04) | 1,273 | -27.42 | -30.09 to -24.74 | <.001 | <.001 |
| Kev 4B (D03) - Gemini 3.8 Flash Low (G05) | 1,273 | -29.07 | -31.66 to -26.47 | <.001 | <.001 |
| Kev 4B (D03) - DeepSeek V4.1 Flash Low (G06) | 1,273 | -27.81 | -30.40 to -25.22 | <.001 | <.001 |

| Model A - Model B | n | Effect, pp | Bootstrap 95% CI, pp | Exact P | Holm P |
|---|---|---|---|---|---|
| Kev 4B (D03) - DeepSeek V4.1 Flash High (G07) | 1,273 | -26.87 | -29.54 to -24.19 | <.001 | <.001 |
| Kev 4B (D03) - Qwen3.8 Max Low (G08) | 1,273 | -21.92 | -24.59 to -19.32 | <.001 | <.001 |
| Kev 4B (D03) - Qwen3.8-27B Low (G09) | 1,273 | -26.39 | -28.99 to -23.80 | <.001 | <.001 |
| Kev 4B (D03) - Qwen3.8-27B Xhigh (G10) | 1,273 | -27.73 | -30.40 to -25.14 | <.001 | <.001 |
| Kev 4B (D03) - Mistral Medium 3.5 None (G11) | 1,273 | -10.60 | -13.20 to -8.01 | <.001 | <.001 |
| Kev 4B (D03) - GLM-5 Native (G12) | 1,273 | -27.26 | -29.85 to -24.67 | <.001 | <.001 |
| Kev 4B (D03) - Kimi K2.5 Native (G13) | 1,273 | -26.79 | -29.38 to -24.12 | <.001 | <.001 |
| Kev 4B (D03) - Llama 4 Maverick Standard (G14) | 1,273 | -16.10 | -18.77 to -13.43 | <.001 | <.001 |
| Kev 4B (D03) - Kimi K2.5 Instant (G15) | 1,273 | -17.28 | -19.87 to -14.69 | <.001 | <.001 |
| Solar Decide (D04) - GPT-6.1 Sol Low (G01) | 1,273 | -14.22 | -16.26 to -12.25 | <.001 | <.001 |
| Solar Decide (D04) - GPT-6 Luna Low (G02) | 1,273 | -11.55 | -13.59 to -9.58 | <.001 | <.001 |
| Solar Decide (D04) - GPT-6 Luna High (G03) | 1,273 | -12.57 | -14.69 to -10.60 | <.001 | <.001 |

| Model A - Model B | n | Effect, pp | Bootstrap 95% CI, pp | Exact P | Holm P |
|---|---|---|---|---|---|
| Solar Decide (D04) - Claude Sonnet 5.5 Low (G04) | 1,273 | -12.49 | -14.53 to -10.53 | <.001 | <.001 |
| Solar Decide (D04) - Gemini 3.8 Flash Low (G05) | 1,273 | -14.14 | -16.18 to -12.10 | <.001 | <.001 |
| Solar Decide (D04) - DeepSeek V4.1 Flash Low (G06) | 1,273 | -12.88 | -14.85 to -11.00 | <.001 | <.001 |
| Solar Decide (D04) - DeepSeek V4.1 Flash High (G07) | 1,273 | -11.94 | -13.90 to -9.98 | <.001 | <.001 |
| Solar Decide (D04) - Qwen3.8 Max Low (G08) | 1,273 | -6.99 | -9.11 to -4.87 | <.001 | <.001 |
| Solar Decide (D04) - Qwen3.8-27B Low (G09) | 1,273 | -11.47 | -13.43 to -9.51 | <.001 | <.001 |
| Solar Decide (D04) - Qwen3.8-27B Xhigh (G10) | 1,273 | -12.80 | -14.77 to -10.84 | <.001 | <.001 |
| Solar Decide (D04) - Mistral Medium 3.5 None (G11) | 1,273 | 4.32 | 2.20 to 6.44 | <.001 | .004 |
| Solar Decide (D04) - GLM-5 Native (G12) | 1,273 | -12.33 | -14.38 to -10.37 | <.001 | <.001 |
| Solar Decide (D04) - Kimi K2.5 Native (G13) | 1,273 | -11.86 | -13.90 to -9.82 | <.001 | <.001 |
| Solar Decide (D04) - Llama 4 Maverick Standard (G14) | 1,273 | -1.18 | -3.38 to 1.10 | .328 | 1.000 |

| Model A - Model B | n | Effect, pp | Bootstrap 95% CI, pp | Exact P | Holm P |
|---|---|---|---|---|---|
| Solar Decide (D04) - Kimi K2.5 Instant (G15) | 1,273 | -2.36 | -4.56 to -0.16 | .043 | 1.000 |
| Mercury Decide (D05) - GPT-6.1 Sol Low (G01) | 1,273 | -11.63 | -13.59 to -9.74 | <.001 | <.001 |
| Mercury Decide (D05) - GPT-6 Luna Low (G02) | 1,273 | -8.96 | -10.92 to -7.07 | <.001 | <.001 |
| Mercury Decide (D05) - GPT-6 Luna High (G03) | 1,273 | -9.98 | -11.94 to -8.09 | <.001 | <.001 |
| Mercury Decide (D05) - Claude Sonnet 5.5 Low (G04) | 1,273 | -9.90 | -11.86 to -8.09 | <.001 | <.001 |
| Mercury Decide (D05) - Gemini 3.8 Flash Low (G05) | 1,273 | -11.55 | -13.51 to -9.58 | <.001 | <.001 |
| Mercury Decide (D05) - DeepSeek V4.1 Flash Low (G06) | 1,273 | -10.29 | -12.18 to -8.48 | <.001 | <.001 |
| Mercury Decide (D05) - DeepSeek V4.1 Flash High (G07) | 1,273 | -9.35 | -11.23 to -7.46 | <.001 | <.001 |
| Mercury Decide (D05) - Qwen3.8 Max Low (G08) | 1,273 | -4.40 | -6.36 to -2.51 | <.001 | <.001 |
| Mercury Decide (D05) - Qwen3.8-27B Low (G09) | 1,273 | -8.88 | -10.68 to -7.15 | <.001 | <.001 |

| Model A - Model B | n | Effect, pp | Bootstrap 95% CI, pp | Exact P | Holm P |
|---|---|---|---|---|---|
| Mercury Decide (D05) - Qwen3.8-27B Xhigh (G10) | 1,273 | -10.21 | -12.02 to -8.41 | <.001 | <.001 |
| Mercury Decide (D05) - Mistral Medium 3.5 None (G11) | 1,273 | 6.91 | 4.71 to 9.03 | <.001 | <.001 |
| Mercury Decide (D05) - GLM-5 Native (G12) | 1,273 | -9.74 | -11.63 to -7.86 | <.001 | <.001 |
| Mercury Decide (D05) - Kimi K2.5 Native (G13) | 1,273 | -9.27 | -11.15 to -7.38 | <.001 | <.001 |
| Mercury Decide (D05) - Llama 4 Maverick Standard (G14) | 1,273 | 1.41 | -0.63 to 3.46 | .197 | 1.000 |
| Mercury Decide (D05) - Kimi K2.5 Instant (G15) | 1,273 | 0.24 | -1.81 to 2.28 | .881 | 1.000 |
| Clef (D06) - GPT-6.1 Sol Low (G01) | 1,273 | -12.10 | -14.06 to -10.13 | <.001 | <.001 |
| Clef (D06) - GPT-6 Luna Low (G02) | 1,273 | -9.43 | -11.39 to -7.54 | <.001 | <.001 |
| Clef (D06) - GPT-6 Luna High (G03) | 1,273 | -10.45 | -12.41 to -8.48 | <.001 | <.001 |
| Clef (D06) - Claude Sonnet 5.5 Low (G04) | 1,273 | -10.37 | -12.41 to -8.48 | <.001 | <.001 |
| Clef (D06) - Gemini 3.8 Flash Low (G05) | 1,273 | -12.02 | -13.98 to -10.05 | <.001 | <.001 |
| Clef (D06) - DeepSeek V4.1 Flash Low (G06) | 1,273 | -10.76 | -12.73 to -8.80 | <.001 | <.001 |

| Model A - Model B | n | Effect, pp | Bootstrap 95% CI, pp | Exact P | Holm P |
|---|---|---|---|---|---|
| Clef (D06) - DeepSeek V4.1 Flash High (G07) | 1,273 | -9.82 | -11.86 to -7.86 | <.001 | <.001 |
| Clef (D06) - Qwen3.8 Max Low (G08) | 1,273 | -4.87 | -6.83 to -2.99 | <.001 | <.001 |
| Clef (D06) - Qwen3.8-27B Low (G09) | 1,273 | -9.35 | -11.23 to -7.54 | <.001 | <.001 |
| Clef (D06) - Qwen3.8-27B Xhigh (G10) | 1,273 | -10.68 | -12.57 to -8.88 | <.001 | <.001 |
| Clef (D06) - Mistral Medium 3.5 None (G11) | 1,273 | 6.44 | 4.24 to 8.64 | <.001 | <.001 |
| Clef (D06) - GLM-5 Native (G12) | 1,273 | -10.21 | -12.18 to -8.25 | <.001 | <.001 |
| Clef (D06) - Kimi K2.5 Native (G13) | 1,273 | -9.74 | -11.70 to -7.78 | <.001 | <.001 |
| Clef (D06) - Llama 4 Maverick Standard (G14) | 1,273 | 0.94 | -1.10 to 2.99 | .407 | 1.000 |
| Clef (D06) - Kimi K2.5 Instant (G15) | 1,273 | -0.24 | -2.28 to 1.81 | .881 | 1.000 |
| Clef Flash (D07) - GPT-6.1 Sol Low (G01) | 1,273 | -23.17 | -25.61 to -20.74 | <.001 | <.001 |
| Clef Flash (D07) - GPT-6 Luna Low (G02) | 1,273 | -20.50 | -23.02 to -18.15 | <.001 | <.001 |
| Clef Flash (D07) - GPT-6 Luna High (G03) | 1,273 | -21.52 | -24.04 to -19.17 | <.001 | <.001 |

| Model A - Model B | n | Effect, pp | Bootstrap 95% CI, pp | Exact P | Holm P |
|---|---|---|---|---|---|
| Clef Flash (D07) - Claude Sonnet 5.5 Low (G04) | 1,273 | -21.45 | -23.88 to -19.01 | <.001 | <.001 |
| Clef Flash (D07) - Gemini 3.8 Flash Low (G05) | 1,273 | -23.10 | -25.53 to -20.66 | <.001 | <.001 |
| Clef Flash (D07) - DeepSeek V4.1 Flash Low (G06) | 1,273 | -21.84 | -24.35 to -19.48 | <.001 | <.001 |
| Clef Flash (D07) - DeepSeek V4.1 Flash High (G07) | 1,273 | -20.90 | -23.41 to -18.46 | <.001 | <.001 |
| Clef Flash (D07) - Qwen3.8 Max Low (G08) | 1,273 | -15.95 | -18.46 to -13.51 | <.001 | <.001 |
| Clef Flash (D07) - Qwen3.8-27B Low (G09) | 1,273 | -20.42 | -22.86 to -17.99 | <.001 | <.001 |
| Clef Flash (D07) - Qwen3.8-27B Xhigh (G10) | 1,273 | -21.76 | -24.19 to -19.40 | <.001 | <.001 |
| Clef Flash (D07) - Mistral Medium 3.5 None (G11) | 1,273 | -4.63 | -6.99 to -2.36 | <.001 | .006 |
| Clef Flash (D07) - GLM-5 Native (G12) | 1,273 | -21.29 | -23.72 to -18.93 | <.001 | <.001 |
| Clef Flash (D07) - Kimi K2.5 Native (G13) | 1,273 | -20.82 | -23.25 to -18.46 | <.001 | <.001 |
| Clef Flash (D07) - Llama 4 Maverick Standard (G14) | 1,273 | -10.13 | -12.57 to -7.70 | <.001 | <.001 |

| Model A - Model B | n | Effect, pp | Bootstrap 95% CI, pp | Exact P | Holm P |
|---|---|---|---|---|---|
| Clef Flash (D07) - Kimi K2.5 Instant (G15) | 1,273 | -11.31 | -13.67 to -8.96 | <.001 | <.001 |
| Liquid d1 (D08) - GPT-6.1 Sol Low (G01) | 1,273 | -8.48 | -10.21 to -6.76 | <.001 | <.001 |
| Liquid d1 (D08) - GPT-6 Luna Low (G02) | 1,273 | -5.81 | -7.62 to -4.01 | <.001 | <.001 |
| Liquid d1 (D08) - GPT-6 Luna High (G03) | 1,273 | -6.83 | -8.64 to -5.11 | <.001 | <.001 |
| Liquid d1 (D08) - Claude Sonnet 5.5 Low (G04) | 1,273 | -6.76 | -8.56 to -5.03 | <.001 | <.001 |
| Liquid d1 (D08) - Gemini 3.8 Flash Low (G05) | 1,273 | -8.41 | -10.13 to -6.68 | <.001 | <.001 |
| Liquid d1 (D08) - DeepSeek V4.1 Flash Low (G06) | 1,273 | -7.15 | -8.88 to -5.50 | <.001 | <.001 |
| Liquid d1 (D08) - DeepSeek V4.1 Flash High (G07) | 1,273 | -6.21 | -7.93 to -4.48 | <.001 | <.001 |
| Liquid d1 (D08) - Qwen3.8 Max Low (G08) | 1,273 | -1.26 | -2.91 to 0.39 | .171 | 1.000 |
| Liquid d1 (D08) - Qwen3.8-27B Low (G09) | 1,273 | -5.73 | -7.38 to -4.08 | <.001 | <.001 |
| Liquid d1 (D08) - Qwen3.8-27B Xhigh (G10) | 1,273 | -7.07 | -8.72 to -5.42 | <.001 | <.001 |

| Model A - Model B | n | Effect, pp | Bootstrap 95% CI, pp | Exact P | Holm P |
|---|---|---|---|---|---|
| Liquid d1 (D08) - Mistral Medium 3.5 None (G11) | 1,273 | 10.05 | 7.93 to 12.18 | <.001 | <.001 |
| Liquid d1 (D08) - GLM-5 Native (G12) | 1,273 | -6.60 | -8.33 to -4.95 | <.001 | <.001 |
| Liquid d1 (D08) - Kimi K2.5 Native (G13) | 1,273 | -6.13 | -7.86 to -4.48 | <.001 | <.001 |
| Liquid d1 (D08) - Llama 4 Maverick Standard (G14) | 1,273 | 4.56 | 2.59 to 6.52 | <.001 | <.001 |
| Liquid d1 (D08) - Kimi K2.5 Instant (G15) | 1,273 | 3.38 | 1.49 to 5.34 | <.001 | .025 |
| Decider V1.1 27B (D09) - GPT-6.1 Sol Low (G01) | 1,273 | -9.03 | -10.84 to -7.31 | <.001 | <.001 |
| Decider V1.1 27B (D09) - GPT-6 Luna Low (G02) | 1,273 | -6.36 | -8.17 to -4.56 | <.001 | <.001 |
| Decider V1.1 27B (D09) - GPT-6 Luna High (G03) | 1,273 | -7.38 | -9.27 to -5.58 | <.001 | <.001 |
| Decider V1.1 27B (D09) - Claude Sonnet 5.5 Low (G04) | 1,273 | -7.31 | -9.11 to -5.50 | <.001 | <.001 |
| Decider V1.1 27B (D09) - Gemini 3.8 Flash Low (G05) | 1,273 | -8.96 | -10.68 to -7.23 | <.001 | <.001 |
| Decider V1.1 27B (D09) - DeepSeek V4.1 Flash Low (G06) | 1,273 | -7.70 | -9.43 to -5.97 | <.001 | <.001 |

| Model A - Model B | n | Effect, pp | Bootstrap 95% CI, pp | Exact P | Holm P |
|---|---|---|---|---|---|
| Decider V1.1 27B (D09) - DeepSeek V4.1 Flash High (G07) | 1,273 | -6.76 | -8.48 to -5.03 | <.001 | <.001 |
| Decider V1.1 27B (D09) - Qwen3.8 Max Low (G08) | 1,273 | -1.81 | -3.69 to 0.00 | .075 | 1.000 |
| Decider V1.1 27B (D09) - Qwen3.8-27B Low (G09) | 1,273 | -6.28 | -8.09 to -4.56 | <.001 | <.001 |
| Decider V1.1 27B (D09) - Qwen3.8-27B Xhigh (G10) | 1,273 | -7.62 | -9.35 to -5.97 | <.001 | <.001 |
| Decider V1.1 27B (D09) - Mistral Medium 3.5 None (G11) | 1,273 | 9.51 | 7.23 to 11.78 | <.001 | <.001 |
| Decider V1.1 27B (D09) - GLM-5 Native (G12) | 1,273 | -7.15 | -8.88 to -5.42 | <.001 | <.001 |
| Decider V1.1 27B (D09) - Kimi K2.5 Native (G13) | 1,273 | -6.68 | -8.41 to -4.87 | <.001 | <.001 |
| Decider V1.1 27B (D09) - Llama 4 Maverick Standard (G14) | 1,273 | 4.01 | 1.96 to 6.13 | <.001 | .006 |
| Decider V1.1 27B (D09) - Kimi K2.5 Instant (G15) | 1,273 | 2.83 | 0.79 to 4.95 | .010 | .261 |
| Solar Decide Flash (D10) - GPT-6.1 Sol Low (G01) | 1,273 | -23.96 | -26.39 to -21.52 | <.001 | <.001 |

| Model A - Model B | n | Effect, pp | Bootstrap 95% CI, pp | Exact P | Holm P |
|---|---|---|---|---|---|
| Solar Decide Flash (D10) - GPT-6 Luna Low (G02) | 1,273 | -21.29 | -23.80 to -18.85 | <.001 | <.001 |
| Solar Decide Flash (D10) - GPT-6 Luna High (G03) | 1,273 | -22.31 | -24.82 to -19.87 | <.001 | <.001 |
| Solar Decide Flash (D10) - Claude Sonnet 5.5 Low (G04) | 1,273 | -22.23 | -24.74 to -19.80 | <.001 | <.001 |
| Solar Decide Flash (D10) - Gemini 3.8 Flash Low (G05) | 1,273 | -23.88 | -26.32 to -21.45 | <.001 | <.001 |
| Solar Decide Flash (D10) - DeepSeek V4.1 Flash Low (G06) | 1,273 | -22.62 | -25.06 to -20.27 | <.001 | <.001 |
| Solar Decide Flash (D10) - DeepSeek V4.1 Flash High (G07) | 1,273 | -21.68 | -24.12 to -19.25 | <.001 | <.001 |
| Solar Decide Flash (D10) - Qwen3.8 Max Low (G08) | 1,273 | -16.73 | -19.25 to -14.30 | <.001 | <.001 |
| Solar Decide Flash (D10) - Qwen3.8-27B Low (G09) | 1,273 | -21.21 | -23.64 to -18.85 | <.001 | <.001 |
| Solar Decide Flash (D10) - Qwen3.8-27B Xhigh (G10) | 1,273 | -22.55 | -24.98 to -20.11 | <.001 | <.001 |
| Solar Decide Flash (D10) - Mistral Medium 3.5 None (G11) | 1,273 | -5.42 | -7.70 to -3.14 | <.001 | <.001 |

| Model A - Model B | n | Effect, pp | Bootstrap 95% CI, pp | Exact P | Holm P |
|---|---|---|---|---|---|
| Solar Decide Flash (D10) - GLM-5 Native (G12) | 1,273 | -22.07 | -24.51 to -19.72 | <.001 | <.001 |
| Solar Decide Flash (D10) - Kimi K2.5 Native (G13) | 1,273 | -21.60 | -24.04 to -19.17 | <.001 | <.001 |
| Solar Decide Flash (D10) - Llama 4 Maverick Standard (G14) | 1,273 | -10.92 | -13.35 to -8.48 | <.001 | <.001 |
| Solar Decide Flash (D10) - Kimi K2.5 Instant (G15) | 1,273 | -12.10 | -14.53 to -9.66 | <.001 | <.001 |
| Clef Omni (D11) - GPT-6.1 Sol Low (G01) | 1,273 | -17.44 | -19.64 to -15.24 | <.001 | <.001 |
| Clef Omni (D11) - GPT-6 Luna Low (G02) | 1,273 | -14.77 | -17.05 to -12.49 | <.001 | <.001 |
| Clef Omni (D11) - GPT-6 Luna High (G03) | 1,273 | -15.79 | -18.15 to -13.43 | <.001 | <.001 |
| Clef Omni (D11) - Claude Sonnet 5.5 Low (G04) | 1,273 | -15.71 | -17.91 to -13.43 | <.001 | <.001 |
| Clef Omni (D11) - Gemini 3.8 Flash Low (G05) | 1,273 | -17.36 | -19.56 to -15.16 | <.001 | <.001 |
| Clef Omni (D11) - DeepSeek V4.1 Flash Low (G06) | 1,273 | -16.10 | -18.30 to -13.83 | <.001 | <.001 |
| Clef Omni (D11) - DeepSeek V4.1 Flash High (G07) | 1,273 | -15.16 | -17.44 to -12.88 | <.001 | <.001 |

| Model A - Model B | n | Effect, pp | Bootstrap 95% CI, pp | Exact P | Holm P |
|---|---|---|---|---|---|
| Clef Omni (D11) - Qwen3.8 Max Low (G08) | 1,273 | -10.21 | -12.41 to -8.01 | <.001 | <.001 |
| Clef Omni (D11) - Qwen3.8-27B Low (G09) | 1,273 | -14.69 | -16.89 to -12.49 | <.001 | <.001 |
| Clef Omni (D11) - Qwen3.8-27B Xhigh (G10) | 1,273 | -16.03 | -18.22 to -13.83 | <.001 | <.001 |
| Clef Omni (D11) - Mistral Medium 3.5 None (G11) | 1,273 | 1.10 | -1.10 to 3.30 | .365 | 1.000 |
| Clef Omni (D11) - GLM-5 Native (G12) | 1,273 | -15.55 | -17.75 to -13.28 | <.001 | <.001 |
| Clef Omni (D11) - Kimi K2.5 Native (G13) | 1,273 | -15.08 | -17.28 to -12.88 | <.001 | <.001 |
| Clef Omni (D11) - Llama 4 Maverick Standard (G14) | 1,273 | -4.40 | -6.68 to -2.12 | <.001 | .008 |
| Clef Omni (D11) - Kimi K2.5 Instant (G15) | 1,273 | -5.58 | -7.70 to -3.46 | <.001 | <.001 |
| Drex v1.5 (D12) - GPT-6.1 Sol Low (G01) | 1,273 | -23.33 | -25.84 to -20.82 | <.001 | <.001 |
| Drex v1.5 (D12) - GPT-6 Luna Low (G02) | 1,273 | -20.66 | -23.25 to -18.15 | <.001 | <.001 |
| Drex v1.5 (D12) - GPT-6 Luna High (G03) | 1,273 | -21.68 | -24.19 to -19.17 | <.001 | <.001 |

| Model A - Model B | n | Effect, pp | Bootstrap 95% CI, pp | Exact P | Holm P |
|---|---|---|---|---|---|
| Drex v1.5 (D12) - Claude Sonnet 5.5 Low (G04) | 1,273 | -21.60 | -24.12 to -19.17 | <.001 | <.001 |
| Drex v1.5 (D12) - Gemini 3.8 Flash Low (G05) | 1,273 | -23.25 | -25.69 to -20.82 | <.001 | <.001 |
| Drex v1.5 (D12) - DeepSeek V4.1 Flash Low (G06) | 1,273 | -22.00 | -24.51 to -19.48 | <.001 | <.001 |
| Drex v1.5 (D12) - DeepSeek V4.1 Flash High (G07) | 1,273 | -21.05 | -23.57 to -18.46 | <.001 | <.001 |
| Drex v1.5 (D12) - Qwen3.8 Max Low (G08) | 1,273 | -16.10 | -18.54 to -13.67 | <.001 | <.001 |
| Drex v1.5 (D12) - Qwen3.8-27B Low (G09) | 1,273 | -20.58 | -23.02 to -18.15 | <.001 | <.001 |
| Drex v1.5 (D12) - Qwen3.8-27B Xhigh (G10) | 1,273 | -21.92 | -24.35 to -19.40 | <.001 | <.001 |
| Drex v1.5 (D12) - Mistral Medium 3.5 None (G11) | 1,273 | -4.79 | -7.31 to -2.20 | <.001 | .011 |
| Drex v1.5 (D12) - GLM-5 Native (G12) | 1,273 | -21.45 | -23.88 to -18.93 | <.001 | <.001 |
| Drex v1.5 (D12) - Kimi K2.5 Native (G13) | 1,273 | -20.97 | -23.49 to -18.54 | <.001 | <.001 |
| Drex v1.5 (D12) - Llama 4 Maverick Standard (G14) | 1,273 | -10.29 | -12.80 to -7.78 | <.001 | <.001 |

| Model A - Model B | n | Effect, pp | Bootstrap 95% CI, pp | Exact P | Holm P |
|---|---|---|---|---|---|
| Drex v1.5 (D12) - Kimi K2.5 Instant (G15) | 1,273 | -11.47 | -13.98 to -8.96 | <.001 | <.001 |
| Microsoft-Decision-1 (D13) - GPT-6.1 Sol Low (G01) | 1,273 | -17.28 | -19.56 to -15.00 | <.001 | <.001 |
| Microsoft-Decision-1 (D13) - GPT-6 Luna Low (G02) | 1,273 | -14.61 | -16.97 to -12.33 | <.001 | <.001 |
| Microsoft-Decision-1 (D13) - GPT-6 Luna High (G03) | 1,273 | -15.63 | -17.99 to -13.35 | <.001 | <.001 |
| Microsoft-Decision-1 (D13) - Claude Sonnet 5.5 Low (G04) | 1,273 | -15.55 | -17.83 to -13.35 | <.001 | <.001 |
| Microsoft-Decision-1 (D13) - Gemini 3.8 Flash Low (G05) | 1,273 | -17.20 | -19.48 to -14.93 | <.001 | <.001 |
| Microsoft-Decision-1 (D13) - DeepSeek V4.1 Flash Low (G06) | 1,273 | -15.95 | -18.22 to -13.75 | <.001 | <.001 |
| Microsoft-Decision-1 (D13) - DeepSeek V4.1 Flash High (G07) | 1,273 | -15.00 | -17.28 to -12.73 | <.001 | <.001 |
| Microsoft-Decision-1 (D13) - Qwen3.8 Max Low (G08) | 1,273 | -10.05 | -12.33 to -7.78 | <.001 | <.001 |
| Microsoft-Decision-1 (D13) - Qwen3.8-27B Low (G09) | 1,273 | -14.53 | -16.81 to -12.33 | <.001 | <.001 |

| Model A - Model B | n | Effect, pp | Bootstrap 95% CI, pp | Exact P | Holm P |
|---|---|---|---|---|---|
| Microsoft-Decision-1 (D13) - Qwen3.8-27B Xhigh (G10) | 1,273 | -15.87 | -18.07 to -13.67 | <.001 | <.001 |
| Microsoft-Decision-1 (D13) - Mistral Medium 3.5 None (G11) | 1,273 | 1.26 | -1.18 to 3.61 | .339 | 1.000 |
| Microsoft-Decision-1 (D13) - GLM-5 Native (G12) | 1,273 | -15.40 | -17.67 to -13.12 | <.001 | <.001 |
| Microsoft-Decision-1 (D13) - Kimi K2.5 Native (G13) | 1,273 | -14.93 | -17.20 to -12.65 | <.001 | <.001 |
| Microsoft-Decision-1 (D13) - Llama 4 Maverick Standard (G14) | 1,273 | -4.24 | -6.52 to -1.96 | <.001 | .015 |
| Microsoft-Decision-1 (D13) - Kimi K2.5 Instant (G15) | 1,273 | -5.42 | -7.70 to -3.14 | <.001 | <.001 |
| JEV 1.13 (D01) - GPT-6 Luna Decisions (D02) | 1,273 | -0.79 | -2.51 to 0.94 | .430 | 1.000 |
| JEV 1.13 (D01) - Kev 4B (D03) | 1,273 | 20.58 | 17.99 to 23.17 | <.001 | <.001 |
| JEV 1.13 (D01) - Solar Decide (D04) | 1,273 | 5.66 | 3.53 to 7.78 | <.001 | <.001 |
| JEV 1.13 (D01) - Mercury Decide (D05) | 1,273 | 3.06 | 1.26 to 4.95 | .002 | .057 |
| JEV 1.13 (D01) - Clef (D06) | 1,273 | 3.53 | 1.65 to 5.42 | <.001 | .011 |
| JEV 1.13 (D01) - Clef Flash (D07) | 1,273 | 14.61 | 12.25 to 17.05 | <.001 | <.001 |

| Model A - Model B | n | Effect, pp | Bootstrap 95% CI, pp | Exact P | Holm P |
|---|---|---|---|---|---|
| JEV 1.13 (D01) - Liquid d1 (D08) | 1,273 | -0.08 | -1.73 to 1.57 | 1.000 | 1.000 |
| JEV 1.13 (D01) - Decider V1.1 27B (D09) | 1,273 | 0.47 | -1.26 to 2.20 | .654 | 1.000 |
| JEV 1.13 (D01) - Solar Decide Flash (D10) | 1,273 | 15.40 | 12.88 to 17.91 | <.001 | <.001 |
| JEV 1.13 (D01) - Clef Omni (D11) | 1,273 | 8.88 | 6.68 to 11.15 | <.001 | <.001 |
| JEV 1.13 (D01) - Drex v1.5 (D12) | 1,273 | 14.77 | 12.25 to 17.36 | <.001 | <.001 |
| JEV 1.13 (D01) - Microsoft-Decision-1 (D13) | 1,273 | 8.72 | 6.52 to 10.92 | <.001 | <.001 |
| GPT-6 Luna Decisions (D02) - Kev 4B (D03) | 1,273 | 21.37 | 18.70 to 24.04 | <.001 | <.001 |
| GPT-6 Luna Decisions (D02) - Solar Decide (D04) | 1,273 | 6.44 | 4.32 to 8.56 | <.001 | <.001 |
| GPT-6 Luna Decisions (D02) - Mercury Decide (D05) | 1,273 | 3.85 | 1.96 to 5.81 | <.001 | .005 |
| GPT-6 Luna Decisions (D02) - Clef (D06) | 1,273 | 4.32 | 2.51 to 6.13 | <.001 | <.001 |
| GPT-6 Luna Decisions (D02) - Clef Flash (D07) | 1,273 | 15.40 | 12.96 to 17.91 | <.001 | <.001 |
| GPT-6 Luna Decisions (D02) - Liquid d1 (D08) | 1,273 | 0.71 | -1.10 to 2.44 | .488 | 1.000 |

| Model A - Model B | n | Effect, pp | Bootstrap 95% CI, pp | Exact P | Holm P |
|---|---|---|---|---|---|
| GPT-6 Luna Decisions (D02) - Decider V1.1 27B (D09) | 1,273 | 1.26 | -0.63 to 3.06 | .214 | 1.000 |
| GPT-6 Luna Decisions (D02) - Solar Decide Flash (D10) | 1,273 | 16.18 | 13.67 to 18.62 | <.001 | <.001 |
| GPT-6 Luna Decisions (D02) - Clef Omni (D11) | 1,273 | 9.66 | 7.46 to 11.94 | <.001 | <.001 |
| GPT-6 Luna Decisions (D02) - Drex v1.5 (D12) | 1,273 | 15.55 | 13.12 to 18.07 | <.001 | <.001 |
| GPT-6 Luna Decisions (D02) - Microsoft-Decision-1 (D13) | 1,273 | 9.51 | 7.23 to 11.86 | <.001 | <.001 |
| Kev 4B (D03) - Solar Decide (D04) | 1,273 | -14.93 | -17.60 to -12.33 | <.001 | <.001 |
| Kev 4B (D03) - Mercury Decide (D05) | 1,273 | -17.52 | -20.19 to -14.93 | <.001 | <.001 |
| Kev 4B (D03) - Clef (D06) | 1,273 | -17.05 | -19.64 to -14.45 | <.001 | <.001 |
| Kev 4B (D03) - Clef Flash (D07) | 1,273 | -5.97 | -8.48 to -3.38 | <.001 | <.001 |
| Kev 4B (D03) - Liquid d1 (D08) | 1,273 | -20.66 | -23.25 to -18.07 | <.001 | <.001 |
| Kev 4B (D03) - Decider V1.1 27B (D09) | 1,273 | -20.11 | -22.78 to -17.44 | <.001 | <.001 |
| Kev 4B (D03) - Solar Decide Flash (D10) | 1,273 | -5.18 | -7.86 to -2.51 | <.001 | .007 |

| Model A - Model B | n | Effect, pp | Bootstrap 95% CI, pp | Exact P | Holm P |
|---|---|---|---|---|---|
| Kev 4B (D03) - Clef Omni (D11) | 1,273 | -11.70 | -14.22 to -9.19 | <.001 | <.001 |
| Kev 4B (D03) - Drex v1.5 (D12) | 1,273 | -5.81 | -8.41 to -3.22 | <.001 | <.001 |
| Kev 4B (D03) - Microsoft-Decision-1 (D13) | 1,273 | -11.86 | -14.45 to -9.27 | <.001 | <.001 |
| Solar Decide (D04) - Mercury Decide (D05) | 1,273 | -2.59 | -4.71 to -0.39 | .026 | .647 |
| Solar Decide (D04) - Clef (D06) | 1,273 | -2.12 | -4.24 to 0.16 | .073 | 1.000 |
| Solar Decide (D04) - Clef Flash (D07) | 1,273 | 8.96 | 6.52 to 11.47 | <.001 | <.001 |
| Solar Decide (D04) - Liquid d1 (D08) | 1,273 | -5.73 | -7.93 to -3.61 | <.001 | <.001 |
| Solar Decide (D04) - Decider V1.1 27B (D09) | 1,273 | -5.18 | -7.38 to -2.99 | <.001 | <.001 |
| Solar Decide (D04) - Solar Decide Flash (D10) | 1,273 | 9.74 | 7.93 to 11.55 | <.001 | <.001 |
| Solar Decide (D04) - Clef Omni (D11) | 1,273 | 3.22 | 0.86 to 5.50 | .009 | .250 |
| Solar Decide (D04) - Drex v1.5 (D12) | 1,273 | 9.11 | 6.44 to 11.79 | <.001 | <.001 |
| Solar Decide (D04) - Microsoft-Decision-1 (D13) | 1,273 | 3.06 | 0.63 to 5.50 | .018 | .469 |
| Mercury Decide (D05) - Clef (D06) | 1,273 | 0.47 | -1.10 to 2.04 | .621 | 1.000 |

| Model A - Model B | n | Effect, pp | Bootstrap 95% CI, pp | Exact P | Holm P |
|---|---|---|---|---|---|
| Mercury Decide (D05) - Clef Flash (D07) | 1,273 | 11.55 | 9.19 to 13.98 | <.001 | <.001 |
| Mercury Decide (D05) - Liquid d1 (D08) | 1,273 | -3.14 | -5.03 to -1.34 | .001 | .035 |
| Mercury Decide (D05) - Decider V1.1 27B (D09) | 1,273 | -2.59 | -4.32 to -0.94 | .003 | .085 |
| Mercury Decide (D05) - Solar Decide Flash (D10) | 1,273 | 12.33 | 9.90 to 14.69 | <.001 | <.001 |
| Mercury Decide (D05) - Clef Omni (D11) | 1,273 | 5.81 | 3.53 to 8.01 | <.001 | <.001 |
| Mercury Decide (D05) - Drex v1.5 (D12) | 1,273 | 11.70 | 9.11 to 14.30 | <.001 | <.001 |
| Mercury Decide (D05) - Microsoft-Decision-1 (D13) | 1,273 | 5.66 | 3.38 to 7.93 | <.001 | <.001 |
| Clef (D06) - Clef Flash (D07) | 1,273 | 11.08 | 8.80 to 13.43 | <.001 | <.001 |
| Clef (D06) - Liquid d1 (D08) | 1,273 | -3.61 | -5.50 to -1.81 | <.001 | .006 |
| Clef (D06) - Decider V1.1 27B (D09) | 1,273 | -3.06 | -4.87 to -1.34 | <.001 | .030 |
| Clef (D06) - Solar Decide Flash (D10) | 1,273 | 11.86 | 9.51 to 14.30 | <.001 | <.001 |
| Clef (D06) - Clef Omni (D11) | 1,273 | 5.34 | 3.14 to 7.46 | <.001 | <.001 |
| Clef (D06) - Drex v1.5 (D12) | 1,273 | 11.23 | 8.72 to 13.75 | <.001 | <.001 |

| Model A - Model B | n | Effect, pp | Bootstrap 95% CI, pp | Exact P | Holm P |
|---|---|---|---|---|---|
| Clef (D06) - Microsoft-Decision-1 (D13) | 1,273 | 5.18 | 2.91 to 7.46 | <.001 | <.001 |
| Clef Flash (D07) - Liquid d1 (D08) | 1,273 | -14.69 | -17.05 to -12.41 | <.001 | <.001 |
| Clef Flash (D07) - Decider V1.1 27B (D09) | 1,273 | -14.14 | -16.58 to -11.78 | <.001 | <.001 |
| Clef Flash (D07) - Solar Decide Flash (D10) | 1,273 | 0.79 | -1.73 to 3.30 | .583 | 1.000 |
| Clef Flash (D07) - Clef Omni (D11) | 1,273 | -5.73 | -8.17 to -3.30 | <.001 | <.001 |
| Clef Flash (D07) - Drex v1.5 (D12) | 1,273 | 0.16 | -2.28 to 2.51 | .948 | 1.000 |
| Clef Flash (D07) - Microsoft-Decision-1 (D13) | 1,273 | -5.89 | -8.17 to -3.53 | <.001 | <.001 |
| Liquid d1 (D08) - Decider V1.1 27B (D09) | 1,273 | 0.55 | -1.34 to 2.36 | .611 | 1.000 |
| Liquid d1 (D08) - Solar Decide Flash (D10) | 1,273 | 15.48 | 13.04 to 17.91 | <.001 | <.001 |
| Liquid d1 (D08) - Clef Omni (D11) | 1,273 | 8.96 | 6.83 to 11.08 | <.001 | <.001 |
| Liquid d1 (D08) - Drex v1.5 (D12) | 1,273 | 14.85 | 12.41 to 17.28 | <.001 | <.001 |
| Liquid d1 (D08) - Microsoft-Decision-1 (D13) | 1,273 | 8.80 | 6.52 to 11.00 | <.001 | <.001 |

| Model A - Model B | n | Effect, pp | Bootstrap 95% CI, pp | Exact P | Holm P |
|---|---|---|---|---|---|
| Decider V1.1 27B (D09) - Solar Decide Flash (D10) | 1,273 | 14.93 | 12.41 to 17.44 | <.001 | <.001 |
| Decider V1.1 27B (D09) - Clef Omni (D11) | 1,273 | 8.41 | 6.21 to 10.68 | <.001 | <.001 |
| Decider V1.1 27B (D09) - Drex v1.5 (D12) | 1,273 | 14.30 | 11.70 to 16.81 | <.001 | <.001 |
| Decider V1.1 27B (D09) - Microsoft-Decision-1 (D13) | 1,273 | 8.25 | 6.05 to 10.53 | <.001 | <.001 |
| Solar Decide Flash (D10) - Clef Omni (D11) | 1,273 | -6.52 | -8.96 to -4.08 | <.001 | <.001 |
| Solar Decide Flash (D10) - Drex v1.5 (D12) | 1,273 | -0.63 | -3.30 to 2.04 | .687 | 1.000 |
| Solar Decide Flash (D10) - Microsoft-Decision-1 (D13) | 1,273 | -6.68 | -9.19 to -4.16 | <.001 | <.001 |
| Clef Omni (D11) - Drex v1.5 (D12) | 1,273 | 5.89 | 3.30 to 8.48 | <.001 | <.001 |
| Clef Omni (D11) - Microsoft-Decision-1 (D13) | 1,273 | -0.16 | -2.52 to 2.28 | .949 | 1.000 |
| Drex v1.5 (D12) - Microsoft-Decision-1 (D13) | 1,273 | -6.05 | -8.48 to -3.69 | <.001 | <.001 |

**Supplementary Table S2. Paired contingency counts and supplementary Newcombe intervals.** Each row compares models A and B on the same questions. The four counts distinguish questions answered correctly by both models, by A alone, by B alone or by neither model, and together sum to the matched denominator n. The two counts in which only one model was correct enter the two-sided exact McNemar

test. The table includes all 273 primary comparisons in family S1 and the three exploratory native-effort comparisons in family S2. Supplementary pointwise 95% confidence intervals for the paired accuracy difference were calculated using Newcombe method 10, in percentage points and with the same A-B direction as Tables S1 and S3. Test conclusions were based on the family-adjusted P values reported in those tables.

| Family | Model A - Model B | n | Both correct | A only correct | B only correct | Both incorrect | Newcombe 95% CI, pp |
|---|---|---|---|---|---|---|---|
| S1 | JEV 1.13 (D01) - GPT-6.1 Sol Low (G01) | 1,273 | 1115 | 9 | 118 | 31 | -10.32 to -6.94 |
| S1 | JEV 1.13 (D01) - GPT-6 Luna Low (G02) | 1,273 | 1097 | 27 | 102 | 47 | -7.67 to -4.19 |
| S1 | JEV 1.13 (D01) - GPT-6 Luna High (G03) | 1,273 | 1103 | 21 | 109 | 40 | -8.70 to -5.22 |
| S1 | JEV 1.13 (D01) - Claude Sonnet 5.5 Low (G04) | 1,273 | 1105 | 19 | 106 | 43 | -8.59 to -5.18 |
| S1 | JEV 1.13 (D01) - Gemini 3.8 Flash Low (G05) | 1,273 | 1114 | 10 | 118 | 31 | -10.25 to -6.85 |
| S1 | JEV 1.13 (D01) - DeepSeek | 1,273 | 1107 | 17 | 109 | 40 | -8.98 to -5.58 |

| Family | Model A - Model B | n | Both correct | A only correct | B only correct | Both incorrect | Newcombe 95% CI, pp |
|---|---|---|---|---|---|---|---|
| | V4.1 Flash Low (G06) | | | | | | |
| S1 | JEV 1.13 (D01) - DeepSeek V4.1 Flash High (G07) | 1,273 | 1101 | 23 | 103 | 46 | -8.04 to -4.61 |
| S1 | JEV 1.13 (D01) - Qwen3.8 Max Low (G08) | 1,273 | 1068 | 56 | 73 | 76 | -3.11 to 0.42 |
| S1 | JEV 1.13 (D01) - Qwen3.8-27B Low (G09) | 1,273 | 1094 | 30 | 104 | 45 | -7.62 to -4.07 |
| S1 | JEV 1.13 (D01) - Qwen3.8-27B Xhigh (G10) | 1,273 | 1106 | 18 | 109 | 40 | -8.91 to -5.49 |
| S1 | JEV 1.13 (D01) - Mistral Medium 3.5 None (G11) | 1,273 | 956 | 168 | 41 | 108 | 7.83 to 12.16 |
| S1 | JEV 1.13 (D01) - GLM-5 Native (G12) | 1,273 | 1101 | 23 | 108 | 41 | -8.47 to -4.98 |

| Family | Model A - Model B | n | Both correct | A only correct | B only correct | Both incorrect | Newcombe 95% CI, pp |
|---|---|---|---|---|---|---|---|
| S1 | JEV 1.13 (D01) - Kimi K2.5 Native (G13) | 1,273 | 1101 | 23 | 102 | 47 | -7.96 to -4.54 |
| S1 | JEV 1.13 (D01) - Llama 4 Maverick Standard (G14) | 1,273 | 1011 | 113 | 56 | 93 | 2.49 to 6.49 |
| S1 | JEV 1.13 (D01) - Kimi K2.5 Instant (G15) | 1,273 | 1021 | 103 | 61 | 88 | 1.33 to 5.29 |
| S1 | GPT-6 Luna Decisions (D02) - GPT-6.1 Sol Low (G01) | 1,273 | 1124 | 10 | 109 | 30 | -9.49 to -6.20 |
| S1 | GPT-6 Luna Decisions (D02) - GPT-6 Luna Low (G02) | 1,273 | 1116 | 18 | 83 | 56 | -6.70 to -3.60 |
| S1 | GPT-6 Luna Decisions (D02) - GPT-6 | 1,273 | 1121 | 13 | 91 | 48 | -7.74 to -4.62 |

| Family | Model A - Model B | n | Both correct | A only correct | B only correct | Both incorrect | Newcombe 95% CI, pp |
|---|---|---|---|---|---|---|---|
| | Luna High (G03) | | | | | | |
| S1 | GPT-6 Luna Decisions (D02) - Claude Sonnet 5.5 Low (G04) | 1,273 | 1117 | 17 | 94 | 45 | -7.71 to -4.49 |
| S1 | GPT-6 Luna Decisions (D02) - Gemini 3.8 Flash Low (G05) | 1,273 | 1121 | 13 | 111 | 28 | -9.44 to -6.08 |
| S1 | GPT-6 Luna Decisions (D02) - DeepSeek V4.1 Flash Low (G06) | 1,273 | 1112 | 22 | 104 | 35 | -8.20 to -4.77 |
| S1 | GPT-6 Luna Decisions (D02) - DeepSeek V4.1 Flash High (G07) | 1,273 | 1105 | 29 | 99 | 40 | -7.27 to -3.80 |
| S1 | GPT-6 Luna Decisions (D02) - Qwen3.8 | 1,273 | 1075 | 59 | 66 | 73 | -2.29 to 1.19 |

| Family | Model A - Model B | n | Both correct | A only correct | B only correct | Both incorrect | Newcombe 95% CI, pp |
|---|---|---|---|---|---|---|---|
| | Max Low (G08) | | | | | | |
| S1 | GPT-6 Luna Decisions (D02) - Qwen3.8-27B Low (G09) | 1,273 | 1106 | 28 | 92 | 47 | -6.75 to -3.38 |
| S1 | GPT-6 Luna Decisions (D02) - Qwen3.8-27B Xhigh (G10) | 1,273 | 1114 | 20 | 101 | 38 | -8.09 to -4.73 |
| S1 | GPT-6 Luna Decisions (D02) - Mistral Medium 3.5 None (G11) | 1,273 | 960 | 174 | 37 | 102 | 8.61 to 12.95 |
| S1 | GPT-6 Luna Decisions (D02) - GLM-5 Native (G12) | 1,273 | 1107 | 27 | 102 | 37 | -7.67 to -4.19 |
| S1 | GPT-6 Luna Decisions (D02) - Kimi K2.5 | 1,273 | 1104 | 30 | 99 | 40 | -7.20 to -3.71 |

| Family | Model A - Model B | n | Both correct | A only correct | B only correct | Both incorrect | Newcombe 95% CI, pp |
|---|---|---|---|---|---|---|---|
| | Native (G13) | | | | | | |
| S1 | GPT-6 Luna Decisions (D02) - Llama 4 Maverick Standard (G14) | 1,273 | 1013 | 121 | 54 | 85 | 3.25 to 7.31 |
| S1 | GPT-6 Luna Decisions (D02) - Kimi K2.5 Instant (G15) | 1,273 | 1028 | 106 | 54 | 85 | 2.15 to 6.05 |
| S1 | Kev 4B (D03) - GPT-6.1 Sol Low (G01) | 1,273 | 849 | 13 | 384 | 27 | -31.78 to -26.53 |
| S1 | Kev 4B (D03) - GPT-6 Luna Low (G02) | 1,273 | 839 | 23 | 360 | 51 | -29.13 to -23.83 |
| S1 | Kev 4B (D03) - GPT-6 Luna High (G03) | 1,273 | 841 | 21 | 371 | 40 | -30.16 to -24.85 |
| S1 | Kev 4B (D03) - Claude | 1,273 | 841 | 21 | 370 | 41 | -30.08 to -24.77 |

| Family | Model A - Model B | n | Both correct | A only correct | B only correct | Both incorrect | Newcombe 95% CI, pp |
|---|---|---|---|---|---|---|---|
| | Sonnet 5.5 Low (G04) | | | | | | |
| S1 | Kev 4B (D03) - Gemini 3.8 Flash Low (G05) | 1,273 | 853 | 9 | 379 | 32 | -31.67 to -26.50 |
| S1 | Kev 4B (D03) - DeepSeek V4.1 Flash Low (G06) | 1,273 | 845 | 17 | 371 | 40 | -30.45 to -25.19 |
| S1 | Kev 4B (D03) - DeepSeek V4.1 Flash High (G07) | 1,273 | 840 | 22 | 364 | 47 | -29.52 to -24.22 |
| S1 | Kev 4B (D03) - Qwen3.8 Max Low (G08) | 1,273 | 822 | 40 | 319 | 92 | -24.58 to -19.25 |
| S1 | Kev 4B (D03) - Qwen3.8-27B Low (G09) | 1,273 | 843 | 19 | 355 | 56 | -29.01 to -23.80 |
| S1 | Kev 4B (D03) - Qwen3.8-27B Xhigh (G10) | 1,273 | 847 | 15 | 368 | 43 | -30.35 to -25.14 |

| Family | Model A - Model B | n | Both correct | A only correct | B only correct | Both incorrect | Newcombe 95% CI, pp |
|---|---|---|---|---|---|---|---|
| S1 | Kev 4B (D03) - Mistral Medium 3.5 None (G11) | 1,273 | 789 | 73 | 208 | 203 | -13.12 to -8.08 |
| S1 | Kev 4B (D03) - GLM-5 Native (G12) | 1,273 | 847 | 15 | 362 | 49 | -29.87 to -24.68 |
| S1 | Kev 4B (D03) - Kimi K2.5 Native (G13) | 1,273 | 840 | 22 | 363 | 48 | -29.44 to -24.15 |
| S1 | Kev 4B (D03) - Llama 4 Maverick Standard (G14) | 1,273 | 804 | 58 | 263 | 148 | -18.72 to -13.48 |
| S1 | Kev 4B (D03) - Kimi K2.5 Instant (G15) | 1,273 | 814 | 48 | 268 | 143 | -19.85 to -14.71 |
| S1 | Solar Decide (D04) - GPT-6.1 Sol Low (G01) | 1,273 | 1044 | 8 | 189 | 32 | -16.31 to -12.25 |
| S1 | Solar Decide | 1,273 | 1029 | 23 | 170 | 51 | -13.65 to -9.53 |

| Family | Model A - Model B | n | Both correct | A only correct | B only correct | Both incorrect | Newcombe 95% CI, pp |
|---|---|---|---|---|---|---|---|
| | (D04) - GPT-6 Luna Low (G02) | | | | | | |
| S1 | Solar Decide (D04) - GPT-6 Luna High (G03) | 1,273 | 1031 | 21 | 181 | 40 | -14.70 to -10.52 |
| S1 | Solar Decide (D04) - Claude Sonnet 5.5 Low (G04) | 1,273 | 1036 | 16 | 175 | 46 | -14.57 to -10.51 |
| S1 | Solar Decide (D04) - Gemini 3.8 Flash Low (G05) | 1,273 | 1041 | 11 | 191 | 30 | -16.26 to -12.14 |
| S1 | Solar Decide (D04) - DeepSeek V4.1 Flash Low (G06) | 1,273 | 1041 | 11 | 175 | 46 | -14.93 to -10.95 |
| S1 | Solar Decide (D04) - DeepSeek V4.1 Flash High (G07) | 1,273 | 1035 | 17 | 169 | 52 | -14.00 to -9.98 |

| Family | Model A - Model B | n | Both correct | A only correct | B only correct | Both incorrect | Newcombe 95% CI, pp |
|---|---|---|---|---|---|---|---|
| S1 | Solar Decide (D04) - Qwen3.8 Max Low (G08) | 1,273 | 997 | 55 | 144 | 77 | -9.16 to -4.86 |
| S1 | Solar Decide (D04) - Qwen3.8-27B Low (G09) | 1,273 | 1032 | 20 | 166 | 55 | -13.53 to -9.49 |
| S1 | Solar Decide (D04) - Qwen3.8-27B Xhigh (G10) | 1,273 | 1040 | 12 | 175 | 46 | -14.86 to -10.86 |
| S1 | Solar Decide (D04) - Mistral Medium 3.5 None (G11) | 1,273 | 927 | 125 | 70 | 151 | 2.18 to 6.47 |
| S1 | Solar Decide (D04) - GLM-5 Native (G12) | 1,273 | 1034 | 18 | 175 | 46 | -14.42 to -10.34 |
| S1 | Solar Decide (D04) - Kimi K2.5 | 1,273 | 1032 | 20 | 171 | 50 | -13.95 to -9.86 |

| Family | Model A - Model B | n | Both correct | A only correct | B only correct | Both incorrect | Newcombe 95% CI, pp |
|---|---|---|---|---|---|---|---|
| | Native (G13) | | | | | | |
| S1 | Solar Decide (D04) - Llama 4 Maverick Standard (G14) | 1,273 | 957 | 95 | 110 | 111 | -3.39 to 1.03 |
| S1 | Solar Decide (D04) - Kimi K2.5 Instant (G15) | 1,273 | 964 | 88 | 118 | 103 | -4.58 to -0.14 |
| S1 | Mercury Decide (D05) - GPT-6.1 Sol Low (G01) | 1,273 | 1069 | 16 | 164 | 24 | -13.66 to -9.70 |
| S1 | Mercury Decide (D05) - GPT-6 Luna Low (G02) | 1,273 | 1055 | 30 | 144 | 44 | -10.98 to -7.00 |
| S1 | Mercury Decide (D05) - GPT-6 Luna High (G03) | 1,273 | 1061 | 24 | 151 | 37 | -12.00 to -8.04 |
| S1 | Mercury Decide | 1,273 | 1064 | 21 | 147 | 41 | -11.88 to -8.01 |

| Family | Model A - Model B | n | Both correct | A only correct | B only correct | Both incorrect | Newcombe 95% CI, pp |
|---|---|---|---|---|---|---|---|
| | (D05) - Claude Sonnet 5.5 Low (G04) | | | | | | |
| S1 | Mercury Decide (D05) - Gemini 3.8 Flash Low (G05) | 1,273 | 1070 | 15 | 162 | 26 | -13.57 to -9.64 |
| S1 | Mercury Decide (D05) - DeepSeek V4.1 Flash Low (G06) | 1,273 | 1068 | 17 | 148 | 40 | -12.25 to -8.43 |
| S1 | Mercury Decide (D05) - DeepSeek V4.1 Flash High (G07) | 1,273 | 1061 | 24 | 143 | 45 | -11.33 to -7.45 |
| S1 | Mercury Decide (D05) - Qwen3.8 Max Low (G08) | 1,273 | 1033 | 52 | 108 | 80 | -6.36 to -2.47 |
| S1 | Mercury Decide (D05) - Qwen3.8-27B Low (G09) | 1,273 | 1068 | 17 | 130 | 58 | -10.74 to -7.10 |

| Family | Model A - Model B | n | Both correct | A only correct | B only correct | Both incorrect | Newcombe 95% CI, pp |
|---|---|---|---|---|---|---|---|
| S1 | Mercury Decide (D05) - Qwen3.8-27B Xhigh (G10) | 1,273 | 1074 | 11 | 141 | 47 | -12.10 to -8.44 |
| S1 | Mercury Decide (D05) - Mistral Medium 3.5 None (G11) | 1,273 | 940 | 145 | 57 | 131 | 4.76 to 9.09 |
| S1 | Mercury Decide (D05) - GLM-5 Native (G12) | 1,273 | 1063 | 22 | 146 | 42 | -11.73 to -7.85 |
| S1 | Mercury Decide (D05) - Kimi K2.5 Native (G13) | 1,273 | 1062 | 23 | 141 | 47 | -11.23 to -7.39 |
| S1 | Mercury Decide (D05) - Llama 4 Maverick Standard (G14) | 1,273 | 989 | 96 | 78 | 110 | -0.63 to 3.46 |
| S1 | Mercury Decide (D05) - Kimi K2.5 | 1,273 | 994 | 91 | 88 | 100 | -1.84 to 2.31 |

| Family | Model A - Model B | n | Both correct | A only correct | B only correct | Both incorrect | Newcombe 95% CI, pp |
|---|---|---|---|---|---|---|---|
| | Instant (G15) | | | | | | |
| S1 | Clef (D06) - GPT-6.1 Sol Low (G01) | 1,273 | 1066 | 13 | 167 | 27 | -14.13 to -10.18 |
| S1 | Clef (D06) - GPT-6 Luna Low (G02) | 1,273 | 1055 | 24 | 144 | 50 | -11.41 to -7.52 |
| S1 | Clef (D06) - GPT-6 Luna High (G03) | 1,273 | 1058 | 21 | 154 | 40 | -12.46 to -8.52 |
| S1 | Clef (D06) - Claude Sonnet 5.5 Low (G04) | 1,273 | 1056 | 23 | 155 | 39 | -12.40 to -8.42 |
| S1 | Clef (D06) - Gemini 3.8 Flash Low (G05) | 1,273 | 1067 | 12 | 165 | 29 | -14.03 to -10.12 |
| S1 | Clef (D06) - DeepSeek V4.1 Flash Low (G06) | 1,273 | 1059 | 20 | 157 | 37 | -12.79 to -8.83 |
| S1 | Clef (D06) - DeepSeek V4.1 Flash High (G07) | 1,273 | 1052 | 27 | 152 | 42 | -11.86 to -7.85 |

| Family | Model A - Model B | n | Both correct | A only correct | B only correct | Both incorrect | Newcombe 95% CI, pp |
|---|---|---|---|---|---|---|---|
| S1 | Clef (D06) - Qwen3.8 Max Low (G08) | 1,273 | 1028 | 51 | 113 | 81 | -6.86 to -2.92 |
| S1 | Clef (D06) - Qwen3.8-27B Low (G09) | 1,273 | 1061 | 18 | 137 | 57 | -11.26 to -7.53 |
| S1 | Clef (D06) - Qwen3.8-27B Xhigh (G10) | 1,273 | 1067 | 12 | 148 | 46 | -12.61 to -8.87 |
| S1 | Clef (D06) - Mistral Medium 3.5 None (G11) | 1,273 | 931 | 148 | 66 | 128 | 4.21 to 8.68 |
| S1 | Clef (D06) - GLM-5 Native (G12) | 1,273 | 1058 | 21 | 151 | 43 | -12.21 to -8.30 |
| S1 | Clef (D06) - Kimi K2.5 Native (G13) | 1,273 | 1053 | 26 | 150 | 44 | -11.77 to -7.79 |
| S1 | Clef (D06) - Llama 4 Maverick Standard (G14) | 1,273 | 985 | 94 | 82 | 112 | -1.11 to 3.00 |
| S1 | Clef (D06) - Kimi K2.5 | 1,273 | 992 | 87 | 90 | 104 | -2.30 to 1.83 |

| Family | Model A - Model B | n | Both correct | A only correct | B only correct | Both incorrect | Newcombe 95% CI, pp |
|---|---|---|---|---|---|---|---|
| | Instant (G15) | | | | | | |
| S1 | Clef Flash (D07) - GPT-6.1 Sol Low (G01) | 1,273 | 923 | 15 | 310 | 25 | -25.68 to -20.73 |
| S1 | Clef Flash (D07) - GPT-6 Luna Low (G02) | 1,273 | 912 | 26 | 287 | 48 | -23.01 to -18.03 |
| S1 | Clef Flash (D07) - GPT-6 Luna High (G03) | 1,273 | 916 | 22 | 296 | 39 | -24.03 to -19.06 |
| S1 | Clef Flash (D07) - Claude Sonnet 5.5 Low (G04) | 1,273 | 917 | 21 | 294 | 41 | -23.94 to -18.99 |
| S1 | Clef Flash (D07) - Gemini 3.8 Flash Low (G05) | 1,273 | 925 | 13 | 307 | 28 | -25.58 to -20.67 |
| S1 | Clef Flash (D07) - DeepSeek V4.1 Flash Low (G06) | 1,273 | 919 | 19 | 297 | 38 | -24.33 to -19.39 |

| Family | Model A - Model B | n | Both correct | A only correct | B only correct | Both incorrect | Newcombe 95% CI, pp |
|---|---|---|---|---|---|---|---|
| S1 | Clef Flash (D07) - DeepSeek V4.1 Flash High (G07) | 1,273 | 912 | 26 | 292 | 43 | -23.42 to -18.41 |
| S1 | Clef Flash (D07) - Qwen3.8 Max Low (G08) | 1,273 | 894 | 44 | 247 | 88 | -18.44 to -13.47 |
| S1 | Clef Flash (D07) - Qwen3.8-27B Low (G09) | 1,273 | 913 | 25 | 285 | 50 | -22.92 to -17.97 |
| S1 | Clef Flash (D07) - Qwen3.8-27B Xhigh (G10) | 1,273 | 921 | 17 | 294 | 41 | -24.24 to -19.34 |
| S1 | Clef Flash (D07) - Mistral Medium 3.5 None (G11) | 1,273 | 851 | 87 | 146 | 189 | -6.98 to -2.29 |
| S1 | Clef Flash (D07) - GLM-5 Native (G12) | 1,273 | 922 | 16 | 287 | 48 | -23.74 to -18.90 |
| S1 | Clef Flash (D07) - Kimi K2.5 | 1,273 | 916 | 22 | 287 | 48 | -23.30 to -18.38 |

| Family | Model A - Model B | n | Both correct | A only correct | B only correct | Both incorrect | Newcombe 95% CI, pp |
|---|---|---|---|---|---|---|---|
| | Native (G13) | | | | | | |
| S1 | Clef Flash (D07) - Llama 4 Maverick Standard (G14) | 1,273 | 868 | 70 | 199 | 136 | -12.61 to -7.67 |
| S1 | Clef Flash (D07) - Kimi K2.5 Instant (G15) | 1,273 | 877 | 61 | 205 | 130 | -13.76 to -8.88 |
| S1 | Liquid d1 (D08) - GPT-6.1 Sol Low (G01) | 1,273 | 1111 | 14 | 122 | 26 | -10.30 to -6.79 |
| S1 | Liquid d1 (D08) - GPT-6 Luna Low (G02) | 1,273 | 1092 | 33 | 107 | 41 | -7.66 to -4.03 |
| S1 | Liquid d1 (D08) - GPT-6 Luna High (G03) | 1,273 | 1099 | 26 | 113 | 35 | -8.67 to -5.08 |
| S1 | Liquid d1 (D08) - Claude Sonnet 5.5 Low (G04) | 1,273 | 1099 | 26 | 112 | 36 | -8.59 to -5.01 |

| Family | Model A - Model B | n | Both correct | A only correct | B only correct | Both incorrect | Newcombe 95% CI, pp |
|---|---|---|---|---|---|---|---|
| S1 | Liquid d1 (D08) - Gemini 3.8 Flash Low (G05) | 1,273 | 1111 | 14 | 121 | 27 | -10.22 to -6.72 |
| S1 | Liquid d1 (D08) - DeepSeek V4.1 Flash Low (G06) | 1,273 | 1107 | 18 | 109 | 39 | -8.91 to -5.49 |
| S1 | Liquid d1 (D08) - DeepSeek V4.1 Flash High (G07) | 1,273 | 1100 | 25 | 104 | 44 | -7.98 to -4.51 |
| S1 | Liquid d1 (D08) - Qwen3.8 Max Low (G08) | 1,273 | 1073 | 52 | 68 | 80 | -2.97 to 0.44 |
| S1 | Liquid d1 (D08) - Qwen3.8-27B Low (G09) | 1,273 | 1102 | 23 | 96 | 52 | -7.45 to -4.10 |
| S1 | Liquid d1 (D08) - Qwen3.8-27B Xhigh (G10) | 1,273 | 1108 | 17 | 107 | 41 | -8.81 to -5.43 |
| S1 | Liquid d1 (D08) - Mistral Medium | 1,273 | 959 | 166 | 38 | 110 | 7.93 to 12.21 |

| Family | Model A - Model B | n | Both correct | A only correct | B only correct | Both incorrect | Newcombe 95% CI, pp |
|---|---|---|---|---|---|---|---|
| | 3.5 None (G11) | | | | | | |
| S1 | Liquid d1 (D08) - GLM-5 Native (G12) | 1,273 | 1102 | 23 | 107 | 41 | -8.38 to -4.90 |
| S1 | Liquid d1 (D08) - Kimi K2.5 Native (G13) | 1,273 | 1101 | 24 | 102 | 46 | -7.89 to -4.45 |
| S1 | Liquid d1 (D08) - Llama 4 Maverick Standard (G14) | 1,273 | 1016 | 109 | 51 | 97 | 2.62 to 6.52 |
| S1 | Liquid d1 (D08) - Kimi K2.5 Instant (G15) | 1,273 | 1025 | 100 | 57 | 91 | 1.45 to 5.33 |
| S1 | Decider V1.1 27B (D09) - GPT-6.1 Sol Low (G01) | 1,273 | 1104 | 14 | 129 | 26 | -10.89 to -7.30 |
| S1 | Decider V1.1 27B (D09) - GPT-6 Luna Low (G02) | 1,273 | 1088 | 30 | 111 | 44 | -8.21 to -4.58 |

| Family | Model A - Model B | n | Both correct | A only correct | B only correct | Both incorrect | Newcombe 95% CI, pp |
|---|---|---|---|---|---|---|---|
| S1 | Decider V1.1 27B (D09) - GPT-6 Luna High (G03) | 1,273 | 1091 | 27 | 121 | 34 | -9.27 to -5.58 |
| S1 | Decider V1.1 27B (D09) - Claude Sonnet 5.5 Low (G04) | 1,273 | 1091 | 27 | 120 | 35 | -9.19 to -5.50 |
| S1 | Decider V1.1 27B (D09) - Gemini 3.8 Flash Low (G05) | 1,273 | 1106 | 12 | 126 | 29 | -10.78 to -7.26 |
| S1 | Decider V1.1 27B (D09) - DeepSeek V4.1 Flash Low (G06) | 1,273 | 1097 | 21 | 119 | 36 | -9.54 to -5.95 |
| S1 | Decider V1.1 27B (D09) - DeepSeek V4.1 Flash High (G07) | 1,273 | 1094 | 24 | 110 | 45 | -8.56 to -5.03 |
| S1 | Decider V1.1 27B (D09) - Qwen3.8 | 1,273 | 1053 | 65 | 88 | 67 | -3.73 to 0.11 |

| Family | Model A - Model B | n | Both correct | A only correct | B only correct | Both incorrect | Newcombe 95% CI, pp |
|---|---|---|---|---|---|---|---|
| | Max Low (G08) | | | | | | |
| S1 | Decider V1.1 27B (D09) - Qwen3.8-27B Low (G09) | 1,273 | 1089 | 29 | 109 | 46 | -8.12 to -4.53 |
| S1 | Decider V1.1 27B (D09) - Qwen3.8-27B Xhigh (G10) | 1,273 | 1104 | 14 | 111 | 44 | -9.37 to -5.98 |
| S1 | Decider V1.1 27B (D09) - Mistral Medium 3.5 None (G11) | 1,273 | 945 | 173 | 52 | 103 | 7.26 to 11.78 |
| S1 | Decider V1.1 27B (D09) - GLM-5 Native (G12) | 1,273 | 1096 | 22 | 113 | 42 | -8.96 to -5.43 |
| S1 | Decider V1.1 27B (D09) - Kimi K2.5 Native (G13) | 1,273 | 1090 | 28 | 113 | 42 | -8.53 to -4.90 |
| S1 | Decider V1.1 27B | 1,273 | 1005 | 113 | 62 | 93 | 1.98 to 6.06 |

| Family | Model A - Model B | n | Both correct | A only correct | B only correct | Both incorrect | Newcombe 95% CI, pp |
|---|---|---|---|---|---|---|---|
| | (D09) - Llama 4 Maverick Standard (G14) | | | | | | |
| S1 | Decider V1.1 27B (D09) - Kimi K2.5 Instant (G15) | 1,273 | 1008 | 110 | 74 | 81 | 0.74 to 4.93 |
| S1 | Solar Decide Flash (D10) - GPT-6.1 Sol Low (G01) | 1,273 | 917 | 11 | 316 | 29 | -26.46 to -21.53 |
| S1 | Solar Decide Flash (D10) - GPT-6 Luna Low (G02) | 1,273 | 904 | 24 | 295 | 50 | -23.81 to -18.81 |
| S1 | Solar Decide Flash (D10) - GPT-6 Luna High (G03) | 1,273 | 907 | 21 | 305 | 40 | -24.84 to -19.83 |
| S1 | Solar Decide Flash (D10) - | 1,273 | 910 | 18 | 301 | 44 | -24.73 to -19.78 |

| Family | Model A - Model B | n | Both correct | A only correct | B only correct | Both incorrect | Newcombe 95% CI, pp |
|---|---|---|---|---|---|---|---|
| | Claude Sonnet 5.5 Low (G04) | | | | | | |
| S1 | Solar Decide Flash (D10) - Gemini 3.8 Flash Low (G05) | 1,273 | 914 | 14 | 318 | 27 | -26.40 to -21.42 |
| S1 | Solar Decide Flash (D10) - DeepSeek V4.1 Flash Low (G06) | 1,273 | 916 | 12 | 300 | 45 | -25.08 to -20.23 |
| S1 | Solar Decide Flash (D10) - DeepSeek V4.1 Flash High (G07) | 1,273 | 910 | 18 | 294 | 51 | -24.16 to -19.25 |
| S1 | Solar Decide Flash (D10) - Qwen3.8 Max Low (G08) | 1,273 | 884 | 44 | 257 | 88 | -19.26 to -14.22 |
| S1 | Solar Decide Flash (D10) - | 1,273 | 908 | 20 | 290 | 55 | -23.69 to -18.78 |

| Family | Model A - Model B | n | Both correct | A only correct | B only correct | Both incorrect | Newcombe 95% CI, pp |
|---|---|---|---|---|---|---|---|
| | Qwen3.8-27B Low (G09) | | | | | | |
| S1 | Solar Decide Flash (D10) - Qwen3.8-27B Xhigh (G10) | 1,273 | 911 | 17 | 304 | 41 | -25.05 to -20.10 |
| S1 | Solar Decide Flash (D10) - Mistral Medium 3.5 None (G11) | 1,273 | 849 | 79 | 148 | 197 | -7.73 to -3.11 |
| S1 | Solar Decide Flash (D10) - GLM-5 Native (G12) | 1,273 | 912 | 16 | 297 | 48 | -24.55 to -19.65 |
| S1 | Solar Decide Flash (D10) - Kimi K2.5 Native (G13) | 1,273 | 910 | 18 | 293 | 52 | -24.08 to -19.18 |
| S1 | Solar Decide Flash (D10) - | 1,273 | 862 | 66 | 205 | 140 | -13.39 to -8.45 |

| Family | Model A - Model B | n | Both correct | A only correct | B only correct | Both incorrect | Newcombe 95% CI, pp |
|---|---|---|---|---|---|---|---|
| | Llama 4 Maverick Standard (G14) | | | | | | |
| S1 | Solar Decide Flash (D10) - Kimi K2.5 Instant (G15) | 1,273 | 868 | 60 | 214 | 131 | -14.57 to -9.63 |
| S1 | Clef Omni (D11) - GPT-6.1 Sol Low (G01) | 1,273 | 998 | 13 | 235 | 27 | -19.73 to -15.25 |
| S1 | Clef Omni (D11) - GPT-6 Luna Low (G02) | 1,273 | 986 | 25 | 213 | 49 | -17.05 to -12.56 |
| S1 | Clef Omni (D11) - GPT-6 Luna High (G03) | 1,273 | 984 | 27 | 228 | 34 | -18.13 to -13.51 |
| S1 | Clef Omni (D11) - Claude Sonnet 5.5 Low (G04) | 1,273 | 988 | 23 | 223 | 39 | -18.01 to -13.48 |
| S1 | Clef Omni (D11) - Gemini | 1,273 | 999 | 12 | 233 | 29 | -19.64 to -15.18 |

| Family | Model A - Model B | n | Both correct | A only correct | B only correct | Both incorrect | Newcombe 95% CI, pp |
|---|---|---|---|---|---|---|---|
| | 3.8 Flash Low (G05) | | | | | | |
| S1 | Clef Omni (D11) - DeepSeek V4.1 Flash Low (G06) | 1,273 | 989 | 22 | 227 | 35 | -18.42 to -13.87 |
| S1 | Clef Omni (D11) - DeepSeek V4.1 Flash High (G07) | 1,273 | 982 | 29 | 222 | 40 | -17.50 to -12.89 |
| S1 | Clef Omni (D11) - Qwen3.8 Max Low (G08) | 1,273 | 966 | 45 | 175 | 87 | -12.46 to -8.01 |
| S1 | Clef Omni (D11) - Qwen3.8-27B Low (G09) | 1,273 | 988 | 23 | 210 | 52 | -16.94 to -12.50 |
| S1 | Clef Omni (D11) - Qwen3.8-27B Xhigh (G10) | 1,273 | 993 | 18 | 222 | 40 | -18.30 to -13.84 |
| S1 | Clef Omni (D11) - Mistral Medium 3.5 None (G11) | 1,273 | 901 | 110 | 96 | 166 | -1.12 to 3.32 |

| Family | Model A - Model B | n | Both correct | A only correct | B only correct | Both incorrect | Newcombe 95% CI, pp |
|---|---|---|---|---|---|---|---|
| S1 | Clef Omni (D11) - GLM-5 Native (G12) | 1,273 | 989 | 22 | 220 | 42 | -17.84 to -13.34 |
| S1 | Clef Omni (D11) - Kimi K2.5 Native (G13) | 1,273 | 989 | 22 | 214 | 48 | -17.35 to -12.89 |
| S1 | Clef Omni (D11) - Llama 4 Maverick Standard (G14) | 1,273 | 928 | 83 | 139 | 123 | -6.69 to -2.11 |
| S1 | Clef Omni (D11) - Kimi K2.5 Instant (G15) | 1,273 | 945 | 66 | 137 | 125 | -7.77 to -3.40 |
| S1 | Drex v1.5 (D12) - GPT-6.1 Sol Low (G01) | 1,273 | 920 | 16 | 313 | 24 | -25.85 to -20.87 |
| S1 | Drex v1.5 (D12) - GPT-6 Luna Low (G02) | 1,273 | 904 | 32 | 295 | 42 | -23.23 to -18.12 |
| S1 | Drex v1.5 (D12) - GPT-6 | 1,273 | 911 | 25 | 301 | 36 | -24.22 to -19.18 |

| Family | Model A - Model B | n | Both correct | A only correct | B only correct | Both incorrect | Newcombe 95% CI, pp |
|---|---|---|---|---|---|---|---|
| | Luna High (G03) | | | | | | |
| S1 | Drex v1.5 (D12) - Claude Sonnet 5.5 Low (G04) | 1,273 | 914 | 22 | 297 | 40 | -24.12 to -19.14 |
| S1 | Drex v1.5 (D12) - Gemini 3.8 Flash Low (G05) | 1,273 | 921 | 15 | 311 | 26 | -25.76 to -20.80 |
| S1 | Drex v1.5 (D12) - DeepSeek V4.1 Flash Low (G06) | 1,273 | 914 | 22 | 302 | 35 | -24.52 to -19.51 |
| S1 | Drex v1.5 (D12) - DeepSeek V4.1 Flash High (G07) | 1,273 | 906 | 30 | 298 | 39 | -23.62 to -18.52 |
| S1 | Drex v1.5 (D12) - Qwen3.8 Max Low (G08) | 1,273 | 898 | 38 | 243 | 94 | -18.55 to -13.68 |
| S1 | Drex v1.5 (D12) - Qwen3.8-27B Low (G09) | 1,273 | 913 | 23 | 285 | 52 | -23.07 to -18.14 |

| Family | Model A - Model B | n | Both correct | A only correct | B only correct | Both incorrect | Newcombe 95% CI, pp |
|---|---|---|---|---|---|---|---|
| S1 | Drex v1.5 (D12) - Qwen3.8-27B Xhigh (G10) | 1,273 | 916 | 20 | 299 | 38 | -24.42 to -19.46 |
| S1 | Drex v1.5 (D12) - Mistral Medium 3.5 None (G11) | 1,273 | 828 | 108 | 169 | 168 | -7.35 to -2.24 |
| S1 | Drex v1.5 (D12) - GLM-5 Native (G12) | 1,273 | 915 | 21 | 294 | 43 | -23.94 to -18.99 |
| S1 | Drex v1.5 (D12) - Kimi K2.5 Native (G13) | 1,273 | 913 | 23 | 290 | 47 | -23.47 to -18.52 |
| S1 | Drex v1.5 (D12) - Llama 4 Maverick Standard (G14) | 1,273 | 860 | 76 | 207 | 130 | -12.83 to -7.76 |
| S1 | Drex v1.5 (D12) - Kimi K2.5 Instant (G15) | 1,273 | 869 | 67 | 213 | 124 | -13.98 to -8.97 |
| S1 | Microsoft-Decision-1 (D13) - | 1,273 | 996 | 17 | 237 | 23 | -19.60 to -15.05 |

| Family | Model A - Model B | n | Both correct | A only correct | B only correct | Both incorrect | Newcombe 95% CI, pp |
|---|---|---|---|---|---|---|---|
| | GPT-6.1 Sol Low (G01) | | | | | | |
| S1 | Microsoft-Decision-1 (D13) - GPT-6 Luna Low (G02) | 1,273 | 982 | 31 | 217 | 43 | -16.94 to -12.34 |
| S1 | Microsoft-Decision-1 (D13) - GPT-6 Luna High (G03) | 1,273 | 986 | 27 | 226 | 34 | -17.97 to -13.36 |
| S1 | Microsoft-Decision-1 (D13) - Claude Sonnet 5.5 Low (G04) | 1,273 | 990 | 23 | 221 | 39 | -17.85 to -13.33 |
| S1 | Microsoft-Decision-1 (D13) - Gemini 3.8 Flash Low (G05) | 1,273 | 996 | 17 | 236 | 24 | -19.52 to -14.97 |
| S1 | Microsoft-Decision-1 (D13) - DeepSeek V4.1 Flash Low (G06) | 1,273 | 993 | 20 | 223 | 37 | -18.23 to -13.74 |

| Family | Model A - Model B | n | Both correct | A only correct | B only correct | Both incorrect | Newcombe 95% CI, pp |
|---|---|---|---|---|---|---|---|
| S1 | Microsoft-Decision-1 (D13) - DeepSeek V4.1 Flash High (G07) | 1,273 | 983 | 30 | 221 | 39 | -17.34 to -12.73 |
| S1 | Microsoft-Decision-1 (D13) - Qwen3.8 Max Low (G08) | 1,273 | 961 | 52 | 180 | 80 | -12.36 to -7.78 |
| S1 | Microsoft-Decision-1 (D13) - Qwen3.8-27B Low (G09) | 1,273 | 989 | 24 | 209 | 51 | -16.79 to -12.34 |
| S1 | Microsoft-Decision-1 (D13) - Qwen3.8-27B Xhigh (G10) | 1,273 | 994 | 19 | 221 | 39 | -18.14 to -13.68 |
| S1 | Microsoft-Decision-1 (D13) - Mistral Medium 3.5 None (G11) | 1,273 | 882 | 131 | 115 | 145 | -1.16 to 3.68 |
| S1 | Microsoft-Decision-1 (D13) - GLM-5 | 1,273 | 987 | 26 | 222 | 38 | -17.71 to -13.15 |

| Family | Model A - Model B | n | Both correct | A only correct | B only correct | Both incorrect | Newcombe 95% CI, pp |
|---|---|---|---|---|---|---|---|
| | Native (G12) | | | | | | |
| S1 | Microsoft-Decision-1 (D13) - Kimi K2.5 Native (G13) | 1,273 | 984 | 29 | 219 | 41 | -17.25 to -12.66 |
| S1 | Microsoft-Decision-1 (D13) - Llama 4 Maverick Standard (G14) | 1,273 | 926 | 87 | 141 | 119 | -6.57 to -1.92 |
| S1 | Microsoft-Decision-1 (D13) - Kimi K2.5 Instant (G15) | 1,273 | 933 | 80 | 149 | 111 | -7.75 to -3.11 |
| S1 | JEV 1.13 (D01) - GPT-6 Luna Decisions (D02) | 1,273 | 1064 | 60 | 70 | 79 | -2.56 to 0.98 |
| S1 | JEV 1.13 (D01) - Kev 4B (D03) | 1,273 | 823 | 301 | 39 | 110 | 17.97 to 23.19 |
| S1 | JEV 1.13 (D01) - Solar | 1,273 | 992 | 132 | 60 | 89 | 3.55 to 7.79 |

| Family | Model A - Model B | n | Both correct | A only correct | B only correct | Both incorrect | Newcombe 95% CI, pp |
|---|---|---|---|---|---|---|---|
| | Decide (D04) | | | | | | |
| S1 | JEV 1.13 (D01) - Mercury Decide (D05) | 1,273 | 1029 | 95 | 56 | 93 | 1.17 to 4.98 |
| S1 | JEV 1.13 (D01) - Clef (D06) | 1,273 | 1026 | 98 | 53 | 96 | 1.65 to 5.45 |
| S1 | JEV 1.13 (D01) - Clef Flash (D07) | 1,273 | 895 | 229 | 43 | 106 | 12.20 to 17.04 |
| S1 | JEV 1.13 (D01) - Liquid d1 (D08) | 1,273 | 1069 | 55 | 56 | 93 | -1.72 to 1.56 |
| S1 | JEV 1.13 (D01) - Decider V1.1 27B (D09) | 1,273 | 1059 | 65 | 59 | 90 | -1.26 to 2.21 |
| S1 | JEV 1.13 (D01) - Solar Decide Flash (D10) | 1,273 | 877 | 247 | 51 | 98 | 12.88 to 17.93 |
| S1 | JEV 1.13 (D01) - Clef Omni (D11) | 1,273 | 961 | 163 | 50 | 99 | 6.69 to 11.10 |

| Family | Model A - Model B | n | Both correct | A only correct | B only correct | Both incorrect | Newcombe 95% CI, pp |
|---|---|---|---|---|---|---|---|
| S1 | JEV 1.13 (D01) - Drex v1.5 (D12) | 1,273 | 882 | 242 | 54 | 95 | 12.24 to 17.30 |
| S1 | JEV 1.13 (D01) - Microsoft-Decision-1 (D13) | 1,273 | 956 | 168 | 57 | 92 | 6.46 to 11.00 |
| S1 | GPT-6 Luna Decisions (D02) - Kev 4B (D03) | 1,273 | 818 | 316 | 44 | 95 | 18.68 to 24.04 |
| S1 | GPT-6 Luna Decisions (D02) - Solar Decide (D04) | 1,273 | 996 | 138 | 56 | 83 | 4.33 to 8.58 |
| S1 | GPT-6 Luna Decisions (D02) - Mercury Decide (D05) | 1,273 | 1030 | 104 | 55 | 84 | 1.92 to 5.81 |
| S1 | GPT-6 Luna Decisions (D02) - Clef (D06) | 1,273 | 1032 | 102 | 47 | 92 | 2.46 to 6.22 |

| Family | Model A - Model B | n | Both correct | A only correct | B only correct | Both incorrect | Newcombe 95% CI, pp |
|---|---|---|---|---|---|---|---|
| S1 | GPT-6 Luna Decisions (D02) - Clef Flash (D07) | 1,273 | 889 | 245 | 49 | 90 | 12.90 to 17.91 |
| S1 | GPT-6 Luna Decisions (D02) - Liquid d1 (D08) | 1,273 | 1063 | 71 | 62 | 77 | -1.08 to 2.50 |
| S1 | GPT-6 Luna Decisions (D02) - Decider V1.1 27B (D09) | 1,273 | 1053 | 81 | 65 | 74 | -0.61 to 3.14 |
| S1 | GPT-6 Luna Decisions (D02) - Solar Decide Flash (D10) | 1,273 | 881 | 253 | 47 | 92 | 13.67 to 18.71 |
| S1 | GPT-6 Luna Decisions (D02) - Clef Omni (D11) | 1,273 | 960 | 174 | 51 | 88 | 7.42 to 11.94 |
| S1 | GPT-6 Luna Decisions | 1,273 | 889 | 245 | 47 | 92 | 13.07 to 18.06 |

| Family | Model A - Model B | n | Both correct | A only correct | B only correct | Both incorrect | Newcombe 95% CI, pp |
|---|---|---|---|---|---|---|---|
| | (D02) - Drex v1.5 (D12) | | | | | | |
| S1 | GPT-6 Luna Decisions (D02) - Microsoft-Decision-1 (D13) | 1,273 | 953 | 181 | 60 | 79 | 7.17 to 11.86 |
| S1 | Kev 4B (D03) - Solar Decide (D04) | 1,273 | 795 | 67 | 257 | 154 | -17.57 to -12.27 |
| S1 | Kev 4B (D03) - Mercury Decide (D05) | 1,273 | 810 | 52 | 275 | 136 | -20.13 to -14.90 |
| S1 | Kev 4B (D03) - Clef (D06) | 1,273 | 811 | 51 | 268 | 143 | -19.63 to -14.45 |
| S1 | Kev 4B (D03) - Clef Flash (D07) | 1,273 | 761 | 101 | 177 | 234 | -8.52 to -3.42 |
| S1 | Kev 4B (D03) - Liquid d1 (D08) | 1,273 | 826 | 36 | 299 | 112 | -23.25 to -18.07 |
| S1 | Kev 4B (D03) - Decider | 1,273 | 816 | 46 | 302 | 109 | -22.76 to -17.45 |

| Family | Model A - Model B | n | Both correct | A only correct | B only correct | Both incorrect | Newcombe 95% CI, pp |
|---|---|---|---|---|---|---|---|
| | V1.1 27B (D09) | | | | | | |
| S1 | Kev 4B (D03) - Solar Decide Flash (D10) | 1,273 | 744 | 118 | 184 | 227 | -7.84 to -2.52 |
| S1 | Kev 4B (D03) - Clef Omni (D11) | 1,273 | 791 | 71 | 220 | 191 | -14.25 to -9.15 |
| S1 | Kev 4B (D03) - Drex v1.5 (D12) | 1,273 | 752 | 110 | 184 | 227 | -8.43 to -3.18 |
| S1 | Kev 4B (D03) - Microsoft-Decision-1 (D13) | 1,273 | 786 | 76 | 227 | 184 | -14.46 to -9.25 |
| S1 | Solar Decide (D04) - Mercury Decide (D05) | 1,273 | 965 | 87 | 120 | 101 | -4.82 to -0.37 |
| S1 | Solar Decide (D04) - Clef (D06) | 1,273 | 960 | 92 | 119 | 102 | -4.37 to 0.12 |
| S1 | Solar Decide (D04) - | 1,273 | 860 | 192 | 78 | 143 | 6.47 to 11.44 |

| Family | Model A - Model B | n | Both correct | A only correct | B only correct | Both incorrect | Newcombe 95% CI, pp |
|---|---|---|---|---|---|---|---|
| | Clef Flash (D07) | | | | | | |
| S1 | Solar Decide (D04) - Liquid d1 (D08) | 1,273 | 985 | 67 | 140 | 81 | -7.95 to -3.54 |
| S1 | Solar Decide (D04) - Decider V1.1 27B (D09) | 1,273 | 981 | 71 | 137 | 84 | -7.41 to -2.98 |
| S1 | Solar Decide (D04) - Solar Decide Flash (D10) | 1,273 | 913 | 139 | 15 | 206 | 7.91 to 11.60 |
| S1 | Solar Decide (D04) - Clef Omni (D11) | 1,273 | 914 | 138 | 97 | 124 | 0.86 to 5.59 |
| S1 | Solar Decide (D04) - Drex v1.5 (D12) | 1,273 | 838 | 214 | 98 | 123 | 6.43 to 11.79 |
| S1 | Solar Decide (D04) - Microsoft- | 1,273 | 903 | 149 | 110 | 111 | 0.58 to 5.55 |

| Family | Model A - Model B | n | Both correct | A only correct | B only correct | Both incorrect | Newcombe 95% CI, pp |
|---|---|---|---|---|---|---|---|
| | Decision-1 (D13) | | | | | | |
| S1 | Mercury Decide (D05) - Clef (D06) | 1,273 | 1031 | 54 | 48 | 140 | -1.10 to 2.05 |
| S1 | Mercury Decide (D05) - Clef Flash (D07) | 1,273 | 876 | 209 | 62 | 126 | 9.09 to 14.01 |
| S1 | Mercury Decide (D05) - Liquid d1 (D08) | 1,273 | 1033 | 52 | 92 | 96 | -5.01 to -1.30 |
| S1 | Mercury Decide (D05) - Decider V1.1 27B (D09) | 1,273 | 1043 | 42 | 75 | 113 | -4.28 to -0.93 |
| S1 | Mercury Decide (D05) - Solar Decide Flash (D10) | 1,273 | 870 | 215 | 58 | 130 | 9.88 to 14.80 |
| S1 | Mercury Decide (D05) - Clef Omni (D11) | 1,273 | 937 | 148 | 74 | 114 | 3.54 to 8.10 |

| Family | Model A - Model B | n | Both correct | A only correct | B only correct | Both incorrect | Newcombe 95% CI, pp |
|---|---|---|---|---|---|---|---|
| S1 | Mercury Decide (D05) - Drex v1.5 (D12) | 1,273 | 866 | 219 | 70 | 118 | 9.16 to 14.25 |
| S1 | Mercury Decide (D05) - Microsoft-Decision-1 (D13) | 1,273 | 937 | 148 | 76 | 112 | 3.37 to 7.96 |
| S1 | Clef (D06) - Clef Flash (D07) | 1,273 | 884 | 195 | 54 | 140 | 8.72 to 13.44 |
| S1 | Clef (D06) - Liquid d1 (D08) | 1,273 | 1031 | 48 | 94 | 100 | -5.47 to -1.79 |
| S1 | Clef (D06) - Decider V1.1 27B (D09) | 1,273 | 1032 | 47 | 86 | 108 | -4.86 to -1.29 |
| S1 | Clef (D06) - Solar Decide Flash (D10) | 1,273 | 871 | 208 | 57 | 137 | 9.44 to 14.29 |
| S1 | Clef (D06) - Clef Omni (D11) | 1,273 | 938 | 141 | 73 | 121 | 3.11 to 7.59 |
| S1 | Clef (D06) - Drex v1.5 (D12) | 1,273 | 866 | 213 | 70 | 124 | 8.71 to 13.76 |

| Family | Model A - Model B | n | Both correct | A only correct | B only correct | Both incorrect | Newcombe 95% CI, pp |
|---|---|---|---|---|---|---|---|
| S1 | Clef (D06) - Microsoft-Decision-1 (D13) | 1,273 | 934 | 145 | 79 | 115 | 2.89 to 7.49 |
| S1 | Clef Flash (D07) - Liquid d1 (D08) | 1,273 | 901 | 37 | 224 | 111 | -17.06 to -12.34 |
| S1 | Clef Flash (D07) - Decider V1.1 27B (D09) | 1,273 | 891 | 47 | 227 | 108 | -16.58 to -11.71 |
| S1 | Clef Flash (D07) - Solar Decide Flash (D10) | 1,273 | 799 | 139 | 129 | 206 | -1.74 to 3.31 |
| S1 | Clef Flash (D07) - Clef Omni (D11) | 1,273 | 847 | 91 | 164 | 171 | -8.18 to -3.29 |
| S1 | Clef Flash (D07) - Drex v1.5 (D12) | 1,273 | 818 | 120 | 118 | 217 | -2.22 to 2.54 |
| S1 | Clef Flash (D07) - Microsoft-Decision-1 (D13) | 1,273 | 860 | 78 | 153 | 182 | -8.22 to -3.57 |

| Family | Model A - Model B | n | Both correct | A only correct | B only correct | Both incorrect | Newcombe 95% CI, pp |
|---|---|---|---|---|---|---|---|
| S1 | Liquid d1 (D08) - Decider V1.1 27B (D09) | 1,273 | 1052 | 73 | 66 | 82 | -1.28 to 2.39 |
| S1 | Liquid d1 (D08) - Solar Decide Flash (D10) | 1,273 | 881 | 244 | 47 | 101 | 12.99 to 17.98 |
| S1 | Liquid d1 (D08) - Clef Omni (D11) | 1,273 | 966 | 159 | 45 | 103 | 6.82 to 11.13 |
| S1 | Liquid d1 (D08) - Drex v1.5 (D12) | 1,273 | 894 | 231 | 42 | 106 | 12.44 to 17.28 |
| S1 | Liquid d1 (D08) - Microsoft-Decision-1 (D13) | 1,273 | 959 | 166 | 54 | 94 | 6.57 to 11.06 |
| S1 | Decider V1.1 27B (D09) - Solar Decide Flash (D10) | 1,273 | 873 | 245 | 55 | 100 | 12.38 to 17.47 |
| S1 | Decider V1.1 27B (D09) - | 1,273 | 953 | 165 | 58 | 97 | 6.15 to 10.68 |

| Family | Model A - Model B | n | Both correct | A only correct | B only correct | Both incorrect | Newcombe 95% CI, pp |
|---|---|---|---|---|---|---|---|
| | Clef Omni (D11) | | | | | | |
| S1 | Decider V1.1 27B (D09) - Drex v1.5 (D12) | 1,273 | 880 | 238 | 56 | 99 | 11.77 to 16.83 |
| S1 | Decider V1.1 27B (D09) - Microsoft-Decision-1 (D13) | 1,273 | 953 | 165 | 60 | 95 | 5.98 to 10.54 |
| S1 | Solar Decide Flash (D10) - Clef Omni (D11) | 1,273 | 835 | 93 | 176 | 169 | -9.03 to -4.01 |
| S1 | Solar Decide Flash (D10) - Drex v1.5 (D12) | 1,273 | 781 | 147 | 155 | 190 | -3.31 to 2.05 |
| S1 | Solar Decide Flash (D10) - Microsoft-Decision-1 (D13) | 1,273 | 829 | 99 | 184 | 161 | -9.25 to -4.11 |
| S1 | Clef Omni (D11) - | 1,273 | 831 | 180 | 105 | 157 | 3.31 to 8.48 |

| Family | Model A - Model B | n | Both correct | A only correct | B only correct | Both incorrect | Newcombe 95% CI, pp |
|---|---|---|---|---|---|---|---|
| | Drex v1.5 (D12) | | | | | | |
| S1 | Clef Omni (D11) - Microsoft-Decision-1 (D13) | 1,273 | 890 | 121 | 123 | 139 | -2.57 to 2.26 |
| S1 | Drex v1.5 (D12) - Microsoft-Decision-1 (D13) | 1,273 | 849 | 87 | 164 | 173 | -8.47 to -3.63 |
| S2 | GPT-6 Luna High (G03) - GPT-6 Luna Low (G02) | 1,273 | 1190 | 22 | 9 | 52 | 0.14 to 1.96 |
| S2 | DeepSeek V4.1 Flash High (G07) - DeepSeek V4.1 Flash Low (G06) | 1,273 | 1198 | 6 | 18 | 51 | -1.79 to -0.16 |
| S2 | Qwen3.8-27B Xhigh (G10) - Qwen3.8-27B Low (G09) | 1,273 | 1178 | 37 | 20 | 38 | 0.16 to 2.56 |

**Supplementary Table S3. Exploratory native-effort accuracy contrasts.** Accuracy at lower native effort was subtracted from accuracy at higher effort using 1,273 matched questions, with pointwise 95% question-bootstrap confidence intervals. Holm adjustment was applied to these three contrasts. Original

failure counts for the lower/higher effort settings were 0/0 for Luna Low/High, 34/25 for DeepSeek Low/High and 0/33 for Qwen Low/Xhigh. One Qwen Xhigh failure remained after retries. Although the pointwise intervals excluded zero, all three adjusted P values were .068, so none met the 0.05 threshold after adjustment for multiple comparisons.

| Higher - lower setting | n | Effect, pp | Bootstrap 95% CI, pp | Exact P | Holm P |
|---|---|---|---|---|---|
| GPT-6 Luna High (G03) - GPT-6 Luna Low (G02) | 1,273 | 1.02 | 0.16 to 1.89 | .029 | .068 |
| DeepSeek V4.1 Flash High (G07) - DeepSeek V4.1 Flash Low (G06) | 1,273 | -0.94 | -1.73 to -0.24 | .023 | .068 |
| Qwen3.8-27B Xhigh (G10) - Qwen3.8-27B Low (G09) | 1,273 | 1.34 | 0.16 to 2.51 | .033 | .068 |

**Supplementary Table S4. Reliability of run-1 native selected-option probabilities.** Native selected-option probabilities were grouped into ten fixed bins, each including its lower boundary and excluding its upper boundary, except the final bin, which included 1. Observed accuracy and its pointwise Wilson interval were calculated from the eligible answers within each bin. Means and intervals remain unavailable for empty bins.

| Model (ID) | Probability bin | n | Correct | Mean probability | Accuracy (%) | Wilson 95% CI (%) |
|---|---|---|---|---|---|---|
| JEV 1.13 (D01) | [0.0, 0.1) | 0 | 0 | Unavailable | Unavailable | Unavailable |
| JEV 1.13 (D01) | [0.1, 0.2) | 0 | 0 | Unavailable | Unavailable | Unavailable |
| JEV 1.13 (D01) | [0.2, 0.3) | 1 | 0 | 0.290 | 0.00 | 0.00 to 79.35 |
| JEV 1.13 (D01) | [0.3, 0.4) | 7 | 5 | 0.376 | 71.43 | 35.89 to 91.78 |

| Model (ID) | Probability bin | n | Correct | Mean probability | Accuracy (%) | Wilson 95% CI (%) |
|---|---|---|---|---|---|---|
| JEV 1.13 (D01) | [0.4, 0.5) | 49 | 21 | 0.460 | 42.86 | 30.02 to 56.73 |
| JEV 1.13 (D01) | [0.5, 0.6) | 65 | 34 | 0.549 | 52.31 | 40.38 to 63.98 |
| JEV 1.13 (D01) | [0.6, 0.7) | 79 | 46 | 0.642 | 58.23 | 47.22 to 68.47 |
| JEV 1.13 (D01) | [0.7, 0.8) | 77 | 52 | 0.753 | 67.53 | 56.46 to 76.94 |
| JEV 1.13 (D01) | [0.8, 0.9) | 118 | 104 | 0.850 | 88.14 | 81.07 to 92.80 |
| JEV 1.13 (D01) | [0.9, 1.0] | 877 | 862 | 0.982 | 98.29 | 97.20 to 98.96 |
| GPT-6 Luna Decisions (D02) | [0.0, 0.1) | 0 | 0 | Unavailable | Unavailable | Unavailable |
| GPT-6 Luna Decisions (D02) | [0.1, 0.2) | 0 | 0 | Unavailable | Unavailable | Unavailable |
| GPT-6 Luna Decisions (D02) | [0.2, 0.3) | 1 | 0 | 0.290 | 0.00 | 0.00 to 79.35 |
| GPT-6 Luna Decisions (D02) | [0.3, 0.4) | 14 | 6 | 0.362 | 42.86 | 21.38 to 67.41 |
| GPT-6 Luna Decisions (D02) | [0.4, 0.5) | 42 | 19 | 0.455 | 45.24 | 31.22 to 60.05 |
| GPT-6 Luna Decisions (D02) | [0.5, 0.6) | 50 | 32 | 0.544 | 64.00 | 50.14 to 75.86 |
| GPT-6 Luna Decisions (D02) | [0.6, 0.7) | 39 | 19 | 0.646 | 48.72 | 33.87 to 63.80 |

| Model (ID) | Probability bin | n | Correct | Mean probability | Accuracy (%) | Wilson 95% CI (%) |
|---|---|---|---|---|---|---|
| GPT-6 Luna Decisions (D02) | [0.7, 0.8) | 64 | 42 | 0.736 | 65.62 | 53.40 to 76.08 |
| GPT-6 Luna Decisions (D02) | [0.8, 0.9) | 82 | 63 | 0.851 | 76.83 | 66.62 to 84.63 |
| GPT-6 Luna Decisions (D02) | [0.9, 1.0] | 979 | 953 | 0.991 | 97.34 | 96.14 to 98.18 |
| Kev 4B (D03) | [0.0, 0.1) | 0 | 0 | Unavailable | Unavailable | Unavailable |
| Kev 4B (D03) | [0.1, 0.2) | 0 | 0 | Unavailable | Unavailable | Unavailable |
| Kev 4B (D03) | [0.2, 0.3) | 12 | 2 | 0.282 | 16.67 | 4.70 to 44.80 |
| Kev 4B (D03) | [0.3, 0.4) | 153 | 61 | 0.356 | 39.87 | 32.45 to 47.78 |
| Kev 4B (D03) | [0.4, 0.5) | 225 | 105 | 0.447 | 46.67 | 40.26 to 53.19 |
| Kev 4B (D03) | [0.5, 0.6) | 206 | 113 | 0.546 | 54.85 | 48.03 to 61.50 |
| Kev 4B (D03) | [0.6, 0.7) | 171 | 116 | 0.650 | 67.84 | 60.51 to 74.38 |
| Kev 4B (D03) | [0.7, 0.8) | 138 | 114 | 0.752 | 82.61 | 75.43 to 88.03 |
| Kev 4B (D03) | [0.8, 0.9) | 169 | 158 | 0.855 | 93.49 | 88.72 to 96.33 |
| Kev 4B (D03) | [0.9, 1.0] | 199 | 193 | 0.941 | 96.98 | 93.58 to 98.61 |
| Solar Decide (D04) | [0.0, 0.1) | 0 | 0 | Unavailable | Unavailable | Unavailable |

| **Model (ID)** | **Probability bin** | **n** | **Correct** | **Mean probability** | **Accuracy (%)** | **Wilson 95% CI (%)** |
|---|---|---|---|---|---|---|
| Solar Decide (D04) | [0.1, 0.2) | 0 | 0 | Unavailable | Unavailable | Unavailable |
| Solar Decide (D04) | [0.2, 0.3) | 2 | 0 | 0.285 | 0.00 | 0.00 to 65.76 |
| Solar Decide (D04) | [0.3, 0.4) | 7 | 2 | 0.351 | 28.57 | 8.22 to 64.11 |
| Solar Decide (D04) | [0.4, 0.5) | 17 | 3 | 0.448 | 17.65 | 6.19 to 41.03 |
| Solar Decide (D04) | [0.5, 0.6) | 16 | 7 | 0.548 | 43.75 | 23.10 to 66.82 |
| Solar Decide (D04) | [0.6, 0.7) | 16 | 10 | 0.656 | 62.50 | 38.64 to 81.52 |
| Solar Decide (D04) | [0.7, 0.8) | 12 | 6 | 0.747 | 50.00 | 25.38 to 74.62 |
| Solar Decide (D04) | [0.8, 0.9) | 186 | 117 | 0.856 | 62.90 | 55.77 to 69.52 |
| Solar Decide (D04) | [0.9, 1.0] | 1017 | 907 | 0.979 | 89.18 | 87.13 to 90.95 |
| Mercury Decide (D05) | [0.0, 0.1) | 0 | 0 | Unavailable | Unavailable | Unavailable |
| Mercury Decide (D05) | [0.1, 0.2) | 0 | 0 | Unavailable | Unavailable | Unavailable |
| Mercury Decide (D05) | [0.2, 0.3) | 2 | 1 | 0.285 | 50.00 | 9.45 to 90.55 |
| Mercury Decide (D05) | [0.3, 0.4) | 3 | 0 | 0.376 | 0.00 | 0.00 to 56.15 |
| Mercury Decide (D05) | [0.4, 0.5) | 9 | 3 | 0.443 | 33.33 | 12.06 to 64.58 |

| Model (ID) | Probability bin | n | Correct | Mean probability | Accuracy (%) | Wilson 95% CI (%) |
|---|---|---|---|---|---|---|
| Mercury Decide (D05) | [0.5, 0.6) | 32 | 13 | 0.544 | 40.62 | 25.52 to 57.74 |
| Mercury Decide (D05) | [0.6, 0.7) | 46 | 21 | 0.652 | 45.65 | 32.15 to 59.82 |
| Mercury Decide (D05) | [0.7, 0.8) | 50 | 23 | 0.754 | 46.00 | 32.97 to 59.60 |
| Mercury Decide (D05) | [0.8, 0.9) | 89 | 59 | 0.858 | 66.29 | 55.98 to 75.26 |
| Mercury Decide (D05) | [0.9, 1.0] | 1042 | 965 | 0.983 | 92.61 | 90.86 to 94.05 |
| Clef (D06) | [0.0, 0.1) | 0 | 0 | Unavailable | Unavailable | Unavailable |
| Clef (D06) | [0.1, 0.2) | 0 | 0 | Unavailable | Unavailable | Unavailable |
| Clef (D06) | [0.2, 0.3) | 1 | 0 | 0.265 | 0.00 | 0.00 to 79.35 |
| Clef (D06) | [0.3, 0.4) | 4 | 0 | 0.352 | 0.00 | 0.00 to 48.99 |
| Clef (D06) | [0.4, 0.5) | 45 | 17 | 0.460 | 37.78 | 25.11 to 52.37 |
| Clef (D06) | [0.5, 0.6) | 55 | 26 | 0.551 | 47.27 | 34.69 to 60.21 |
| Clef (D06) | [0.6, 0.7) | 70 | 38 | 0.651 | 54.29 | 42.70 to 65.43 |
| Clef (D06) | [0.7, 0.8) | 94 | 56 | 0.754 | 59.57 | 49.47 to 68.93 |
| Clef (D06) | [0.8, 0.9) | 172 | 138 | 0.857 | 80.23 | 73.65 to 85.50 |

| Model (ID) | Probability bin | n | Correct | Mean probability | Accuracy (%) | Wilson 95% CI (%) |
|---|---|---|---|---|---|---|
| Clef (D06) | [0.9, 1.0] | 832 | 804 | 0.966 | 96.63 | 95.18 to 97.66 |
| Clef Flash (D07) | [0.0, 0.1) | 0 | 0 | Unavailable | Unavailable | Unavailable |
| Clef Flash (D07) | [0.1, 0.2) | 0 | 0 | Unavailable | Unavailable | Unavailable |
| Clef Flash (D07) | [0.2, 0.3) | 4 | 1 | 0.285 | 25.00 | 4.56 to 69.94 |
| Clef Flash (D07) | [0.3, 0.4) | 22 | 5 | 0.360 | 22.73 | 10.12 to 43.44 |
| Clef Flash (D07) | [0.4, 0.5) | 61 | 21 | 0.451 | 34.43 | 23.75 to 46.95 |
| Clef Flash (D07) | [0.5, 0.6) | 91 | 39 | 0.554 | 42.86 | 33.18 to 53.11 |
| Clef Flash (D07) | [0.6, 0.7) | 113 | 57 | 0.654 | 50.44 | 41.36 to 59.49 |
| Clef Flash (D07) | [0.7, 0.8) | 129 | 83 | 0.747 | 64.34 | 55.77 to 72.08 |
| Clef Flash (D07) | [0.8, 0.9) | 367 | 293 | 0.858 | 79.84 | 75.43 to 83.62 |
| Clef Flash (D07) | [0.9, 1.0] | 486 | 439 | 0.936 | 90.33 | 87.38 to 92.65 |
| Liquid d1 (D08) | [0.0, 0.1) | 0 | 0 | Unavailable | Unavailable | Unavailable |
| Liquid d1 (D08) | [0.1, 0.2) | 0 | 0 | Unavailable | Unavailable | Unavailable |
| Liquid d1 (D08) | [0.2, 0.3) | 1 | 1 | 0.297 | 100.00 | 20.65 to 100.00 |
| Liquid d1 (D08) | [0.3, 0.4) | 11 | 4 | 0.372 | 36.36 | 15.17 to 64.62 |
| Liquid d1 (D08) | [0.4, 0.5) | 36 | 16 | 0.456 | 44.44 | 29.54 to 60.42 |

| Model (ID) | Probability bin | n | Correct | Mean probability | Accuracy (%) | Wilson 95% CI (%) |
|---|---|---|---|---|---|---|
| Liquid d1 (D08) | [0.5, 0.6) | 80 | 42 | 0.553 | 52.50 | 41.70 to 63.08 |
| Liquid d1 (D08) | [0.6, 0.7) | 60 | 38 | 0.655 | 63.33 | 50.68 to 74.38 |
| Liquid d1 (D08) | [0.7, 0.8) | 81 | 56 | 0.755 | 69.14 | 58.40 to 78.14 |
| Liquid d1 (D08) | [0.8, 0.9) | 137 | 117 | 0.855 | 85.40 | 78.53 to 90.35 |
| Liquid d1 (D08) | [0.9, 1.0] | 867 | 851 | 0.983 | 98.15 | 97.02 to 98.86 |
| Decider V1.1 27B (D09) | [0.0, 0.1) | 0 | 0 | Unavailable | Unavailable | Unavailable |
| Decider V1.1 27B (D09) | [0.1, 0.2) | 0 | 0 | Unavailable | Unavailable | Unavailable |
| Decider V1.1 27B (D09) | [0.2, 0.3) | 1 | 1 | 0.267 | 100.00 | 20.65 to 100.00 |
| Decider V1.1 27B (D09) | [0.3, 0.4) | 9 | 0 | 0.371 | 0.00 | 0.00 to 29.91 |
| Decider V1.1 27B (D09) | [0.4, 0.5) | 33 | 13 | 0.460 | 39.39 | 24.68 to 56.32 |
| Decider V1.1 27B (D09) | [0.5, 0.6) | 65 | 34 | 0.548 | 52.31 | 40.38 to 63.98 |
| Decider V1.1 27B (D09) | [0.6, 0.7) | 53 | 31 | 0.654 | 58.49 | 45.09 to 70.74 |

| Model (ID) | Probability bin | n | Correct | Mean probability | Accuracy (%) | Wilson 95% CI (%) |
|---|---|---|---|---|---|---|
| Decider V1.1 27B (D09) | [0.7, 0.8) | 55 | 32 | 0.754 | 58.18 | 45.03 to 70.26 |
| Decider V1.1 27B (D09) | [0.8, 0.9) | 94 | 73 | 0.854 | 77.66 | 68.25 to 84.90 |
| Decider V1.1 27B (D09) | [0.9, 1.0] | 963 | 934 | 0.988 | 96.99 | 95.71 to 97.90 |
| Solar Decide Flash (D10) | [0.0, 0.1) | 0 | 0 | Unavailable | Unavailable | Unavailable |
| Solar Decide Flash (D10) | [0.1, 0.2) | 0 | 0 | Unavailable | Unavailable | Unavailable |
| Solar Decide Flash (D10) | [0.2, 0.3) | 1 | 0 | 0.299 | 0.00 | 0.00 to 79.35 |
| Solar Decide Flash (D10) | [0.3, 0.4) | 24 | 3 | 0.361 | 12.50 | 4.34 to 31.00 |
| Solar Decide Flash (D10) | [0.4, 0.5) | 56 | 29 | 0.459 | 51.79 | 39.01 to 64.33 |
| Solar Decide Flash (D10) | [0.5, 0.6) | 82 | 33 | 0.548 | 40.24 | 30.30 to 51.06 |
| Solar Decide Flash (D10) | [0.6, 0.7) | 97 | 47 | 0.654 | 48.45 | 38.76 to 58.27 |
| Solar Decide Flash (D10) | [0.7, 0.8) | 105 | 63 | 0.749 | 60.00 | 50.44 to 68.86 |
| Solar Decide Flash (D10) | [0.8, 0.9) | 156 | 97 | 0.853 | 62.18 | 54.36 to 69.41 |
| Solar Decide Flash (D10) | [0.9, 1.0] | 752 | 656 | 0.972 | 87.23 | 84.66 to 89.43 |
| Clef Omni (D11) | [0.0, 0.1) | 0 | 0 | Unavailable | Unavailable | Unavailable |

| Model (ID) | Probability bin | n | Correct | Mean probability | Accuracy (%) | Wilson 95% CI (%) |
|---|---|---|---|---|---|---|
| Clef Omni (D11) | [0.1, 0.2) | 0 | 0 | Unavailable | Unavailable | Unavailable |
| Clef Omni (D11) | [0.2, 0.3) | 1 | 1 | 0.290 | 100.00 | 20.65 to 100.00 |
| Clef Omni (D11) | [0.3, 0.4) | 33 | 14 | 0.368 | 42.42 | 27.24 to 59.19 |
| Clef Omni (D11) | [0.4, 0.5) | 78 | 31 | 0.456 | 39.74 | 29.61 to 50.84 |
| Clef Omni (D11) | [0.5, 0.6) | 85 | 37 | 0.551 | 43.53 | 33.50 to 54.12 |
| Clef Omni (D11) | [0.6, 0.7) | 113 | 71 | 0.651 | 62.83 | 53.64 to 71.18 |
| Clef Omni (D11) | [0.7, 0.8) | 131 | 97 | 0.757 | 74.05 | 65.93 to 80.79 |
| Clef Omni (D11) | [0.8, 0.9) | 328 | 283 | 0.861 | 86.28 | 82.13 to 89.59 |
| Clef Omni (D11) | [0.9, 1.0] | 504 | 477 | 0.928 | 94.64 | 92.32 to 96.29 |
| Drex v1.5 (D12) | [0.0, 0.1) | 0 | 0 | Unavailable | Unavailable | Unavailable |
| Drex v1.5 (D12) | [0.1, 0.2) | 0 | 0 | Unavailable | Unavailable | Unavailable |
| Drex v1.5 (D12) | [0.2, 0.3) | 0 | 0 | Unavailable | Unavailable | Unavailable |
| Drex v1.5 (D12) | [0.3, 0.4) | 14 | 6 | 0.368 | 42.86 | 21.38 to 67.41 |
| Drex v1.5 (D12) | [0.4, 0.5) | 66 | 33 | 0.458 | 50.00 | 38.27 to 61.73 |
| Drex v1.5 (D12) | [0.5, 0.6) | 94 | 39 | 0.552 | 41.49 | 32.06 to 51.59 |
| Drex v1.5 (D12) | [0.6, 0.7) | 115 | 54 | 0.648 | 46.96 | 38.08 to 56.03 |

| Model (ID) | Probability bin | n | Correct | Mean probability | Accuracy (%) | Wilson 95% CI (%) |
|---|---|---|---|---|---|---|
| Drex v1.5 (D12) | [0.7, 0.8) | 124 | 82 | 0.751 | 66.13 | 57.43 to 73.86 |
| Drex v1.5 (D12) | [0.8, 0.9) | 163 | 105 | 0.854 | 64.42 | 56.81 to 71.36 |
| Drex v1.5 (D12) | [0.9, 1.0] | 697 | 617 | 0.971 | 88.52 | 85.94 to 90.68 |
| Microsoft-Decision-1 (D13) | [0.0, 0.1) | 0 | 0 | Unavailable | Unavailable | Unavailable |
| Microsoft-Decision-1 (D13) | [0.1, 0.2) | 0 | 0 | Unavailable | Unavailable | Unavailable |
| Microsoft-Decision-1 (D13) | [0.2, 0.3) | 0 | 0 | Unavailable | Unavailable | Unavailable |
| Microsoft-Decision-1 (D13) | [0.3, 0.4) | 23 | 7 | 0.367 | 30.43 | 15.60 to 50.87 |
| Microsoft-Decision-1 (D13) | [0.4, 0.5) | 65 | 27 | 0.456 | 41.54 | 30.36 to 53.66 |
| Microsoft-Decision-1 (D13) | [0.5, 0.6) | 98 | 50 | 0.549 | 51.02 | 41.27 to 60.69 |
| Microsoft-Decision-1 (D13) | [0.6, 0.7) | 81 | 50 | 0.656 | 61.73 | 50.84 to 71.55 |
| Microsoft-Decision-1 (D13) | [0.7, 0.8) | 109 | 67 | 0.750 | 61.47 | 52.09 to 70.07 |
| Microsoft-Decision-1 (D13) | [0.8, 0.9) | 199 | 150 | 0.855 | 75.38 | 68.95 to 80.84 |

| Model (ID) | Probability bin | n | Correct | Mean probability | Accuracy (%) | Wilson 95% CI (%) |
|---|---|---|---|---|---|---|
| Microsoft-Decision-1 (D13) | [0.9, 1.0] | 698 | 662 | 0.967 | 94.84 | 92.94 to 96.25 |

**Supplementary Table S5. Support of native selected-option probabilities in run 1.** Scores were the native probabilities assigned to the selected labels in run 1. Counts of distinct values and values at the endpoints used the stored probabilities. The two invalid final Luna Decisions responses were excluded. Differences in the precision returned by providers affected the counts of distinct scores and exact matches.

| Model (ID) | Eligible n | Excluded n | Distinct scores | Minimum | Maximum | Exactly 0, n | Exactly 1, n | Correct n | Incorrect n |
|---|---|---|---|---|---|---|---|---|---|
| JEV 1.13 (D01) | 1273 | 0 | 73 | 0.290000 | 1.000000 | 0 | 450 | 1124 | 149 |
| GPT-6 Luna Decisions (D02) | 1271 | 2 | 68 | 0.290000 | 1.000000 | 0 | 707 | 1134 | 137 |
| Kev 4B (D03) | 1273 | 0 | 1165 | 0.261600 | 0.992100 | 0 | 0 | 862 | 411 |
| Solar Decide (D04) | 1273 | 0 | 1186 | 0.282183 | 0.999975 | 0 | 0 | 1052 | 221 |
| Mercury Decide (D05) | 1273 | 0 | 1273 | 0.271202 | 0.999793 | 0 | 0 | 1085 | 188 |
| Clef (D06) | 1273 | 0 | 884 | 0.265400 | 0.994000 | 0 | 0 | 1079 | 194 |
| Clef Flash (D07) | 1273 | 0 | 1051 | 0.267400 | 0.987000 | 0 | 0 | 938 | 335 |

| Model (ID) | Eligible n | Excluded n | Distinct scores | Minimum | Maximum | Exactly 0, n | Exactly 1, n | Correct n | Incorrect n |
|---|---|---|---|---|---|---|---|---|---|
| Liquid d1 (D08) | 1273 | 0 | 1273 | 0.296602 | 0.999987 | 0 | 0 | 1125 | 148 |
| Decider V1.1 27B (D09) | 1273 | 0 | 1273 | 0.267218 | 0.999960 | 0 | 0 | 1118 | 155 |
| Solar Decide Flash (D10) | 1273 | 0 | 1225 | 0.298633 | 0.999208 | 0 | 0 | 928 | 345 |
| Clef Omni (D11) | 1273 | 0 | 1037 | 0.290200 | 0.973000 | 0 | 0 | 1011 | 262 |
| Drex v1.5 (D12) | 1273 | 0 | 954 | 0.323100 | 0.999800 | 0 | 0 | 936 | 337 |
| Microsoft-Decision-1 (D13) | 1273 | 0 | 1269 | 0.318945 | 0.997324 | 0 | 0 | 1013 | 260 |

**Supplementary Table S6. Question-matched correctness and failure transitions.** Transitions between correctness and failure states were counted for all 1,273 matched questions, including failed outcomes. Luna Decisions failed on the same two questions in every run, while the other decision models returned valid final answers throughout.

| Model (ID) | Runs | Correct → correct | Correct → incorrect | Incorrect → correct | Incorrect → incorrect | Transitions involving failure | Valid pairs | Same selected label |
|---|---|---|---|---|---|---|---|---|
| JEV 1.13 (D01) | 1-2 | 1114 | 10 | 6 | 143 | 0 | 1273 | 1255 |

| Model (ID) | Runs | Correct → correct | Correct → incorrect | Incorrect → correct | Incorrect → incorrect | Transitions involving failure | Valid pairs | Same selected label |
|---|---|---|---|---|---|---|---|---|
| JEV 1.13 (D01) | 1-3 | 1117 | 7 | 5 | 144 | 0 | 1273 | 1261 |
| JEV 1.13 (D01) | 2-3 | 1113 | 7 | 9 | 144 | 0 | 1273 | 1255 |
| GPT-6 Luna Decisions (D02) | 1-2 | 1134 | 0 | 0 | 137 | 2 | 1271 | 1271 |
| GPT-6 Luna Decisions (D02) | 1-3 | 1134 | 0 | 0 | 137 | 2 | 1271 | 1271 |
| GPT-6 Luna Decisions (D02) | 2-3 | 1134 | 0 | 0 | 137 | 2 | 1271 | 1271 |
| Kev 4B (D03) | 1-2 | 862 | 0 | 0 | 411 | 0 | 1273 | 1271 |
| Kev 4B (D03) | 1-3 | 862 | 0 | 0 | 411 | 0 | 1273 | 1271 |
| Kev 4B (D03) | 2-3 | 862 | 0 | 0 | 411 | 0 | 1273 | 1273 |
| Solar Decide (D04) | 1-2 | 1021 | 31 | 27 | 194 | 0 | 1273 | 1202 |
| Solar Decide (D04) | 1-3 | 1023 | 29 | 31 | 190 | 0 | 1273 | 1200 |
| Solar Decide (D04) | 2-3 | 1021 | 27 | 33 | 192 | 0 | 1273 | 1205 |

| Model (ID) | Runs | Correct → correct | Correct → incorrect | Incorrect → correct | Incorrect → incorrect | Transitions involving failure | Valid pairs | Same selected label |
|---|---|---|---|---|---|---|---|---|
| Mercury Decide (D05) | 1-2 | 1085 | 0 | 8 | 180 | 0 | 1273 | 1263 |
| Mercury Decide (D05) | 1-3 | 1082 | 3 | 7 | 181 | 0 | 1273 | 1261 |
| Mercury Decide (D05) | 2-3 | 1086 | 7 | 3 | 177 | 0 | 1273 | 1263 |
| Clef (D06) | 1-2 | 1079 | 0 | 0 | 194 | 0 | 1273 | 1273 |
| Clef (D06) | 1-3 | 1079 | 0 | 0 | 194 | 0 | 1273 | 1273 |
| Clef (D06) | 2-3 | 1079 | 0 | 0 | 194 | 0 | 1273 | 1273 |
| Clef Flash (D07) | 1-2 | 938 | 0 | 0 | 335 | 0 | 1273 | 1273 |
| Clef Flash (D07) | 1-3 | 938 | 0 | 0 | 335 | 0 | 1273 | 1273 |
| Clef Flash (D07) | 2-3 | 938 | 0 | 0 | 335 | 0 | 1273 | 1273 |
| Liquid d1 (D08) | 1-2 | 1120 | 5 | 10 | 138 | 0 | 1273 | 1257 |
| Liquid d1 (D08) | 1-3 | 1117 | 8 | 10 | 138 | 0 | 1273 | 1254 |
| Liquid d1 (D08) | 2-3 | 1126 | 4 | 1 | 142 | 0 | 1273 | 1266 |

| Model (ID) | Runs | Correct → correct | Correct → incorrect | Incorrect → correct | Incorrect → incorrect | Transitions involving failure | Valid pairs | Same selected label |
|---|---|---|---|---|---|---|---|---|
| Decider V1.1 27B (D09) | 1-2 | 1118 | 0 | 0 | 155 | 0 | 1273 | 1273 |
| Decider V1.1 27B (D09) | 1-3 | 1118 | 0 | 0 | 155 | 0 | 1273 | 1273 |
| Decider V1.1 27B (D09) | 2-3 | 1118 | 0 | 0 | 155 | 0 | 1273 | 1273 |
| Solar Decide Flash (D10) | 1-2 | 917 | 11 | 12 | 333 | 0 | 1273 | 1247 |
| Solar Decide Flash (D10) | 1-3 | 911 | 17 | 12 | 333 | 0 | 1273 | 1240 |
| Solar Decide Flash (D10) | 2-3 | 913 | 16 | 10 | 334 | 0 | 1273 | 1242 |
| Clef Omni (D11) | 1-2 | 1011 | 0 | 1 | 261 | 0 | 1273 | 1272 |
| Clef Omni (D11) | 1-3 | 1011 | 0 | 1 | 261 | 0 | 1273 | 1272 |
| Clef Omni (D11) | 2-3 | 1012 | 0 | 0 | 261 | 0 | 1273 | 1273 |

| Model (ID) | Runs | Correct → correct | Correct → incorrect | Incorrect → correct | Incorrect → incorrect | Transitions involving failure | Valid pairs | Same selected label |
|---|---|---|---|---|---|---|---|---|
| Drex v1.5 (D12) | 1-2 | 921 | 15 | 10 | 327 | 0 | 1273 | 1246 |
| Drex v1.5 (D12) | 1-3 | 921 | 15 | 10 | 327 | 0 | 1273 | 1246 |
| Drex v1.5 (D12) | 2-3 | 931 | 0 | 0 | 342 | 0 | 1273 | 1273 |
| Microsoft-Decision-1 (D13) | 1-2 | 1004 | 9 | 12 | 248 | 0 | 1273 | 1247 |
| Microsoft-Decision-1 (D13) | 1-3 | 1005 | 8 | 10 | 250 | 0 | 1273 | 1246 |
| Microsoft-Decision-1 (D13) | 2-3 | 1001 | 15 | 14 | 243 | 0 | 1273 | 1233 |

**Supplementary Table S7. Exact three-run equality of decision-model probabilities.** Each measure included only questions with usable values in all three runs, with percentages calculated from these eligible triples. Exact equality was assessed using recorded native probabilities before display rounding, without normalizing the four-option vectors. The proportion with exactly equal vectors ranged from 0% for Solar Decide Flash to 100% for Luna Decisions, Clef, Clef Flash and Decider V1.1 27B. The precision of the recorded probabilities differed between providers.

| Model (ID) | Measure | Eligible triples | Excluded triples | Exactly unchanged triples | Exactly unchanged (%) |
|---|---|---|---|---|---|
| JEV 1.13 (D01) | Native selected- | 1,273 | 0 | 574 | 45.09 |

| Model (ID) | Measure | Eligible triples | Excluded triples | Exactly unchanged triples | Exactly unchanged (%) |
|---|---|---|---|---|---|
| | answer probability | | | | |
| JEV 1.13 (D01) | Native four-option vector | 1,273 | 0 | 568 | 44.62 |
| GPT-6 Luna Decisions (D02) | Native selected-answer probability | 1,271 | 2 | 1,271 | 100.00 |
| GPT-6 Luna Decisions (D02) | Native four-option vector | 1,271 | 2 | 1,271 | 100.00 |
| Kev 4B (D03) | Native selected-answer probability | 1,273 | 0 | 1,000 | 78.55 |
| Kev 4B (D03) | Native four-option vector | 1,273 | 0 | 998 | 78.40 |
| Solar Decide (D04) | Native selected-answer probability | 1,273 | 0 | 2 | 0.16 |
| Solar Decide (D04) | Native four-option vector | 1,273 | 0 | 2 | 0.16 |
| Mercury Decide (D05) | Native selected-answer probability | 1,273 | 0 | 100 | 7.86 |
| Mercury Decide (D05) | Native four-option vector | 1,273 | 0 | 100 | 7.86 |
| Clef (D06) | Native selected-answer probability | 1,273 | 0 | 1,273 | 100.00 |

| Model (ID) | Measure | Eligible triples | Excluded triples | Exactly unchanged triples | Exactly unchanged (%) |
|---|---|---|---|---|---|
| Clef (D06) | Native four-option vector | 1,273 | 0 | 1,273 | 100.00 |
| Clef Flash (D07) | Native selected-answer probability | 1,273 | 0 | 1,273 | 100.00 |
| Clef Flash (D07) | Native four-option vector | 1,273 | 0 | 1,273 | 100.00 |
| Liquid d1 (D08) | Native selected-answer probability | 1,273 | 0 | 800 | 62.84 |
| Liquid d1 (D08) | Native four-option vector | 1,273 | 0 | 800 | 62.84 |
| Decider V1.1 27B (D09) | Native selected-answer probability | 1,273 | 0 | 1,273 | 100.00 |
| Decider V1.1 27B (D09) | Native four-option vector | 1,273 | 0 | 1,273 | 100.00 |
| Solar Decide Flash (D10) | Native selected-answer probability | 1,273 | 0 | 0 | 0.00 |
| Solar Decide Flash (D10) | Native four-option vector | 1,273 | 0 | 0 | 0.00 |
| Clef Omni (D11) | Native selected-answer probability | 1,273 | 0 | 1,243 | 97.64 |
| Clef Omni (D11) | Native four-option vector | 1,273 | 0 | 1,243 | 97.64 |

| Model (ID) | Measure | Eligible triples | Excluded triples | Exactly unchanged triples | Exactly unchanged (%) |
|---|---|---|---|---|---|
| Drex v1.5 (D12) | Native selected-answer probability | 1,273 | 0 | 603 | 47.37 |
| Drex v1.5 (D12) | Native four-option vector | 1,273 | 0 | 597 | 46.90 |
| Microsoft-Decision-1 (D13) | Native selected-answer probability | 1,273 | 0 | 21 | 1.65 |
| Microsoft-Decision-1 (D13) | Native four-option vector | 1,273 | 0 | 21 | 1.65 |

**Supplementary Table S8. Pairwise changes in decision-model probabilities.** Runs were matched by question for each pairwise comparison. Selected-answer probability compares the native probability assigned to each run's own selected label. Whole-vector total variation was calculated from normalized distributions. Exact unchanged counts used the stored values before display rounding. The mean, median and 95th percentile are shown to four decimals, so small nonzero changes can appear as 0.0000. All eligible pairs were unchanged for Luna Decisions, Clef, Clef Flash and Decider V1.1 27B.

| Model (ID) | Runs | Measure | Eligible pairs | Exactly unchanged n | Exactly unchanged (%) | Mean absolute change | Median | p95 |
|---|---|---|---|---|---|---|---|---|
| JEV 1.13 (D01) | 1-2 | Native selected-answer probability | 1,273 | 691 | 54.28 | 0.0147 | 0.0000 | 0.0600 |
| JEV 1.13 (D01) | 1-2 | Earlier selected-option probability | 1,273 | 690 | 54.20 | 0.0153 | 0.0000 | 0.0700 |

| Model (ID) | Runs | Measure | Eligible pairs | Exactly unchanged n | Exactly unchanged (%) | Mean absolute change | Median | p95 |
|---|---|---|---|---|---|---|---|---|
| JEV 1.13 (D01) | 1-2 | Normalized vector total variation | 1,273 | 661 | 51.92 | 0.0169 | 0.0000 | 0.0700 |
| JEV 1.13 (D01) | 1-3 | Native selected-answer probability | 1,273 | 693 | 54.44 | 0.0144 | 0.0000 | 0.0700 |
| JEV 1.13 (D01) | 1-3 | Earlier selected-option probability | 1,273 | 692 | 54.36 | 0.0149 | 0.0000 | 0.0700 |
| JEV 1.13 (D01) | 1-3 | Normalized vector total variation | 1,273 | 661 | 51.92 | 0.0163 | 0.0000 | 0.0724 |
| JEV 1.13 (D01) | 2-3 | Native selected-answer probability | 1,273 | 673 | 52.87 | 0.0146 | 0.0000 | 0.0700 |
| JEV 1.13 (D01) | 2-3 | Earlier selected-option probability | 1,273 | 673 | 52.87 | 0.0153 | 0.0000 | 0.0700 |
| JEV 1.13 (D01) | 2-3 | Normalized vector total variation | 1,273 | 655 | 51.45 | 0.0168 | 0.0000 | 0.0700 |

| Model (ID) | Runs | Measure | Eligible pairs | Exactly unchanged n | Exactly unchanged (%) | Mean absolute change | Median | p95 |
|---|---|---|---|---|---|---|---|---|
| GPT-6 Luna Decisions (D02) | 1-2 | Native selected-answer probability | 1,271 | 1,271 | 100.00 | 0.0000 | 0.0000 | 0.0000 |
| GPT-6 Luna Decisions (D02) | 1-2 | Earlier selected-option probability | 1,271 | 1,271 | 100.00 | 0.0000 | 0.0000 | 0.0000 |
| GPT-6 Luna Decisions (D02) | 1-2 | Normalized vector total variation | 1,271 | 1,271 | 100.00 | 0.0000 | 0.0000 | 0.0000 |
| GPT-6 Luna Decisions (D02) | 1-3 | Native selected-answer probability | 1,271 | 1,271 | 100.00 | 0.0000 | 0.0000 | 0.0000 |
| GPT-6 Luna Decisions (D02) | 1-3 | Earlier selected-option probability | 1,271 | 1,271 | 100.00 | 0.0000 | 0.0000 | 0.0000 |
| GPT-6 Luna Decisions (D02) | 1-3 | Normalized vector total variation | 1,271 | 1,271 | 100.00 | 0.0000 | 0.0000 | 0.0000 |
| GPT-6 Luna Decisions (D02) | 2-3 | Native selected-answer probability | 1,271 | 1,271 | 100.00 | 0.0000 | 0.0000 | 0.0000 |

| Model (ID) | Runs | Measure | Eligible pairs | Exactly unchanged n | Exactly unchanged (%) | Mean absolute change | Median | p95 |
|---|---|---|---|---|---|---|---|---|
| GPT-6 Luna Decisions (D02) | 2-3 | Earlier selected-option probability | 1,271 | 1,271 | 100.00 | 0.0000 | 0.0000 | 0.0000 |
| GPT-6 Luna Decisions (D02) | 2-3 | Normalized vector total variation | 1,271 | 1,271 | 100.00 | 0.0000 | 0.0000 | 0.0000 |
| Kev 4B (D03) | 1-2 | Native selected-answer probability | 1,273 | 1,085 | 85.23 | 0.0006 | 0.0000 | 0.0039 |
| Kev 4B (D03) | 1-2 | Earlier selected-option probability | 1,273 | 1,085 | 85.23 | 0.0006 | 0.0000 | 0.0039 |
| Kev 4B (D03) | 1-2 | Normalized vector total variation | 1,273 | 1,083 | 85.07 | 0.0007 | 0.0000 | 0.0050 |
| Kev 4B (D03) | 1-3 | Native selected-answer probability | 1,273 | 1,089 | 85.55 | 0.0005 | 0.0000 | 0.0040 |
| Kev 4B (D03) | 1-3 | Earlier selected-option probability | 1,273 | 1,089 | 85.55 | 0.0005 | 0.0000 | 0.0040 |

| Model (ID) | Runs | Measure | Eligible pairs | Exactly unchanged n | Exactly unchanged (%) | Mean absolute change | Median | p95 |
|---|---|---|---|---|---|---|---|---|
| Kev 4B (D03) | 1-3 | Normalized vector total variation | 1,273 | 1,088 | 85.47 | 0.0007 | 0.0000 | 0.0050 |
| Kev 4B (D03) | 2-3 | Native selected-answer probability | 1,273 | 1,070 | 84.05 | 0.0006 | 0.0000 | 0.0044 |
| Kev 4B (D03) | 2-3 | Earlier selected-option probability | 1,273 | 1,070 | 84.05 | 0.0006 | 0.0000 | 0.0044 |
| Kev 4B (D03) | 2-3 | Normalized vector total variation | 1,273 | 1,067 | 83.82 | 0.0007 | 0.0000 | 0.0053 |
| Solar Decide (D04) | 1-2 | Native selected-answer probability | 1,273 | 18 | 1.41 | 0.0335 | 0.0031 | 0.1993 |
| Solar Decide (D04) | 1-2 | Earlier selected-option probability | 1,273 | 18 | 1.41 | 0.0546 | 0.0031 | 0.3831 |
| Solar Decide (D04) | 1-2 | Normalized vector total variation | 1,273 | 17 | 1.34 | 0.0595 | 0.0035 | 0.4284 |

| Model (ID) | Runs | Measure | Eligible pairs | Exactly unchanged n | Exactly unchanged (%) | Mean absolute change | Median | p95 |
|---|---|---|---|---|---|---|---|---|
| Solar Decide (D04) | 1-3 | Native selected-answer probability | 1,273 | 24 | 1.89 | 0.0313 | 0.0033 | 0.1905 |
| Solar Decide (D04) | 1-3 | Earlier selected-option probability | 1,273 | 24 | 1.89 | 0.0563 | 0.0034 | 0.3855 |
| Solar Decide (D04) | 1-3 | Normalized vector total variation | 1,273 | 24 | 1.89 | 0.0604 | 0.0038 | 0.4180 |
| Solar Decide (D04) | 2-3 | Native selected-answer probability | 1,273 | 24 | 1.89 | 0.0332 | 0.0034 | 0.1937 |
| Solar Decide (D04) | 2-3 | Earlier selected-option probability | 1,273 | 24 | 1.89 | 0.0596 | 0.0034 | 0.4043 |
| Solar Decide (D04) | 2-3 | Normalized vector total variation | 1,273 | 23 | 1.81 | 0.0628 | 0.0036 | 0.4536 |
| Mercury Decide (D05) | 1-2 | Native selected-answer probability | 1,273 | 291 | 22.86 | 0.0106 | 0.0004 | 0.0633 |

| Model (ID) | Runs | Measure | Eligible pairs | Exactly unchanged n | Exactly unchanged (%) | Mean absolute change | Median | p95 |
|---|---|---|---|---|---|---|---|---|
| Mercury Decide (D05) | 1-2 | Earlier selected-option probability | 1,273 | 291 | 22.86 | 0.0116 | 0.0004 | 0.0633 |
| Mercury Decide (D05) | 1-2 | Normalized vector total variation | 1,273 | 291 | 22.86 | 0.0124 | 0.0004 | 0.0636 |
| Mercury Decide (D05) | 1-3 | Native selected-answer probability | 1,273 | 315 | 24.74 | 0.0108 | 0.0004 | 0.0634 |
| Mercury Decide (D05) | 1-3 | Earlier selected-option probability | 1,273 | 315 | 24.74 | 0.0122 | 0.0004 | 0.0648 |
| Mercury Decide (D05) | 1-3 | Normalized vector total variation | 1,273 | 315 | 24.74 | 0.0129 | 0.0004 | 0.0676 |
| Mercury Decide (D05) | 2-3 | Native selected-answer probability | 1,273 | 382 | 30.01 | 0.0099 | 0.0003 | 0.0616 |
| Mercury Decide (D05) | 2-3 | Earlier selected-option probability | 1,273 | 382 | 30.01 | 0.0117 | 0.0003 | 0.0628 |

| Model (ID) | Runs | Measure | Eligible pairs | Exactly unchanged n | Exactly unchanged (%) | Mean absolute change | Median | p95 |
|---|---|---|---|---|---|---|---|---|
| Mercury Decide (D05) | 2-3 | Normalized vector total variation | 1,273 | 382 | 30.01 | 0.0124 | 0.0003 | 0.0649 |
| Clef (D06) | 1-2 | Native selected-answer probability | 1,273 | 1,273 | 100.00 | 0.0000 | 0.0000 | 0.0000 |
| Clef (D06) | 1-2 | Earlier selected-option probability | 1,273 | 1,273 | 100.00 | 0.0000 | 0.0000 | 0.0000 |
| Clef (D06) | 1-2 | Normalized vector total variation | 1,273 | 1,273 | 100.00 | 0.0000 | 0.0000 | 0.0000 |
| Clef (D06) | 1-3 | Native selected-answer probability | 1,273 | 1,273 | 100.00 | 0.0000 | 0.0000 | 0.0000 |
| Clef (D06) | 1-3 | Earlier selected-option probability | 1,273 | 1,273 | 100.00 | 0.0000 | 0.0000 | 0.0000 |
| Clef (D06) | 1-3 | Normalized vector total variation | 1,273 | 1,273 | 100.00 | 0.0000 | 0.0000 | 0.0000 |

| Model (ID) | Runs | Measure | Eligible pairs | Exactly unchanged n | Exactly unchanged (%) | Mean absolute change | Median | p95 |
|---|---|---|---|---|---|---|---|---|
| Clef (D06) | 2-3 | Native selected-answer probability | 1,273 | 1,273 | 100.00 | 0.0000 | 0.0000 | 0.0000 |
| Clef (D06) | 2-3 | Earlier selected-option probability | 1,273 | 1,273 | 100.00 | 0.0000 | 0.0000 | 0.0000 |
| Clef (D06) | 2-3 | Normalized vector total variation | 1,273 | 1,273 | 100.00 | 0.0000 | 0.0000 | 0.0000 |
| Clef Flash (D07) | 1-2 | Native selected-answer probability | 1,273 | 1,273 | 100.00 | 0.0000 | 0.0000 | 0.0000 |
| Clef Flash (D07) | 1-2 | Earlier selected-option probability | 1,273 | 1,273 | 100.00 | 0.0000 | 0.0000 | 0.0000 |
| Clef Flash (D07) | 1-2 | Normalized vector total variation | 1,273 | 1,273 | 100.00 | 0.0000 | 0.0000 | 0.0000 |
| Clef Flash (D07) | 1-3 | Native selected-answer probability | 1,273 | 1,273 | 100.00 | 0.0000 | 0.0000 | 0.0000 |

| Model (ID) | Runs | Measure | Eligible pairs | Exactly unchanged n | Exactly unchanged (%) | Mean absolute change | Median | p95 |
|---|---|---|---|---|---|---|---|---|
| Clef Flash (D07) | 1-3 | Earlier selected-option probability | 1,273 | 1,273 | 100.00 | 0.0000 | 0.0000 | 0.0000 |
| Clef Flash (D07) | 1-3 | Normalized vector total variation | 1,273 | 1,273 | 100.00 | 0.0000 | 0.0000 | 0.0000 |
| Clef Flash (D07) | 2-3 | Native selected-answer probability | 1,273 | 1,273 | 100.00 | 0.0000 | 0.0000 | 0.0000 |
| Clef Flash (D07) | 2-3 | Earlier selected-option probability | 1,273 | 1,273 | 100.00 | 0.0000 | 0.0000 | 0.0000 |
| Clef Flash (D07) | 2-3 | Normalized vector total variation | 1,273 | 1,273 | 100.00 | 0.0000 | 0.0000 | 0.0000 |
| Liquid d1 (D08) | 1-2 | Native selected-answer probability | 1,273 | 888 | 69.76 | 0.0101 | 0.0000 | 0.0674 |
| Liquid d1 (D08) | 1-2 | Earlier selected-option probability | 1,273 | 888 | 69.76 | 0.0116 | 0.0000 | 0.0758 |

| Model (ID) | Runs | Measure | Eligible pairs | Exactly unchanged n | Exactly unchanged (%) | Mean absolute change | Median | p95 |
|---|---|---|---|---|---|---|---|---|
| Liquid d1 (D08) | 1-2 | Normalized vector total variation | 1,273 | 888 | 69.76 | 0.0127 | 0.0000 | 0.0898 |
| Liquid d1 (D08) | 1-3 | Native selected-answer probability | 1,273 | 871 | 68.42 | 0.0108 | 0.0000 | 0.0753 |
| Liquid d1 (D08) | 1-3 | Earlier selected-option probability | 1,273 | 871 | 68.42 | 0.0131 | 0.0000 | 0.0868 |
| Liquid d1 (D08) | 1-3 | Normalized vector total variation | 1,273 | 871 | 68.42 | 0.0142 | 0.0000 | 0.0945 |
| Liquid d1 (D08) | 2-3 | Native selected-answer probability | 1,273 | 1,114 | 87.51 | 0.0034 | 0.0000 | 0.0124 |
| Liquid d1 (D08) | 2-3 | Earlier selected-option probability | 1,273 | 1,114 | 87.51 | 0.0044 | 0.0000 | 0.0139 |
| Liquid d1 (D08) | 2-3 | Normalized vector total variation | 1,273 | 1,114 | 87.51 | 0.0049 | 0.0000 | 0.0165 |

| Model (ID) | Runs | Measure | Eligible pairs | Exactly unchanged n | Exactly unchanged (%) | Mean absolute change | Median | p95 |
|---|---|---|---|---|---|---|---|---|
| Decider V1.1 27B (D09) | 1-2 | Native selected-answer probability | 1,273 | 1,273 | 100.00 | 0.0000 | 0.0000 | 0.0000 |
| Decider V1.1 27B (D09) | 1-2 | Earlier selected-option probability | 1,273 | 1,273 | 100.00 | 0.0000 | 0.0000 | 0.0000 |
| Decider V1.1 27B (D09) | 1-2 | Normalized vector total variation | 1,273 | 1,273 | 100.00 | 0.0000 | 0.0000 | 0.0000 |
| Decider V1.1 27B (D09) | 1-3 | Native selected-answer probability | 1,273 | 1,273 | 100.00 | 0.0000 | 0.0000 | 0.0000 |
| Decider V1.1 27B (D09) | 1-3 | Earlier selected-option probability | 1,273 | 1,273 | 100.00 | 0.0000 | 0.0000 | 0.0000 |
| Decider V1.1 27B (D09) | 1-3 | Normalized vector total variation | 1,273 | 1,273 | 100.00 | 0.0000 | 0.0000 | 0.0000 |
| Decider V1.1 27B (D09) | 2-3 | Native selected-answer probability | 1,273 | 1,273 | 100.00 | 0.0000 | 0.0000 | 0.0000 |

| Model (ID) | Runs | Measure | Eligible pairs | Exactly unchanged n | Exactly unchanged (%) | Mean absolute change | Median | p95 |
|---|---|---|---|---|---|---|---|---|
| Decider V1.1 27B (D09) | 2-3 | Earlier selected-option probability | 1,273 | 1,273 | 100.00 | 0.0000 | 0.0000 | 0.0000 |
| Decider V1.1 27B (D09) | 2-3 | Normalized vector total variation | 1,273 | 1,273 | 100.00 | 0.0000 | 0.0000 | 0.0000 |
| Solar Decide Flash (D10) | 1-2 | Native selected-answer probability | 1,273 | 19 | 1.49 | 0.0205 | 0.0065 | 0.0858 |
| Solar Decide Flash (D10) | 1-2 | Earlier selected-option probability | 1,273 | 19 | 1.49 | 0.0212 | 0.0065 | 0.0890 |
| Solar Decide Flash (D10) | 1-2 | Normalized vector total variation | 1,273 | 18 | 1.41 | 0.0234 | 0.0082 | 0.0961 |
| Solar Decide Flash (D10) | 1-3 | Native selected-answer probability | 1,273 | 20 | 1.57 | 0.0206 | 0.0063 | 0.0871 |
| Solar Decide Flash (D10) | 1-3 | Earlier selected-option probability | 1,273 | 19 | 1.49 | 0.0218 | 0.0065 | 0.0931 |

| Model (ID) | Runs | Measure | Eligible pairs | Exactly unchanged n | Exactly unchanged (%) | Mean absolute change | Median | p95 |
|---|---|---|---|---|---|---|---|---|
| Solar Decide Flash (D10) | 1-3 | Normalized vector total variation | 1,273 | 17 | 1.34 | 0.0237 | 0.0071 | 0.0969 |
| Solar Decide Flash (D10) | 2-3 | Native selected-answer probability | 1,273 | 28 | 2.20 | 0.0197 | 0.0058 | 0.0798 |
| Solar Decide Flash (D10) | 2-3 | Earlier selected-option probability | 1,273 | 28 | 2.20 | 0.0208 | 0.0059 | 0.0899 |
| Solar Decide Flash (D10) | 2-3 | Normalized vector total variation | 1,273 | 25 | 1.96 | 0.0231 | 0.0074 | 0.0958 |
| Clef Omni (D11) | 1-2 | Native selected-answer probability | 1,273 | 1,267 | 99.53 | 0.0001 | 0.0000 | 0.0000 |
| Clef Omni (D11) | 1-2 | Earlier selected-option probability | 1,273 | 1,267 | 99.53 | 0.0002 | 0.0000 | 0.0000 |
| Clef Omni (D11) | 1-2 | Normalized vector total variation | 1,273 | 1,267 | 99.53 | 0.0003 | 0.0000 | 0.0000 |

| Model (ID) | Runs | Measure | Eligible pairs | Exactly unchanged n | Exactly unchanged (%) | Mean absolute change | Median | p95 |
|---|---|---|---|---|---|---|---|---|
| Clef Omni (D11) | 1-3 | Native selected-answer probability | 1,273 | 1,243 | 97.64 | 0.0003 | 0.0000 | 0.0000 |
| Clef Omni (D11) | 1-3 | Earlier selected-option probability | 1,273 | 1,243 | 97.64 | 0.0004 | 0.0000 | 0.0000 |
| Clef Omni (D11) | 1-3 | Normalized vector total variation | 1,273 | 1,243 | 97.64 | 0.0005 | 0.0000 | 0.0000 |
| Clef Omni (D11) | 2-3 | Native selected-answer probability | 1,273 | 1,249 | 98.11 | 0.0002 | 0.0000 | 0.0000 |
| Clef Omni (D11) | 2-3 | Earlier selected-option probability | 1,273 | 1,249 | 98.11 | 0.0002 | 0.0000 | 0.0000 |
| Clef Omni (D11) | 2-3 | Normalized vector total variation | 1,273 | 1,249 | 98.11 | 0.0003 | 0.0000 | 0.0000 |
| Drex v1.5 (D12) | 1-2 | Native selected-answer probability | 1,273 | 609 | 47.84 | 0.0170 | 0.0002 | 0.0977 |

| Model (ID) | Runs | Measure | Eligible pairs | Exactly unchanged n | Exactly unchanged (%) | Mean absolute change | Median | p95 |
|---|---|---|---|---|---|---|---|---|
| Drex v1.5 (D12) | 1-2 | Earlier selected-option probability | 1,273 | 609 | 47.84 | 0.0190 | 0.0002 | 0.1130 |
| Drex v1.5 (D12) | 1-2 | Normalized vector total variation | 1,273 | 603 | 47.37 | 0.0205 | 0.0004 | 0.1150 |
| Drex v1.5 (D12) | 1-3 | Native selected-answer probability | 1,273 | 611 | 48.00 | 0.0168 | 0.0002 | 0.0971 |
| Drex v1.5 (D12) | 1-3 | Earlier selected-option probability | 1,273 | 611 | 48.00 | 0.0188 | 0.0002 | 0.1114 |
| Drex v1.5 (D12) | 1-3 | Normalized vector total variation | 1,273 | 605 | 47.53 | 0.0204 | 0.0004 | 0.1141 |
| Drex v1.5 (D12) | 2-3 | Native selected-answer probability | 1,273 | 1,256 | 98.66 | 0.0003 | 0.0000 | 0.0000 |
| Drex v1.5 (D12) | 2-3 | Earlier selected-option probability | 1,273 | 1,256 | 98.66 | 0.0003 | 0.0000 | 0.0000 |

| Model (ID) | Runs | Measure | Eligible pairs | Exactly unchanged n | Exactly unchanged (%) | Mean absolute change | Median | p95 |
|---|---|---|---|---|---|---|---|---|
| Drex v1.5 (D12) | 2-3 | Normalized vector total variation | 1,273 | 1,256 | 98.66 | 0.0003 | 0.0000 | 0.0000 |
| Microsoft-Decision-1 (D13) | 1-2 | Native selected-answer probability | 1,273 | 145 | 11.39 | 0.0218 | 0.0048 | 0.1031 |
| Microsoft-Decision-1 (D13) | 1-2 | Earlier selected-option probability | 1,273 | 145 | 11.39 | 0.0234 | 0.0048 | 0.1105 |
| Microsoft-Decision-1 (D13) | 1-2 | Normalized vector total variation | 1,273 | 142 | 11.15 | 0.0256 | 0.0059 | 0.1156 |
| Microsoft-Decision-1 (D13) | 1-3 | Native selected-answer probability | 1,273 | 143 | 11.23 | 0.0209 | 0.0050 | 0.1007 |
| Microsoft-Decision-1 (D13) | 1-3 | Earlier selected-option probability | 1,273 | 143 | 11.23 | 0.0218 | 0.0050 | 0.1062 |
| Microsoft-Decision-1 (D13) | 1-3 | Normalized vector total variation | 1,273 | 143 | 11.23 | 0.0241 | 0.0058 | 0.1114 |

| Model (ID) | Runs | Measure | Eligible pairs | Exactly unchanged n | Exactly unchanged (%) | Mean absolute change | Median | p95 |
|---|---|---|---|---|---|---|---|---|
| Microsoft-Decision-1 (D13) | 2-3 | Native selected-answer probability | 1,273 | 161 | 12.65 | 0.0216 | 0.0047 | 0.1013 |
| Microsoft-Decision-1 (D13) | 2-3 | Earlier selected-option probability | 1,273 | 161 | 12.65 | 0.0230 | 0.0047 | 0.1055 |
| Microsoft-Decision-1 (D13) | 2-3 | Normalized vector total variation | 1,273 | 161 | 12.65 | 0.0254 | 0.0058 | 0.1129 |

**Supplementary Table S9. Original-first-attempt and visual-nomination accuracy sensitivities.** Primary and repeat runs were assessed separately. Full and original-first-attempt accuracy were calculated using all 1,273 questions, with failures counted as incorrect. Recovery change was calculated as the difference between first-valid and first-attempt accuracy within the same run. Visual filtering excluded the fixed 108 questions nominated post hoc, leaving 1,165 questions per run. Among primary configurations, recovery increased accuracy for 15 models by 0.08-5.26 percentage points, while visual filtering changed accuracy by -0.04 to 1.08 points. The questions have not been assessed whether the image is actually required to provide answer a question.

| Model (ID) | Run | Full n | First-valid accuracy (%) | First-attempt accuracy (%) | Recovery change, pp | Visual-filtered n | Visual-filtered accuracy (%) |
|---|---|---|---|---|---|---|---|
| GPT-6.1 Sol Low (G01) | 1 | 1,273 | 96.86 | 96.86 | 0.00 | 1,165 | 96.91 |
| GPT-6 Luna Low (G02) | 1 | 1,273 | 94.19 | 94.19 | 0.00 | 1,165 | 94.59 |

| Model (ID) | Run | Full n | First-valid accuracy (%) | First-attempt accuracy (%) | Recovery change, pp | Visual-filtered n | Visual-filtered accuracy (%) |
|---|---|---|---|---|---|---|---|
| GPT-6 Luna High (G03) | 1 | 1,273 | 95.21 | 95.21 | 0.00 | 1,165 | 95.54 |
| Claude Sonnet 5.5 Low (G04) | 1 | 1,273 | 95.13 | 94.42 | 0.71 | 1,165 | 95.28 |
| Gemini 3.8 Flash Low (G05) | 1 | 1,273 | 96.78 | 96.70 | 0.08 | 1,165 | 96.82 |
| DeepSeek V4.1 Flash Low (G06) | 1 | 1,273 | 95.52 | 92.93 | 2.59 | 1,165 | 95.62 |
| DeepSeek V4.1 Flash High (G07) | 1 | 1,273 | 94.58 | 92.69 | 1.89 | 1,165 | 94.76 |
| Qwen3.8 Max Low (G08) | 1 | 1,273 | 89.63 | 89.63 | 0.00 | 1,165 | 89.96 |
| Qwen3.8-27B Low (G09) | 1 | 1,273 | 94.11 | 94.11 | 0.00 | 1,165 | 94.42 |
| Qwen3.8-27B Xhigh (G10) | 1 | 1,273 | 95.44 | 93.79 | 1.65 | 1,165 | 95.79 |
| Mistral Medium 3.5 None (G11) | 1 | 1,273 | 78.32 | 78.16 | 0.16 | 1,165 | 78.28 |
| GLM-5 Native (G12) | 1 | 1,273 | 94.97 | 89.71 | 5.26 | 1,165 | 95.28 |

| Model (ID) | Run | Full n | First-valid accuracy (%) | First-attempt accuracy (%) | Recovery change, pp | Visual-filtered n | Visual-filtered accuracy (%) |
|---|---|---|---|---|---|---|---|
| Kimi K2.5 Native (G13) | 1 | 1,273 | 94.50 | 94.34 | 0.16 | 1,165 | 94.76 |
| Llama 4 Maverick Standard (G14) | 1 | 1,273 | 83.82 | 83.82 | 0.00 | 1,165 | 83.95 |
| Kimi K2.5 Instant (G15) | 1 | 1,273 | 85.00 | 85.00 | 0.00 | 1,165 | 85.49 |
| JEV 1.13 (D01) | 1 | 1,273 | 88.30 | 88.30 | 0.00 | 1,165 | 89.10 |
| GPT-6 Luna Decisions (D02) | 1 | 1,273 | 89.08 | 88.92 | 0.16 | 1,165 | 89.36 |
| Kev 4B (D03) | 1 | 1,273 | 67.71 | 67.71 | 0.00 | 1,165 | 68.15 |
| Solar Decide (D04) | 1 | 1,273 | 82.64 | 82.64 | 0.00 | 1,165 | 82.83 |
| Mercury Decide (D05) | 1 | 1,273 | 85.23 | 85.23 | 0.00 | 1,165 | 85.67 |
| Clef (D06) | 1 | 1,273 | 84.76 | 84.60 | 0.16 | 1,165 | 85.24 |
| Clef Flash (D07) | 1 | 1,273 | 73.68 | 73.61 | 0.08 | 1,165 | 74.76 |
| Liquid d1 (D08) | 1 | 1,273 | 88.37 | 88.37 | 0.00 | 1,165 | 88.67 |
| Decider V1.1 27B (D09) | 1 | 1,273 | 87.82 | 87.82 | 0.00 | 1,165 | 88.41 |

| Model (ID) | Run | Full n | First-valid accuracy (%) | First-attempt accuracy (%) | Recovery change, pp | Visual-filtered n | Visual-filtered accuracy (%) |
|---|---|---|---|---|---|---|---|
| Solar Decide Flash (D10) | 1 | 1,273 | 72.90 | 72.82 | 0.08 | 1,165 | 72.96 |
| Clef Omni (D11) | 1 | 1,273 | 79.42 | 78.71 | 0.71 | 1,165 | 80.00 |
| Drex v1.5 (D12) | 1 | 1,273 | 73.53 | 73.45 | 0.08 | 1,165 | 74.16 |
| Microsoft-Decision-1 (D13) | 1 | 1,273 | 79.58 | 79.42 | 0.16 | 1,165 | 80.43 |
| JEV 1.13 (D01) | 2 | 1,273 | 87.98 | 87.98 | 0.00 | 1,165 | 88.84 |
| GPT-6 Luna Decisions (D02) | 2 | 1,273 | 89.08 | 89.08 | 0.00 | 1,165 | 89.36 |
| Kev 4B (D03) | 2 | 1,273 | 67.71 | 67.24 | 0.47 | 1,165 | 68.15 |
| Solar Decide (D04) | 2 | 1,273 | 82.33 | 82.25 | 0.08 | 1,165 | 82.66 |
| Mercury Decide (D05) | 2 | 1,273 | 85.86 | 85.86 | 0.00 | 1,165 | 86.27 |
| Clef (D06) | 2 | 1,273 | 84.76 | 84.76 | 0.00 | 1,165 | 85.24 |
| Clef Flash (D07) | 2 | 1,273 | 73.68 | 73.21 | 0.47 | 1,165 | 74.76 |
| Liquid d1 (D08) | 2 | 1,273 | 88.77 | 88.77 | 0.00 | 1,165 | 88.93 |

| Model (ID) | Run | Full n | First-valid accuracy (%) | First-attempt accuracy (%) | Recovery change, pp | Visual-filtered n | Visual-filtered accuracy (%) |
|---|---|---|---|---|---|---|---|
| Decider V1.1 27B (D09) | 2 | 1,273 | 87.82 | 87.82 | 0.00 | 1,165 | 88.41 |
| Solar Decide Flash (D10) | 2 | 1,273 | 72.98 | 72.98 | 0.00 | 1,165 | 73.05 |
| Clef Omni (D11) | 2 | 1,273 | 79.50 | 79.42 | 0.08 | 1,165 | 80.09 |
| Drex v1.5 (D12) | 2 | 1,273 | 73.13 | 73.13 | 0.00 | 1,165 | 73.82 |
| Microsoft-Decision-1 (D13) | 2 | 1,273 | 79.81 | 79.81 | 0.00 | 1,165 | 80.60 |
| JEV 1.13 (D01) | 3 | 1,273 | 88.14 | 88.14 | 0.00 | 1,165 | 89.01 |
| GPT-6 Luna Decisions (D02) | 3 | 1,273 | 89.08 | 89.08 | 0.00 | 1,165 | 89.36 |
| Kev 4B (D03) | 3 | 1,273 | 67.71 | 67.64 | 0.08 | 1,165 | 68.15 |
| Solar Decide (D04) | 3 | 1,273 | 82.80 | 82.80 | 0.00 | 1,165 | 82.92 |
| Mercury Decide (D05) | 3 | 1,273 | 85.55 | 85.55 | 0.00 | 1,165 | 85.92 |
| Clef (D06) | 3 | 1,273 | 84.76 | 84.68 | 0.08 | 1,165 | 85.24 |
| Clef Flash (D07) | 3 | 1,273 | 73.68 | 73.61 | 0.08 | 1,165 | 74.76 |

| Model (ID) | Run | Full n | First-valid accuracy (%) | First-attempt accuracy (%) | Recovery change, pp | Visual-filtered n | Visual-filtered accuracy (%) |
|---|---|---|---|---|---|---|---|
| Liquid d1 (D08) | 3 | 1,273 | 88.53 | 88.53 | 0.00 | 1,165 | 88.67 |
| Decider V1.1 27B (D09) | 3 | 1,273 | 87.82 | 87.82 | 0.00 | 1,165 | 88.41 |
| Solar Decide Flash (D10) | 3 | 1,273 | 72.51 | 72.51 | 0.00 | 1,165 | 72.70 |
| Clef Omni (D11) | 3 | 1,273 | 79.50 | 79.50 | 0.00 | 1,165 | 80.09 |
| Drex v1.5 (D12) | 3 | 1,273 | 73.13 | 73.13 | 0.00 | 1,165 | 73.82 |
| Microsoft-Decision-1 (D13) | 3 | 1,273 | 79.73 | 79.73 | 0.00 | 1,165 | 80.69 |

**Supplementary Table S10. Complete-valid and visual-filtered paired accuracy sensitivities.** Complete-valid comparisons included only questions with valid selected answers from both configurations. Therefore, denominators varied between pairs. Visual filtering used the same 1,165 questions for every comparison and continued to count failures as incorrect. Effects are expressed as A-B in percentage points, with pointwise 95% question-bootstrap intervals. Complete-valid denominators ranged from 1,260 to 1,273. The JEV-Luna Decisions difference was -0.87 points among 1,271 complete-valid pairs and -0.26 points among the 1,165 visual-filtered questions. The questions have not been assessed whether the image is actually required to provide answer a question.

| **Family** | **Model A - Model B** | **Complete -valid n** | **Complete -valid effect, pp** | **Complete -valid 95% CI, pp** | **Visual-filtered n** | **Visual-filtered effect, pp** | **Visual-filtered 95% CI, pp** |
|---|---|---|---|---|---|---|---|
| S1 | JEV 1.13 (D01) - GPT-6.1 | 1,273 | -8.56 | -10.21 to -6.91 | 1,165 | -7.81 | -9.53 to -6.18 |

| Family | Model A - Model B | Complete-valid n | Complete-valid effect, pp | Complete-valid 95% CI, pp | Visual-filtered n | Visual-filtered effect, pp | Visual-filtered 95% CI, pp |
|---|---|---|---|---|---|---|---|
| | Sol Low (G01) | | | | | | |
| S1 | JEV 1.13 (D01) - GPT-6 Luna Low (G02) | 1,273 | -5.89 | -7.62 to -4.16 | 1,165 | -5.49 | -7.21 to -3.78 |
| S1 | JEV 1.13 (D01) - GPT-6 Luna High (G03) | 1,273 | -6.91 | -8.64 to -5.18 | 1,165 | -6.44 | -8.24 to -4.72 |
| S1 | JEV 1.13 (D01) - Claude Sonnet 5.5 Low (G04) | 1,272 | -6.92 | -8.58 to -5.26 | 1,165 | -6.18 | -7.90 to -4.46 |
| S1 | JEV 1.13 (D01) - Gemini 3.8 Flash Low (G05) | 1,273 | -8.48 | -10.21 to -6.83 | 1,165 | -7.73 | -9.44 to -6.01 |
| S1 | JEV 1.13 (D01) - DeepSeek V4.1 Flash Low (G06) | 1,273 | -7.23 | -8.88 to -5.58 | 1,165 | -6.52 | -8.24 to -4.81 |
| S1 | JEV 1.13 (D01) - DeepSeek V4.1 Flash High (G07) | 1,273 | -6.28 | -7.93 to -4.56 | 1,165 | -5.67 | -7.38 to -3.95 |

| Family | Model A - Model B | Complete-valid n | Complete-valid effect, pp | Complete-valid 95% CI, pp | Visual-filtered n | Visual-filtered effect, pp | Visual-filtered 95% CI, pp |
|---|---|---|---|---|---|---|---|
| S1 | JEV 1.13 (D01) - Qwen3.8 Max Low (G08) | 1,273 | -1.34 | -3.06 to 0.39 | 1,165 | -0.86 | -2.66 to 0.94 |
| S1 | JEV 1.13 (D01) - Qwen3.8-27B Low (G09) | 1,273 | -5.81 | -7.62 to -4.01 | 1,165 | -5.32 | -7.12 to -3.52 |
| S1 | JEV 1.13 (D01) - Qwen3.8-27B Xhigh (G10) | 1,272 | -7.15 | -8.88 to -5.42 | 1,165 | -6.70 | -8.41 to -4.98 |
| S1 | JEV 1.13 (D01) - Mistral Medium 3.5 None (G11) | 1,273 | 9.98 | 7.86 to 12.10 | 1,165 | 10.82 | 8.58 to 13.13 |
| S1 | JEV 1.13 (D01) - GLM-5 Native (G12) | 1,262 | -7.05 | -8.78 to -5.32 | 1,165 | -6.18 | -7.98 to -4.46 |
| S1 | JEV 1.13 (D01) - Kimi K2.5 Native (G13) | 1,273 | -6.21 | -7.93 to -4.48 | 1,165 | -5.67 | -7.38 to -3.95 |
| S1 | JEV 1.13 (D01) - Llama 4 Maverick | 1,273 | 4.48 | 2.51 to 6.52 | 1,165 | 5.15 | 3.09 to 7.21 |

| Family | Model A - Model B | Complete -valid n | Complete -valid effect, pp | Complete -valid 95% CI, pp | Visual-filtered n | Visual-filtered effect, pp | Visual-filtered 95% CI, pp |
|---|---|---|---|---|---|---|---|
| | Standard (G14) | | | | | | |
| S1 | JEV 1.13 (D01) - Kimi K2.5 Instant (G15) | 1,273 | 3.30 | 1.41 to 5.26 | 1,165 | 3.61 | 1.55 to 5.67 |
| S1 | GPT-6 Luna Decisions (D02) - GPT-6.1 Sol Low (G01) | 1,271 | -7.63 | -9.28 to -6.06 | 1,165 | -7.55 | -9.27 to -5.84 |
| S1 | GPT-6 Luna Decisions (D02) - GPT-6 Luna Low (G02) | 1,271 | -4.96 | -6.46 to -3.46 | 1,165 | -5.24 | -6.87 to -3.61 |
| S1 | GPT-6 Luna Decisions (D02) - GPT-6 Luna High (G03) | 1,271 | -6.06 | -7.63 to -4.56 | 1,165 | -6.18 | -7.81 to -4.64 |
| S1 | GPT-6 Luna Decisions (D02) - Claude Sonnet | 1,270 | -6.06 | -7.66 to -4.56 | 1,165 | -5.92 | -7.64 to -4.29 |

| Family | Model A - Model B | Complete-valid n | Complete-valid effect, pp | Complete-valid 95% CI, pp | Visual-filtered n | Visual-filtered effect, pp | Visual-filtered 95% CI, pp |
|---|---|---|---|---|---|---|---|
| | 5.5 Low (G04) | | | | | | |
| S1 | GPT-6 Luna Decisions (D02) - Gemini 3.8 Flash Low (G05) | 1,271 | -7.55 | -9.22 to -5.91 | 1,165 | -7.47 | -9.18 to -5.75 |
| S1 | GPT-6 Luna Decisions (D02) - DeepSeek V4.1 Flash Low (G06) | 1,271 | -6.29 | -7.97 to -4.63 | 1,165 | -6.27 | -8.07 to -4.46 |
| S1 | GPT-6 Luna Decisions (D02) - DeepSeek V4.1 Flash High (G07) | 1,271 | -5.35 | -7.01 to -3.69 | 1,165 | -5.41 | -7.21 to -3.61 |
| S1 | GPT-6 Luna Decisions (D02) - Qwen3.8 Max Low (G08) | 1,271 | -0.39 | -2.12 to 1.26 | 1,165 | -0.60 | -2.40 to 1.20 |
| S1 | GPT-6 Luna Decisions (D02) - Qwen3.8- | 1,271 | -4.88 | -6.53 to -3.23 | 1,165 | -5.06 | -6.78 to -3.35 |

| Family | Model A - Model B | Complete-valid n | Complete-valid effect, pp | Complete-valid 95% CI, pp | Visual-filtered n | Visual-filtered effect, pp | Visual-filtered 95% CI, pp |
|---|---|---|---|---|---|---|---|
| | 27B Low (G09) | | | | | | |
| S1 | GPT-6 Luna Decisions (D02) - Qwen3.8-27B Xhigh (G10) | 1,270 | -6.22 | -7.88 to -4.57 | 1,165 | -6.44 | -8.15 to -4.72 |
| S1 | GPT-6 Luna Decisions (D02) - Mistral Medium 3.5 None (G11) | 1,271 | 10.94 | 8.81 to 13.05 | 1,165 | 11.07 | 8.93 to 13.30 |
| S1 | GPT-6 Luna Decisions (D02) - GLM-5 Native (G12) | 1,260 | -6.11 | -7.80 to -4.43 | 1,165 | -5.92 | -7.73 to -4.12 |
| S1 | GPT-6 Luna Decisions (D02) - Kimi K2.5 Native (G13) | 1,271 | -5.35 | -7.08 to -3.62 | 1,165 | -5.41 | -7.21 to -3.61 |
| S1 | GPT-6 Luna Decisions (D02) - Llama 4 | 1,271 | 5.35 | 3.38 to 7.39 | 1,165 | 5.41 | 3.35 to 7.55 |

| Family | Model A - Model B | Complete -valid n | Complete -valid effect, pp | Complete -valid 95% CI, pp | Visual-filtered n | Visual-filtered effect, pp | Visual-filtered 95% CI, pp |
|---|---|---|---|---|---|---|---|
| | Maverick Standard (G14) | | | | | | |
| S1 | GPT-6 Luna Decisions (D02) - Kimi K2.5 Instant (G15) | 1,271 | 4.17 | 2.28 to 6.13 | 1,165 | 3.86 | 1.89 to 5.84 |
| S1 | Kev 4B (D03) - GPT-6.1 Sol Low (G01) | 1,273 | -29.14 | -31.74 to -26.55 | 1,165 | -28.76 | -31.50 to -26.09 |
| S1 | Kev 4B (D03) - GPT-6 Luna Low (G02) | 1,273 | -26.47 | -29.14 to -23.88 | 1,165 | -26.44 | -29.18 to -23.61 |
| S1 | Kev 4B (D03) - GPT-6 Luna High (G03) | 1,273 | -27.49 | -30.16 to -24.90 | 1,165 | -27.38 | -30.13 to -24.64 |
| S1 | Kev 4B (D03) - Claude Sonnet 5.5 Low (G04) | 1,272 | -27.44 | -30.09 to -24.78 | 1,165 | -27.12 | -29.87 to -24.29 |
| S1 | Kev 4B (D03) - Gemini | 1,273 | -29.07 | -31.66 to -26.47 | 1,165 | -28.67 | -31.33 to -26.01 |

| Family | Model A - Model B | Complete -valid n | Complete -valid effect, pp | Complete -valid 95% CI, pp | Visual-filtered n | Visual-filtered effect, pp | Visual-filtered 95% CI, pp |
|---|---|---|---|---|---|---|---|
| | 3.8 Flash Low (G05) | | | | | | |
| S1 | Kev 4B (D03) - DeepSeek V4.1 Flash Low (G06) | 1,273 | -27.81 | -30.40 to -25.22 | 1,165 | -27.47 | -30.21 to -24.72 |
| S1 | Kev 4B (D03) - DeepSeek V4.1 Flash High (G07) | 1,273 | -26.87 | -29.54 to -24.19 | 1,165 | -26.61 | -29.36 to -23.86 |
| S1 | Kev 4B (D03) - Qwen3.8 Max Low (G08) | 1,273 | -21.92 | -24.59 to -19.32 | 1,165 | -21.80 | -24.64 to -19.06 |
| S1 | Kev 4B (D03) - Qwen3.8-27B Low (G09) | 1,273 | -26.39 | -28.99 to -23.80 | 1,165 | -26.27 | -29.01 to -23.61 |
| S1 | Kev 4B (D03) - Qwen3.8-27B Xhigh (G10) | 1,272 | -27.75 | -30.42 to -25.16 | 1,165 | -27.64 | -30.30 to -24.89 |
| S1 | Kev 4B (D03) - Mistral Medium 3.5 None (G11) | 1,273 | -10.60 | -13.20 to -8.01 | 1,165 | -10.13 | -12.71 to -7.55 |

| Family | Model A - Model B | Complete-valid n | Complete-valid effect, pp | Complete-valid 95% CI, pp | Visual-filtered n | Visual-filtered effect, pp | Visual-filtered 95% CI, pp |
|---|---|---|---|---|---|---|---|
| S1 | Kev 4B (D03) - GLM-5 Native (G12) | 1,262 | -27.65 | -30.23 to -25.10 | 1,165 | -27.12 | -29.87 to -24.46 |
| S1 | Kev 4B (D03) - Kimi K2.5 Native (G13) | 1,273 | -26.79 | -29.38 to -24.12 | 1,165 | -26.61 | -29.36 to -23.86 |
| S1 | Kev 4B (D03) - Llama 4 Maverick Standard (G14) | 1,273 | -16.10 | -18.77 to -13.43 | 1,165 | -15.79 | -18.54 to -13.05 |
| S1 | Kev 4B (D03) - Kimi K2.5 Instant (G15) | 1,273 | -17.28 | -19.87 to -14.69 | 1,165 | -17.34 | -20.00 to -14.68 |
| S1 | Solar Decide (D04) - GPT-6.1 Sol Low (G01) | 1,273 | -14.22 | -16.26 to -12.25 | 1,165 | -14.08 | -16.22 to -12.02 |
| S1 | Solar Decide (D04) - GPT-6 Luna Low (G02) | 1,273 | -11.55 | -13.59 to -9.58 | 1,165 | -11.76 | -13.91 to -9.70 |
| S1 | Solar Decide | 1,273 | -12.57 | -14.69 to -10.60 | 1,165 | -12.70 | -14.85 to -10.56 |

| Family | Model A - Model B | Complete -valid n | Complete -valid effect, pp | Complete -valid 95% CI, pp | Visual-filtered n | Visual-filtered effect, pp | Visual-filtered 95% CI, pp |
|---|---|---|---|---|---|---|---|
| | (D04) - GPT-6 Luna High (G03) | | | | | | |
| S1 | Solar Decide (D04) - Claude Sonnet 5.5 Low (G04) | 1,272 | -12.58 | -14.56 to -10.61 | 1,165 | -12.45 | -14.59 to -10.39 |
| S1 | Solar Decide (D04) - Gemini 3.8 Flash Low (G05) | 1,273 | -14.14 | -16.18 to -12.10 | 1,165 | -13.99 | -16.14 to -11.85 |
| S1 | Solar Decide (D04) - DeepSeek V4.1 Flash Low (G06) | 1,273 | -12.88 | -14.85 to -11.00 | 1,165 | -12.79 | -14.94 to -10.82 |
| S1 | Solar Decide (D04) - DeepSeek V4.1 Flash High (G07) | 1,273 | -11.94 | -13.90 to -9.98 | 1,165 | -11.93 | -14.08 to -9.87 |
| S1 | Solar Decide (D04) - Qwen3.8 Max Low (G08) | 1,273 | -6.99 | -9.11 to -4.87 | 1,165 | -7.12 | -9.36 to -4.89 |

| Family | Model A - Model B | Complete-valid n | Complete-valid effect, pp | Complete-valid 95% CI, pp | Visual-filtered n | Visual-filtered effect, pp | Visual-filtered 95% CI, pp |
|---|---|---|---|---|---|---|---|
| S1 | Solar Decide (D04) - Qwen3.8-27B Low (G09) | 1,273 | -11.47 | -13.43 to -9.51 | 1,165 | -11.59 | -13.73 to -9.53 |
| S1 | Solar Decide (D04) - Qwen3.8-27B Xhigh (G10) | 1,272 | -12.81 | -14.80 to -10.85 | 1,165 | -12.96 | -15.02 to -10.90 |
| S1 | Solar Decide (D04) - Mistral Medium 3.5 None (G11) | 1,273 | 4.32 | 2.20 to 6.44 | 1,165 | 4.55 | 2.32 to 6.78 |
| S1 | Solar Decide (D04) - GLM-5 Native (G12) | 1,262 | -12.76 | -14.81 to -10.79 | 1,165 | -12.45 | -14.59 to -10.30 |
| S1 | Solar Decide (D04) - Kimi K2.5 Native (G13) | 1,273 | -11.86 | -13.90 to -9.82 | 1,165 | -11.93 | -14.08 to -9.87 |
| S1 | Solar Decide (D04) - Llama 4 Maverick | 1,273 | -1.18 | -3.38 to 1.10 | 1,165 | -1.12 | -3.43 to 1.20 |

| Family | Model A - Model B | Complete-valid n | Complete-valid effect, pp | Complete-valid 95% CI, pp | Visual-filtered n | Visual-filtered effect, pp | Visual-filtered 95% CI, pp |
|---|---|---|---|---|---|---|---|
| | Standard (G14) | | | | | | |
| S1 | Solar Decide (D04) - Kimi K2.5 Instant (G15) | 1,273 | -2.36 | -4.56 to -0.16 | 1,165 | -2.66 | -4.98 to -0.34 |
| S1 | Mercury Decide (D05) - GPT-6.1 Sol Low (G01) | 1,273 | -11.63 | -13.59 to -9.74 | 1,165 | -11.24 | -13.30 to -9.18 |
| S1 | Mercury Decide (D05) - GPT-6 Luna Low (G02) | 1,273 | -8.96 | -10.92 to -7.07 | 1,165 | -8.93 | -10.90 to -6.95 |
| S1 | Mercury Decide (D05) - GPT-6 Luna High (G03) | 1,273 | -9.98 | -11.94 to -8.09 | 1,165 | -9.87 | -11.85 to -7.90 |
| S1 | Mercury Decide (D05) - Claude Sonnet 5.5 Low (G04) | 1,272 | -9.98 | -11.89 to -8.17 | 1,165 | -9.61 | -11.59 to -7.64 |
| S1 | Mercury Decide | 1,273 | -11.55 | -13.51 to -9.58 | 1,165 | -11.16 | -13.13 to -9.18 |

| Family | Model A - Model B | Complete-valid n | Complete-valid effect, pp | Complete-valid 95% CI, pp | Visual-filtered n | Visual-filtered effect, pp | Visual-filtered 95% CI, pp |
|---|---|---|---|---|---|---|---|
| | (D05) - Gemini 3.8 Flash Low (G05) | | | | | | |
| S1 | Mercury Decide (D05) - DeepSeek V4.1 Flash Low (G06) | 1,273 | -10.29 | -12.18 to -8.48 | 1,165 | -9.96 | -11.93 to -8.07 |
| S1 | Mercury Decide (D05) - DeepSeek V4.1 Flash High (G07) | 1,273 | -9.35 | -11.23 to -7.46 | 1,165 | -9.10 | -11.16 to -7.12 |
| S1 | Mercury Decide (D05) - Qwen3.8 Max Low (G08) | 1,273 | -4.40 | -6.36 to -2.51 | 1,165 | -4.29 | -6.35 to -2.32 |
| S1 | Mercury Decide (D05) - Qwen3.8-27B Low (G09) | 1,273 | -8.88 | -10.68 to -7.15 | 1,165 | -8.76 | -10.64 to -6.95 |
| S1 | Mercury Decide (D05) - Qwen3.8-27B Xhigh (G10) | 1,272 | -10.22 | -12.04 to -8.43 | 1,165 | -10.13 | -12.02 to -8.33 |

| Family | Model A - Model B | Complete-valid n | Complete-valid effect, pp | Complete-valid 95% CI, pp | Visual-filtered n | Visual-filtered effect, pp | Visual-filtered 95% CI, pp |
|---|---|---|---|---|---|---|---|
| S1 | Mercury Decide (D05) - Mistral Medium 3.5 None (G11) | 1,273 | 6.91 | 4.71 to 9.03 | 1,165 | 7.38 | 5.15 to 9.61 |
| S1 | Mercury Decide (D05) - GLM-5 Native (G12) | 1,262 | -10.06 | -11.98 to -8.17 | 1,165 | -9.61 | -11.59 to -7.73 |
| S1 | Mercury Decide (D05) - Kimi K2.5 Native (G13) | 1,273 | -9.27 | -11.15 to -7.38 | 1,165 | -9.10 | -11.07 to -7.12 |
| S1 | Mercury Decide (D05) - Llama 4 Maverick Standard (G14) | 1,273 | 1.41 | -0.63 to 3.46 | 1,165 | 1.72 | -0.43 to 3.86 |
| S1 | Mercury Decide (D05) - Kimi K2.5 Instant (G15) | 1,273 | 0.24 | -1.81 to 2.28 | 1,165 | 0.17 | -1.97 to 2.32 |
| S1 | Clef (D06) - GPT-6.1 | 1,273 | -12.10 | -14.06 to -10.13 | 1,165 | -11.67 | -13.65 to -9.70 |

| Family | Model A - Model B | Complete-valid n | Complete-valid effect, pp | Complete-valid 95% CI, pp | Visual-filtered n | Visual-filtered effect, pp | Visual-filtered 95% CI, pp |
|---|---|---|---|---|---|---|---|
| | Sol Low (G01) | | | | | | |
| S1 | Clef (D06) - GPT-6 Luna Low (G02) | 1,273 | -9.43 | -11.39 to -7.54 | 1,165 | -9.36 | -11.33 to -7.38 |
| S1 | Clef (D06) - GPT-6 Luna High (G03) | 1,273 | -10.45 | -12.41 to -8.48 | 1,165 | -10.30 | -12.27 to -8.33 |
| S1 | Clef (D06) - Claude Sonnet 5.5 Low (G04) | 1,272 | -10.46 | -12.49 to -8.57 | 1,165 | -10.04 | -12.02 to -7.98 |
| S1 | Clef (D06) - Gemini 3.8 Flash Low (G05) | 1,273 | -12.02 | -13.98 to -10.05 | 1,165 | -11.59 | -13.56 to -9.61 |
| S1 | Clef (D06) - DeepSeek V4.1 Flash Low (G06) | 1,273 | -10.76 | -12.73 to -8.80 | 1,165 | -10.39 | -12.36 to -8.33 |
| S1 | Clef (D06) - DeepSeek V4.1 Flash High (G07) | 1,273 | -9.82 | -11.86 to -7.86 | 1,165 | -9.53 | -11.50 to -7.47 |
| S1 | Clef (D06) - Qwen3.8 Max Low (G08) | 1,273 | -4.87 | -6.83 to -2.99 | 1,165 | -4.72 | -6.78 to -2.66 |

| Family | Model A - Model B | Complete -valid n | Complete -valid effect, pp | Complete -valid 95% CI, pp | Visual-filtered n | Visual-filtered effect, pp | Visual-filtered 95% CI, pp |
|---|---|---|---|---|---|---|---|
| S1 | Clef (D06) - Qwen3.8-27B Low (G09) | 1,273 | -9.35 | -11.23 to -7.54 | 1,165 | -9.18 | -11.07 to -7.30 |
| S1 | Clef (D06) - Qwen3.8-27B Xhigh (G10) | 1,272 | -10.69 | -12.58 to -8.88 | 1,165 | -10.56 | -12.45 to -8.67 |
| S1 | Clef (D06) - Mistral Medium 3.5 None (G11) | 1,273 | 6.44 | 4.24 to 8.64 | 1,165 | 6.95 | 4.72 to 9.27 |
| S1 | Clef (D06) - GLM-5 Native (G12) | 1,262 | -10.70 | -12.66 to -8.79 | 1,165 | -10.04 | -12.10 to -7.98 |
| S1 | Clef (D06) - Kimi K2.5 Native (G13) | 1,273 | -9.74 | -11.70 to -7.78 | 1,165 | -9.53 | -11.59 to -7.47 |
| S1 | Clef (D06) - Llama 4 Maverick Standard (G14) | 1,273 | 0.94 | -1.10 to 2.99 | 1,165 | 1.29 | -0.86 to 3.43 |
| S1 | Clef (D06) - Kimi K2.5 Instant (G15) | 1,273 | -0.24 | -2.28 to 1.81 | 1,165 | -0.26 | -2.40 to 1.89 |
| S1 | Clef Flash (D07) - GPT-6.1 | 1,273 | -23.17 | -25.61 to -20.74 | 1,165 | -22.15 | -24.72 to -19.66 |

| Family | Model A - Model B | Complete-valid n | Complete-valid effect, pp | Complete-valid 95% CI, pp | Visual-filtered n | Visual-filtered effect, pp | Visual-filtered 95% CI, pp |
|---|---|---|---|---|---|---|---|
| | Sol Low (G01) | | | | | | |
| S1 | Clef Flash (D07) - GPT-6 Luna Low (G02) | 1,273 | -20.50 | -23.02 to -18.15 | 1,165 | -19.83 | -22.40 to -17.34 |
| S1 | Clef Flash (D07) - GPT-6 Luna High (G03) | 1,273 | -21.52 | -24.04 to -19.17 | 1,165 | -20.77 | -23.35 to -18.37 |
| S1 | Clef Flash (D07) - Claude Sonnet 5.5 Low (G04) | 1,272 | -21.54 | -24.00 to -19.15 | 1,165 | -20.52 | -23.09 to -18.03 |
| S1 | Clef Flash (D07) - Gemini 3.8 Flash Low (G05) | 1,273 | -23.10 | -25.53 to -20.66 | 1,165 | -22.06 | -24.64 to -19.66 |
| S1 | Clef Flash (D07) - DeepSeek V4.1 Flash Low (G06) | 1,273 | -21.84 | -24.35 to -19.48 | 1,165 | -20.86 | -23.43 to -18.37 |
| S1 | Clef Flash (D07) - DeepSeek V4.1 Flash High (G07) | 1,273 | -20.90 | -23.41 to -18.46 | 1,165 | -20.00 | -22.66 to -17.51 |

| Family | Model A - Model B | Complete-valid n | Complete-valid effect, pp | Complete-valid 95% CI, pp | Visual-filtered n | Visual-filtered effect, pp | Visual-filtered 95% CI, pp |
|---|---|---|---|---|---|---|---|
| S1 | Clef Flash (D07) - Qwen3.8 Max Low (G08) | 1,273 | -15.95 | -18.46 to -13.51 | 1,165 | -15.19 | -17.77 to -12.62 |
| S1 | Clef Flash (D07) - Qwen3.8-27B Low (G09) | 1,273 | -20.42 | -22.86 to -17.99 | 1,165 | -19.66 | -22.23 to -17.17 |
| S1 | Clef Flash (D07) - Qwen3.8-27B Xhigh (G10) | 1,272 | -21.78 | -24.23 to -19.40 | 1,165 | -21.03 | -23.61 to -18.63 |
| S1 | Clef Flash (D07) - Mistral Medium 3.5 None (G11) | 1,273 | -4.63 | -6.99 to -2.36 | 1,165 | -3.52 | -5.92 to -1.20 |
| S1 | Clef Flash (D07) - GLM-5 Native (G12) | 1,262 | -21.55 | -23.99 to -19.19 | 1,165 | -20.52 | -23.00 to -18.03 |
| S1 | Clef Flash (D07) - Kimi K2.5 Native (G13) | 1,273 | -20.82 | -23.25 to -18.46 | 1,165 | -20.00 | -22.58 to -17.51 |
| S1 | Clef Flash (D07) - Llama 4 Maverick | 1,273 | -10.13 | -12.57 to -7.70 | 1,165 | -9.18 | -11.76 to -6.70 |

| Family | Model A - Model B | Complete-valid n | Complete-valid effect, pp | Complete-valid 95% CI, pp | Visual-filtered n | Visual-filtered effect, pp | Visual-filtered 95% CI, pp |
|---|---|---|---|---|---|---|---|
| | Standard (G14) | | | | | | |
| S1 | Clef Flash (D07) - Kimi K2.5 Instant (G15) | 1,273 | -11.31 | -13.67 to -8.96 | 1,165 | -10.73 | -13.30 to -8.24 |
| S1 | Liquid d1 (D08) - GPT-6.1 Sol Low (G01) | 1,273 | -8.48 | -10.21 to -6.76 | 1,165 | -8.24 | -10.04 to -6.52 |
| S1 | Liquid d1 (D08) - GPT-6 Luna Low (G02) | 1,273 | -5.81 | -7.62 to -4.01 | 1,165 | -5.92 | -7.73 to -4.12 |
| S1 | Liquid d1 (D08) - GPT-6 Luna High (G03) | 1,273 | -6.83 | -8.64 to -5.11 | 1,165 | -6.87 | -8.67 to -5.06 |
| S1 | Liquid d1 (D08) - Claude Sonnet 5.5 Low (G04) | 1,272 | -6.84 | -8.64 to -5.11 | 1,165 | -6.61 | -8.41 to -4.81 |
| S1 | Liquid d1 (D08) - Gemini 3.8 Flash Low (G05) | 1,273 | -8.41 | -10.13 to -6.68 | 1,165 | -8.15 | -9.96 to -6.44 |

| Family | Model A - Model B | Complete-valid n | Complete-valid effect, pp | Complete-valid 95% CI, pp | Visual-filtered n | Visual-filtered effect, pp | Visual-filtered 95% CI, pp |
|---|---|---|---|---|---|---|---|
| S1 | Liquid d1 (D08) - DeepSeek V4.1 Flash Low (G06) | 1,273 | -7.15 | -8.88 to -5.50 | 1,165 | -6.95 | -8.76 to -5.24 |
| S1 | Liquid d1 (D08) - DeepSeek V4.1 Flash High (G07) | 1,273 | -6.21 | -7.93 to -4.48 | 1,165 | -6.09 | -7.90 to -4.38 |
| S1 | Liquid d1 (D08) - Qwen3.8 Max Low (G08) | 1,273 | -1.26 | -2.91 to 0.39 | 1,165 | -1.29 | -3.09 to 0.52 |
| S1 | Liquid d1 (D08) - Qwen3.8-27B Low (G09) | 1,273 | -5.73 | -7.38 to -4.08 | 1,165 | -5.75 | -7.47 to -4.12 |
| S1 | Liquid d1 (D08) - Qwen3.8-27B Xhigh (G10) | 1,272 | -7.08 | -8.74 to -5.42 | 1,165 | -7.12 | -8.84 to -5.41 |
| S1 | Liquid d1 (D08) - Mistral Medium 3.5 None (G11) | 1,273 | 10.05 | 7.93 to 12.18 | 1,165 | 10.39 | 8.15 to 12.62 |
| S1 | Liquid d1 (D08) - GLM-5 | 1,262 | -7.05 | -8.76 to -5.37 | 1,165 | -6.61 | -8.41 to -4.89 |

| Family | Model A - Model B | Complete-valid n | Complete-valid effect, pp | Complete-valid 95% CI, pp | Visual-filtered n | Visual-filtered effect, pp | Visual-filtered 95% CI, pp |
|---|---|---|---|---|---|---|---|
| | Native (G12) | | | | | | |
| S1 | Liquid d1 (D08) - Kimi K2.5 Native (G13) | 1,273 | -6.13 | -7.86 to -4.48 | 1,165 | -6.09 | -7.81 to -4.38 |
| S1 | Liquid d1 (D08) - Llama 4 Maverick Standard (G14) | 1,273 | 4.56 | 2.59 to 6.52 | 1,165 | 4.72 | 2.75 to 6.70 |
| S1 | Liquid d1 (D08) - Kimi K2.5 Instant (G15) | 1,273 | 3.38 | 1.49 to 5.34 | 1,165 | 3.18 | 1.20 to 5.15 |
| S1 | Decider V1.1 27B (D09) - GPT-6.1 Sol Low (G01) | 1,273 | -9.03 | -10.84 to -7.31 | 1,165 | -8.50 | -10.39 to -6.70 |
| S1 | Decider V1.1 27B (D09) - GPT-6 Luna Low (G02) | 1,273 | -6.36 | -8.17 to -4.56 | 1,165 | -6.18 | -8.07 to -4.38 |
| S1 | Decider V1.1 27B (D09) - GPT-6 | 1,273 | -7.38 | -9.27 to -5.58 | 1,165 | -7.12 | -9.01 to -5.32 |

| Family | Model A - Model B | Complete-valid n | Complete-valid effect, pp | Complete-valid 95% CI, pp | Visual-filtered n | Visual-filtered effect, pp | Visual-filtered 95% CI, pp |
|---|---|---|---|---|---|---|---|
| | Luna High (G03) | | | | | | |
| S1 | Decider V1.1 27B (D09) - Claude Sonnet 5.5 Low (G04) | 1,272 | -7.39 | -9.20 to -5.58 | 1,165 | -6.87 | -8.76 to -4.98 |
| S1 | Decider V1.1 27B (D09) - Gemini 3.8 Flash Low (G05) | 1,273 | -8.96 | -10.68 to -7.23 | 1,165 | -8.41 | -10.21 to -6.61 |
| S1 | Decider V1.1 27B (D09) - DeepSeek V4.1 Flash Low (G06) | 1,273 | -7.70 | -9.43 to -5.97 | 1,165 | -7.21 | -9.10 to -5.41 |
| S1 | Decider V1.1 27B (D09) - DeepSeek V4.1 Flash High (G07) | 1,273 | -6.76 | -8.48 to -5.03 | 1,165 | -6.35 | -8.24 to -4.55 |
| S1 | Decider V1.1 27B (D09) - Qwen3.8 Max Low (G08) | 1,273 | -1.81 | -3.69 to 0.00 | 1,165 | -1.55 | -3.52 to 0.43 |
| S1 | Decider V1.1 27B | 1,273 | -6.28 | -8.09 to -4.56 | 1,165 | -6.01 | -7.90 to -4.21 |

| Family | Model A - Model B | Complete-valid n | Complete-valid effect, pp | Complete-valid 95% CI, pp | Visual-filtered n | Visual-filtered effect, pp | Visual-filtered 95% CI, pp |
|---|---|---|---|---|---|---|---|
| | (D09) - Qwen3.8-27B Low (G09) | | | | | | |
| S1 | Decider V1.1 27B (D09) - Qwen3.8-27B Xhigh (G10) | 1,272 | -7.63 | -9.35 to -5.97 | 1,165 | -7.38 | -9.10 to -5.67 |
| S1 | Decider V1.1 27B (D09) - Mistral Medium 3.5 None (G11) | 1,273 | 9.51 | 7.23 to 11.78 | 1,165 | 10.13 | 7.81 to 12.45 |
| S1 | Decider V1.1 27B (D09) - GLM-5 Native (G12) | 1,262 | -7.45 | -9.21 to -5.72 | 1,165 | -6.87 | -8.67 to -5.06 |
| S1 | Decider V1.1 27B (D09) - Kimi K2.5 Native (G13) | 1,273 | -6.68 | -8.41 to -4.87 | 1,165 | -6.35 | -8.24 to -4.55 |
| S1 | Decider V1.1 27B (D09) - Llama 4 Maverick | 1,273 | 4.01 | 1.96 to 6.13 | 1,165 | 4.46 | 2.32 to 6.52 |

| Family | Model A - Model B | Complete-valid n | Complete-valid effect, pp | Complete-valid 95% CI, pp | Visual-filtered n | Visual-filtered effect, pp | Visual-filtered 95% CI, pp |
|---|---|---|---|---|---|---|---|
| | Standard (G14) | | | | | | |
| S1 | Decider V1.1 27B (D09) - Kimi K2.5 Instant (G15) | 1,273 | 2.83 | 0.79 to 4.95 | 1,165 | 2.92 | 0.77 to 5.06 |
| S1 | Solar Decide Flash (D10) - GPT-6.1 Sol Low (G01) | 1,273 | -23.96 | -26.39 to -21.52 | 1,165 | -23.95 | -26.52 to -21.46 |
| S1 | Solar Decide Flash (D10) - GPT-6 Luna Low (G02) | 1,273 | -21.29 | -23.80 to -18.85 | 1,165 | -21.63 | -24.21 to -19.06 |
| S1 | Solar Decide Flash (D10) - GPT-6 Luna High (G03) | 1,273 | -22.31 | -24.82 to -19.87 | 1,165 | -22.58 | -25.15 to -20.00 |
| S1 | Solar Decide Flash (D10) - Claude Sonnet | 1,272 | -22.33 | -24.82 to -19.92 | 1,165 | -22.32 | -24.89 to -19.83 |

| Family | Model A - Model B | Complete-valid n | Complete-valid effect, pp | Complete-valid 95% CI, pp | Visual-filtered n | Visual-filtered effect, pp | Visual-filtered 95% CI, pp |
|---|---|---|---|---|---|---|---|
| | 5.5 Low (G04) | | | | | | |
| S1 | Solar Decide Flash (D10) - Gemini 3.8 Flash Low (G05) | 1,273 | -23.88 | -26.32 to -21.45 | 1,165 | -23.86 | -26.44 to -21.29 |
| S1 | Solar Decide Flash (D10) - DeepSeek V4.1 Flash Low (G06) | 1,273 | -22.62 | -25.06 to -20.27 | 1,165 | -22.66 | -25.24 to -20.17 |
| S1 | Solar Decide Flash (D10) - DeepSeek V4.1 Flash High (G07) | 1,273 | -21.68 | -24.12 to -19.25 | 1,165 | -21.80 | -24.38 to -19.23 |
| S1 | Solar Decide Flash (D10) - Qwen3.8 Max Low (G08) | 1,273 | -16.73 | -19.25 to -14.30 | 1,165 | -17.00 | -19.66 to -14.33 |
| S1 | Solar Decide Flash (D10) - Qwen3.8- | 1,273 | -21.21 | -23.64 to -18.85 | 1,165 | -21.46 | -23.95 to -18.97 |

| Family | Model A - Model B | Complete-valid n | Complete-valid effect, pp | Complete-valid 95% CI, pp | Visual-filtered n | Visual-filtered effect, pp | Visual-filtered 95% CI, pp |
|---|---|---|---|---|---|---|---|
| | 27B Low (G09) | | | | | | |
| S1 | Solar Decide Flash (D10) - Qwen3.8-27B Xhigh (G10) | 1,272 | -22.56 | -25.00 to -20.13 | 1,165 | -22.83 | -25.41 to -20.34 |
| S1 | Solar Decide Flash (D10) - Mistral Medium 3.5 None (G11) | 1,273 | -5.42 | -7.70 to -3.14 | 1,165 | -5.32 | -7.73 to -2.92 |
| S1 | Solar Decide Flash (D10) - GLM-5 Native (G12) | 1,262 | -22.50 | -24.92 to -20.09 | 1,165 | -22.32 | -24.89 to -19.74 |
| S1 | Solar Decide Flash (D10) - Kimi K2.5 Native (G13) | 1,273 | -21.60 | -24.04 to -19.17 | 1,165 | -21.80 | -24.38 to -19.31 |
| S1 | Solar Decide Flash (D10) - Llama 4 | 1,273 | -10.92 | -13.35 to -8.48 | 1,165 | -10.99 | -13.56 to -8.33 |

| Family | Model A - Model B | Complete -valid n | Complete -valid effect, pp | Complete -valid 95% CI, pp | Visual-filtered n | Visual-filtered effect, pp | Visual-filtered 95% CI, pp |
|---|---|---|---|---|---|---|---|
| | Maverick Standard (G14) | | | | | | |
| S1 | Solar Decide Flash (D10) - Kimi K2.5 Instant (G15) | 1,273 | -12.10 | -14.53 to -9.66 | 1,165 | -12.53 | -15.11 to -9.96 |
| S1 | Clef Omni (D11) - GPT-6.1 Sol Low (G01) | 1,273 | -17.44 | -19.64 to -15.24 | 1,165 | -16.91 | -19.23 to -14.68 |
| S1 | Clef Omni (D11) - GPT-6 Luna Low (G02) | 1,273 | -14.77 | -17.05 to -12.49 | 1,165 | -14.59 | -16.91 to -12.36 |
| S1 | Clef Omni (D11) - GPT-6 Luna High (G03) | 1,273 | -15.79 | -18.15 to -13.43 | 1,165 | -15.54 | -17.94 to -13.22 |
| S1 | Clef Omni (D11) - Claude Sonnet 5.5 Low (G04) | 1,272 | -15.80 | -18.03 to -13.55 | 1,165 | -15.28 | -17.60 to -13.05 |
| S1 | Clef Omni (D11) - Gemini | 1,273 | -17.36 | -19.56 to -15.16 | 1,165 | -16.82 | -19.06 to -14.59 |

| Family | Model A - Model B | Complete-valid n | Complete-valid effect, pp | Complete-valid 95% CI, pp | Visual-filtered n | Visual-filtered effect, pp | Visual-filtered 95% CI, pp |
|---|---|---|---|---|---|---|---|
| | 3.8 Flash Low (G05) | | | | | | |
| S1 | Clef Omni (D11) - DeepSeek V4.1 Flash Low (G06) | 1,273 | -16.10 | -18.30 to -13.83 | 1,165 | -15.62 | -18.03 to -13.30 |
| S1 | Clef Omni (D11) - DeepSeek V4.1 Flash High (G07) | 1,273 | -15.16 | -17.44 to -12.88 | 1,165 | -14.76 | -17.17 to -12.45 |
| S1 | Clef Omni (D11) - Qwen3.8 Max Low (G08) | 1,273 | -10.21 | -12.41 to -8.01 | 1,165 | -9.96 | -12.27 to -7.73 |
| S1 | Clef Omni (D11) - Qwen3.8-27B Low (G09) | 1,273 | -14.69 | -16.89 to -12.49 | 1,165 | -14.42 | -16.65 to -12.19 |
| S1 | Clef Omni (D11) - Qwen3.8-27B Xhigh (G10) | 1,272 | -16.04 | -18.24 to -13.83 | 1,165 | -15.79 | -18.11 to -13.56 |
| S1 | Clef Omni (D11) - Mistral Medium 3.5 None (G11) | 1,273 | 1.10 | -1.10 to 3.30 | 1,165 | 1.72 | -0.60 to 3.95 |

| Family | Model A - Model B | Complete -valid n | Complete -valid effect, pp | Complete -valid 95% CI, pp | Visual-filtered n | Visual-filtered effect, pp | Visual-filtered 95% CI, pp |
|---|---|---|---|---|---|---|---|
| S1 | Clef Omni (D11) - GLM-5 Native (G12) | 1,262 | -16.01 | -18.20 to -13.81 | 1,165 | -15.28 | -17.60 to -13.05 |
| S1 | Clef Omni (D11) - Kimi K2.5 Native (G13) | 1,273 | -15.08 | -17.28 to -12.88 | 1,165 | -14.76 | -17.08 to -12.53 |
| S1 | Clef Omni (D11) - Llama 4 Maverick Standard (G14) | 1,273 | -4.40 | -6.68 to -2.12 | 1,165 | -3.95 | -6.35 to -1.55 |
| S1 | Clef Omni (D11) - Kimi K2.5 Instant (G15) | 1,273 | -5.58 | -7.70 to -3.46 | 1,165 | -5.49 | -7.73 to -3.26 |
| S1 | Drex v1.5 (D12) - GPT-6.1 Sol Low (G01) | 1,273 | -23.33 | -25.84 to -20.82 | 1,165 | -22.75 | -25.32 to -20.17 |
| S1 | Drex v1.5 (D12) - GPT-6 Luna Low (G02) | 1,273 | -20.66 | -23.25 to -18.15 | 1,165 | -20.43 | -23.09 to -17.85 |
| S1 | Drex v1.5 (D12) - GPT-6 | 1,273 | -21.68 | -24.19 to -19.17 | 1,165 | -21.37 | -23.95 to -18.80 |

| Family | Model A - Model B | Complete-valid n | Complete-valid effect, pp | Complete-valid 95% CI, pp | Visual-filtered n | Visual-filtered effect, pp | Visual-filtered 95% CI, pp |
|---|---|---|---|---|---|---|---|
| | Luna High (G03) | | | | | | |
| S1 | Drex v1.5 (D12) - Claude Sonnet 5.5 Low (G04) | 1,272 | -21.62 | -24.14 to -19.17 | 1,165 | -21.12 | -23.69 to -18.63 |
| S1 | Drex v1.5 (D12) - Gemini 3.8 Flash Low (G05) | 1,273 | -23.25 | -25.69 to -20.82 | 1,165 | -22.66 | -25.24 to -20.09 |
| S1 | Drex v1.5 (D12) - DeepSeek V4.1 Flash Low (G06) | 1,273 | -22.00 | -24.51 to -19.48 | 1,165 | -21.46 | -24.12 to -18.88 |
| S1 | Drex v1.5 (D12) - DeepSeek V4.1 Flash High (G07) | 1,273 | -21.05 | -23.57 to -18.46 | 1,165 | -20.60 | -23.26 to -17.94 |
| S1 | Drex v1.5 (D12) - Qwen3.8 Max Low (G08) | 1,273 | -16.10 | -18.54 to -13.67 | 1,165 | -15.79 | -18.28 to -13.30 |
| S1 | Drex v1.5 (D12) - Qwen3.8-27B Low (G09) | 1,273 | -20.58 | -23.02 to -18.15 | 1,165 | -20.26 | -22.83 to -17.77 |

| Family | Model A - Model B | Complete-valid n | Complete-valid effect, pp | Complete-valid 95% CI, pp | Visual-filtered n | Visual-filtered effect, pp | Visual-filtered 95% CI, pp |
|---|---|---|---|---|---|---|---|
| S1 | Drex v1.5 (D12) - Qwen3.8-27B Xhigh (G10) | 1,272 | -21.93 | -24.43 to -19.42 | 1,165 | -21.63 | -24.21 to -19.06 |
| S1 | Drex v1.5 (D12) - Mistral Medium 3.5 None (G11) | 1,273 | -4.79 | -7.31 to -2.20 | 1,165 | -4.12 | -6.78 to -1.46 |
| S1 | Drex v1.5 (D12) - GLM-5 Native (G12) | 1,262 | -21.95 | -24.42 to -19.48 | 1,165 | -21.12 | -23.69 to -18.54 |
| S1 | Drex v1.5 (D12) - Kimi K2.5 Native (G13) | 1,273 | -20.97 | -23.49 to -18.54 | 1,165 | -20.60 | -23.18 to -18.03 |
| S1 | Drex v1.5 (D12) - Llama 4 Maverick Standard (G14) | 1,273 | -10.29 | -12.80 to -7.78 | 1,165 | -9.79 | -12.45 to -7.12 |
| S1 | Drex v1.5 (D12) - Kimi K2.5 Instant (G15) | 1,273 | -11.47 | -13.98 to -8.96 | 1,165 | -11.33 | -13.91 to -8.76 |
| S1 | Microsoft-Decision-1 (D13) - | 1,273 | -17.28 | -19.56 to -15.00 | 1,165 | -16.48 | -18.80 to -14.25 |

| Family | Model A - Model B | Complete -valid n | Complete -valid effect, pp | Complete -valid 95% CI, pp | Visual-filtered n | Visual-filtered effect, pp | Visual-filtered 95% CI, pp |
|---|---|---|---|---|---|---|---|
| | GPT-6.1 Sol Low (G01) | | | | | | |
| S1 | Microsoft-Decision-1 (D13) - GPT-6 Luna Low (G02) | 1,273 | -14.61 | -16.97 to -12.33 | 1,165 | -14.16 | -16.48 to -11.85 |
| S1 | Microsoft-Decision-1 (D13) - GPT-6 Luna High (G03) | 1,273 | -15.63 | -17.99 to -13.35 | 1,165 | -15.11 | -17.51 to -12.79 |
| S1 | Microsoft-Decision-1 (D13) - Claude Sonnet 5.5 Low (G04) | 1,272 | -15.64 | -17.92 to -13.44 | 1,165 | -14.85 | -17.17 to -12.62 |
| S1 | Microsoft-Decision-1 (D13) - Gemini 3.8 Flash Low (G05) | 1,273 | -17.20 | -19.48 to -14.93 | 1,165 | -16.39 | -18.80 to -14.08 |
| S1 | Microsoft-Decision-1 (D13) - DeepSeek V4.1 Flash Low (G06) | 1,273 | -15.95 | -18.22 to -13.75 | 1,165 | -15.19 | -17.60 to -12.96 |

| Family | Model A - Model B | Complete-valid n | Complete-valid effect, pp | Complete-valid 95% CI, pp | Visual-filtered n | Visual-filtered effect, pp | Visual-filtered 95% CI, pp |
|---|---|---|---|---|---|---|---|
| S1 | Microsoft-Decision-1 (D13) - DeepSeek V4.1 Flash High (G07) | 1,273 | -15.00 | -17.28 to -12.73 | 1,165 | -14.33 | -16.74 to -12.02 |
| S1 | Microsoft-Decision-1 (D13) - Qwen3.8 Max Low (G08) | 1,273 | -10.05 | -12.33 to -7.78 | 1,165 | -9.53 | -11.93 to -7.21 |
| S1 | Microsoft-Decision-1 (D13) - Qwen3.8-27B Low (G09) | 1,273 | -14.53 | -16.81 to -12.33 | 1,165 | -13.99 | -16.22 to -11.76 |
| S1 | Microsoft-Decision-1 (D13) - Qwen3.8-27B Xhigh (G10) | 1,272 | -15.88 | -18.10 to -13.68 | 1,165 | -15.36 | -17.68 to -13.13 |
| S1 | Microsoft-Decision-1 (D13) - Mistral Medium 3.5 None (G11) | 1,273 | 1.26 | -1.18 to 3.61 | 1,165 | 2.15 | -0.34 to 4.64 |
| S1 | Microsoft-Decision-1 (D13) - GLM-5 | 1,262 | -15.85 | -18.13 to -13.61 | 1,165 | -14.85 | -17.25 to -12.53 |

| Family | Model A - Model B | Complete-valid n | Complete-valid effect, pp | Complete-valid 95% CI, pp | Visual-filtered n | Visual-filtered effect, pp | Visual-filtered 95% CI, pp |
|---|---|---|---|---|---|---|---|
| | Native (G12) | | | | | | |
| S1 | Microsoft-Decision-1 (D13) - Kimi K2.5 Native (G13) | 1,273 | -14.93 | -17.20 to -12.65 | 1,165 | -14.33 | -16.74 to -12.02 |
| S1 | Microsoft-Decision-1 (D13) - Llama 4 Maverick Standard (G14) | 1,273 | -4.24 | -6.52 to -1.96 | 1,165 | -3.52 | -5.84 to -1.12 |
| S1 | Microsoft-Decision-1 (D13) - Kimi K2.5 Instant (G15) | 1,273 | -5.42 | -7.70 to -3.14 | 1,165 | -5.06 | -7.47 to -2.75 |
| S1 | JEV 1.13 (D01) - GPT-6 Luna Decisions (D02) | 1,271 | -0.87 | -2.59 to 0.87 | 1,165 | -0.26 | -2.06 to 1.55 |
| S1 | JEV 1.13 (D01) - Kev 4B (D03) | 1,273 | 20.58 | 17.99 to 23.17 | 1,165 | 20.94 | 18.28 to 23.69 |
| S1 | JEV 1.13 (D01) - Solar | 1,273 | 5.66 | 3.53 to 7.78 | 1,165 | 6.27 | 4.03 to 8.50 |

| Family | Model A - Model B | Complete-valid n | Complete-valid effect, pp | Complete-valid 95% CI, pp | Visual-filtered n | Visual-filtered effect, pp | Visual-filtered 95% CI, pp |
|---|---|---|---|---|---|---|---|
| | Decide (D04) | | | | | | |
| S1 | JEV 1.13 (D01) - Mercury Decide (D05) | 1,273 | 3.06 | 1.26 to 4.95 | 1,165 | 3.43 | 1.46 to 5.41 |
| S1 | JEV 1.13 (D01) - Clef (D06) | 1,273 | 3.53 | 1.65 to 5.42 | 1,165 | 3.86 | 1.89 to 5.75 |
| S1 | JEV 1.13 (D01) - Clef Flash (D07) | 1,273 | 14.61 | 12.25 to 17.05 | 1,165 | 14.33 | 11.85 to 16.82 |
| S1 | JEV 1.13 (D01) - Liquid d1 (D08) | 1,273 | -0.08 | -1.73 to 1.57 | 1,165 | 0.43 | -1.29 to 2.15 |
| S1 | JEV 1.13 (D01) - Decider V1.1 27B (D09) | 1,273 | 0.47 | -1.26 to 2.20 | 1,165 | 0.69 | -1.03 to 2.40 |
| S1 | JEV 1.13 (D01) - Solar Decide Flash (D10) | 1,273 | 15.40 | 12.88 to 17.91 | 1,165 | 16.14 | 13.48 to 18.71 |
| S1 | JEV 1.13 (D01) - Clef Omni (D11) | 1,273 | 8.88 | 6.68 to 11.15 | 1,165 | 9.10 | 6.87 to 11.33 |

| Family | Model A - Model B | Complete-valid n | Complete-valid effect, pp | Complete-valid 95% CI, pp | Visual-filtered n | Visual-filtered effect, pp | Visual-filtered 95% CI, pp |
|---|---|---|---|---|---|---|---|
| S1 | JEV 1.13 (D01) - Drex v1.5 (D12) | 1,273 | 14.77 | 12.25 to 17.36 | 1,165 | 14.94 | 12.27 to 17.51 |
| S1 | JEV 1.13 (D01) - Microsoft-Decision-1 (D13) | 1,273 | 8.72 | 6.52 to 10.92 | 1,165 | 8.67 | 6.44 to 10.99 |
| S1 | GPT-6 Luna Decisions (D02) - Kev 4B (D03) | 1,271 | 21.48 | 18.79 to 24.14 | 1,165 | 21.20 | 18.45 to 24.03 |
| S1 | GPT-6 Luna Decisions (D02) - Solar Decide (D04) | 1,271 | 6.61 | 4.50 to 8.73 | 1,165 | 6.52 | 4.38 to 8.67 |
| S1 | GPT-6 Luna Decisions (D02) - Mercury Decide (D05) | 1,271 | 3.93 | 2.05 to 5.89 | 1,165 | 3.69 | 1.72 to 5.67 |
| S1 | GPT-6 Luna Decisions (D02) - Clef (D06) | 1,271 | 4.48 | 2.68 to 6.30 | 1,165 | 4.12 | 2.23 to 6.09 |

| Family | Model A - Model B | Complete-valid n | Complete-valid effect, pp | Complete-valid 95% CI, pp | Visual-filtered n | Visual-filtered effect, pp | Visual-filtered 95% CI, pp |
|---|---|---|---|---|---|---|---|
| S1 | GPT-6 Luna Decisions (D02) - Clef Flash (D07) | 1,271 | 15.58 | 13.15 to 18.08 | 1,165 | 14.59 | 12.10 to 17.08 |
| S1 | GPT-6 Luna Decisions (D02) - Liquid d1 (D08) | 1,271 | 0.87 | -0.94 to 2.60 | 1,165 | 0.69 | -1.12 to 2.49 |
| S1 | GPT-6 Luna Decisions (D02) - Decider V1.1 27B (D09) | 1,271 | 1.42 | -0.47 to 3.23 | 1,165 | 0.94 | -0.94 to 2.92 |
| S1 | GPT-6 Luna Decisions (D02) - Solar Decide Flash (D10) | 1,271 | 16.37 | 13.92 to 18.80 | 1,165 | 16.39 | 13.82 to 18.97 |
| S1 | GPT-6 Luna Decisions (D02) - Clef Omni (D11) | 1,271 | 9.83 | 7.63 to 12.06 | 1,165 | 9.36 | 7.04 to 11.67 |
| S1 | GPT-6 Luna Decisions | 1,271 | 15.66 | 13.21 to 18.16 | 1,165 | 15.19 | 12.62 to 17.85 |

| Family | Model A - Model B | Complete-valid n | Complete-valid effect, pp | Complete-valid 95% CI, pp | Visual-filtered n | Visual-filtered effect, pp | Visual-filtered 95% CI, pp |
|---|---|---|---|---|---|---|---|
| | (D02) - Drex v1.5 (D12) | | | | | | |
| S1 | GPT-6 Luna Decisions (D02) - Microsoft-Decision-1 (D13) | 1,271 | 9.68 | 7.40 to 12.02 | 1,165 | 8.93 | 6.52 to 11.42 |
| S1 | Kev 4B (D03) - Solar Decide (D04) | 1,273 | -14.93 | -17.60 to -12.33 | 1,165 | -14.68 | -17.51 to -11.93 |
| S1 | Kev 4B (D03) - Mercury Decide (D05) | 1,273 | -17.52 | -20.19 to -14.93 | 1,165 | -17.51 | -20.26 to -14.85 |
| S1 | Kev 4B (D03) - Clef (D06) | 1,273 | -17.05 | -19.64 to -14.45 | 1,165 | -17.08 | -19.83 to -14.42 |
| S1 | Kev 4B (D03) - Clef Flash (D07) | 1,273 | -5.97 | -8.48 to -3.38 | 1,165 | -6.61 | -9.27 to -3.95 |
| S1 | Kev 4B (D03) - Liquid d1 (D08) | 1,273 | -20.66 | -23.25 to -18.07 | 1,165 | -20.52 | -23.26 to -17.85 |
| S1 | Kev 4B (D03) - Decider | 1,273 | -20.11 | -22.78 to -17.44 | 1,165 | -20.26 | -23.00 to -17.51 |

| Family | Model A - Model B | Complete-valid n | Complete-valid effect, pp | Complete-valid 95% CI, pp | Visual-filtered n | Visual-filtered effect, pp | Visual-filtered 95% CI, pp |
|---|---|---|---|---|---|---|---|
| | V1.1 27B (D09) | | | | | | |
| S1 | Kev 4B (D03) - Solar Decide Flash (D10) | 1,273 | -5.18 | -7.86 to -2.51 | 1,165 | -4.81 | -7.55 to -2.06 |
| S1 | Kev 4B (D03) - Clef Omni (D11) | 1,273 | -11.70 | -14.22 to -9.19 | 1,165 | -11.85 | -14.42 to -9.18 |
| S1 | Kev 4B (D03) - Drex v1.5 (D12) | 1,273 | -5.81 | -8.41 to -3.22 | 1,165 | -6.01 | -8.76 to -3.26 |
| S1 | Kev 4B (D03) - Microsoft-Decision-1 (D13) | 1,273 | -11.86 | -14.45 to -9.27 | 1,165 | -12.27 | -14.94 to -9.61 |
| S1 | Solar Decide (D04) - Mercury Decide (D05) | 1,273 | -2.59 | -4.71 to -0.39 | 1,165 | -2.83 | -5.15 to -0.60 |
| S1 | Solar Decide (D04) - Clef (D06) | 1,273 | -2.12 | -4.24 to 0.16 | 1,165 | -2.40 | -4.81 to -0.09 |
| S1 | Solar Decide (D04) - | 1,273 | 8.96 | 6.52 to 11.47 | 1,165 | 8.07 | 5.49 to 10.73 |

| Family | Model A - Model B | Complete-valid n | Complete-valid effect, pp | Complete-valid 95% CI, pp | Visual-filtered n | Visual-filtered effect, pp | Visual-filtered 95% CI, pp |
|---|---|---|---|---|---|---|---|
| | Clef Flash (D07) | | | | | | |
| S1 | Solar Decide (D04) - Liquid d1 (D08) | 1,273 | -5.73 | -7.93 to -3.61 | 1,165 | -5.84 | -8.07 to -3.52 |
| S1 | Solar Decide (D04) - Decider V1.1 27B (D09) | 1,273 | -5.18 | -7.38 to -2.99 | 1,165 | -5.58 | -7.90 to -3.26 |
| S1 | Solar Decide (D04) - Solar Decide Flash (D10) | 1,273 | 9.74 | 7.93 to 11.55 | 1,165 | 9.87 | 7.98 to 11.85 |
| S1 | Solar Decide (D04) - Clef Omni (D11) | 1,273 | 3.22 | 0.86 to 5.50 | 1,165 | 2.83 | 0.34 to 5.24 |
| S1 | Solar Decide (D04) - Drex v1.5 (D12) | 1,273 | 9.11 | 6.44 to 11.79 | 1,165 | 8.67 | 5.84 to 11.50 |
| S1 | Solar Decide (D04) - Microsoft- | 1,273 | 3.06 | 0.63 to 5.50 | 1,165 | 2.40 | -0.09 to 4.98 |

| Family | Model A - Model B | Complete-valid n | Complete-valid effect, pp | Complete-valid 95% CI, pp | Visual-filtered n | Visual-filtered effect, pp | Visual-filtered 95% CI, pp |
|---|---|---|---|---|---|---|---|
| | Decision-1 (D13) | | | | | | |
| S1 | Mercury Decide (D05) - Clef (D06) | 1,273 | 0.47 | -1.10 to 2.04 | 1,165 | 0.43 | -1.20 to 2.06 |
| S1 | Mercury Decide (D05) - Clef Flash (D07) | 1,273 | 11.55 | 9.19 to 13.98 | 1,165 | 10.90 | 8.33 to 13.48 |
| S1 | Mercury Decide (D05) - Liquid d1 (D08) | 1,273 | -3.14 | -5.03 to -1.34 | 1,165 | -3.00 | -4.98 to -1.12 |
| S1 | Mercury Decide (D05) - Decider V1.1 27B (D09) | 1,273 | -2.59 | -4.32 to -0.94 | 1,165 | -2.75 | -4.55 to -1.03 |
| S1 | Mercury Decide (D05) - Solar Decide Flash (D10) | 1,273 | 12.33 | 9.90 to 14.69 | 1,165 | 12.70 | 10.13 to 15.28 |
| S1 | Mercury Decide (D05) - Clef Omni (D11) | 1,273 | 5.81 | 3.53 to 8.01 | 1,165 | 5.67 | 3.35 to 7.98 |

| Family | Model A - Model B | Complete-valid n | Complete-valid effect, pp | Complete-valid 95% CI, pp | Visual-filtered n | Visual-filtered effect, pp | Visual-filtered 95% CI, pp |
|---|---|---|---|---|---|---|---|
| S1 | Mercury Decide (D05) - Drex v1.5 (D12) | 1,273 | 11.70 | 9.11 to 14.30 | 1,165 | 11.50 | 8.93 to 14.08 |
| S1 | Mercury Decide (D05) - Microsoft-Decision-1 (D13) | 1,273 | 5.66 | 3.38 to 7.93 | 1,165 | 5.24 | 2.92 to 7.64 |
| S1 | Clef (D06) - Clef Flash (D07) | 1,273 | 11.08 | 8.80 to 13.43 | 1,165 | 10.47 | 8.07 to 12.96 |
| S1 | Clef (D06) - Liquid d1 (D08) | 1,273 | -3.61 | -5.50 to -1.81 | 1,165 | -3.43 | -5.32 to -1.63 |
| S1 | Clef (D06) - Decider V1.1 27B (D09) | 1,273 | -3.06 | -4.87 to -1.34 | 1,165 | -3.18 | -4.98 to -1.29 |
| S1 | Clef (D06) - Solar Decide Flash (D10) | 1,273 | 11.86 | 9.51 to 14.30 | 1,165 | 12.27 | 9.79 to 14.85 |
| S1 | Clef (D06) - Clef Omni (D11) | 1,273 | 5.34 | 3.14 to 7.46 | 1,165 | 5.24 | 2.92 to 7.55 |
| S1 | Clef (D06) - Drex v1.5 (D12) | 1,273 | 11.23 | 8.72 to 13.75 | 1,165 | 11.07 | 8.50 to 13.65 |

| Family | Model A - Model B | Complete-valid n | Complete-valid effect, pp | Complete-valid 95% CI, pp | Visual-filtered n | Visual-filtered effect, pp | Visual-filtered 95% CI, pp |
|---|---|---|---|---|---|---|---|
| S1 | Clef (D06) - Microsoft-Decision-1 (D13) | 1,273 | 5.18 | 2.91 to 7.46 | 1,165 | 4.81 | 2.49 to 7.21 |
| S1 | Clef Flash (D07) - Liquid d1 (D08) | 1,273 | -14.69 | -17.05 to -12.41 | 1,165 | -13.91 | -16.39 to -11.50 |
| S1 | Clef Flash (D07) - Decider V1.1 27B (D09) | 1,273 | -14.14 | -16.58 to -11.78 | 1,165 | -13.65 | -16.22 to -11.16 |
| S1 | Clef Flash (D07) - Solar Decide Flash (D10) | 1,273 | 0.79 | -1.73 to 3.30 | 1,165 | 1.80 | -0.86 to 4.46 |
| S1 | Clef Flash (D07) - Clef Omni (D11) | 1,273 | -5.73 | -8.17 to -3.30 | 1,165 | -5.24 | -7.81 to -2.75 |
| S1 | Clef Flash (D07) - Drex v1.5 (D12) | 1,273 | 0.16 | -2.28 to 2.51 | 1,165 | 0.60 | -1.89 to 3.09 |
| S1 | Clef Flash (D07) - Microsoft-Decision-1 (D13) | 1,273 | -5.89 | -8.17 to -3.53 | 1,165 | -5.67 | -8.07 to -3.26 |

| Family | Model A - Model B | Complete-valid n | Complete-valid effect, pp | Complete-valid 95% CI, pp | Visual-filtered n | Visual-filtered effect, pp | Visual-filtered 95% CI, pp |
|---|---|---|---|---|---|---|---|
| S1 | Liquid d1 (D08) - Decider V1.1 27B (D09) | 1,273 | 0.55 | -1.34 to 2.36 | 1,165 | 0.26 | -1.63 to 2.23 |
| S1 | Liquid d1 (D08) - Solar Decide Flash (D10) | 1,273 | 15.48 | 13.04 to 17.91 | 1,165 | 15.71 | 13.13 to 18.28 |
| S1 | Liquid d1 (D08) - Clef Omni (D11) | 1,273 | 8.96 | 6.83 to 11.08 | 1,165 | 8.67 | 6.44 to 10.90 |
| S1 | Liquid d1 (D08) - Drex v1.5 (D12) | 1,273 | 14.85 | 12.41 to 17.28 | 1,165 | 14.51 | 12.02 to 17.00 |
| S1 | Liquid d1 (D08) - Microsoft-Decision-1 (D13) | 1,273 | 8.80 | 6.52 to 11.00 | 1,165 | 8.24 | 6.01 to 10.56 |
| S1 | Decider V1.1 27B (D09) - Solar Decide Flash (D10) | 1,273 | 14.93 | 12.41 to 17.44 | 1,165 | 15.45 | 12.79 to 18.11 |
| S1 | Decider V1.1 27B (D09) - | 1,273 | 8.41 | 6.21 to 10.68 | 1,165 | 8.41 | 6.09 to 10.73 |

| Family | Model A - Model B | Complete -valid n | Complete -valid effect, pp | Complete -valid 95% CI, pp | Visual-filtered n | Visual-filtered effect, pp | Visual-filtered 95% CI, pp |
|---|---|---|---|---|---|---|---|
| | Clef Omni (D11) | | | | | | |
| S1 | Decider V1.1 27B (D09) - Drex v1.5 (D12) | 1,273 | 14.30 | 11.70 to 16.81 | 1,165 | 14.25 | 11.67 to 16.91 |
| S1 | Decider V1.1 27B (D09) - Microsoft-Decision-1 (D13) | 1,273 | 8.25 | 6.05 to 10.53 | 1,165 | 7.98 | 5.67 to 10.30 |
| S1 | Solar Decide Flash (D10) - Clef Omni (D11) | 1,273 | -6.52 | -8.96 to -4.08 | 1,165 | -7.04 | -9.61 to -4.46 |
| S1 | Solar Decide Flash (D10) - Drex v1.5 (D12) | 1,273 | -0.63 | -3.30 to 2.04 | 1,165 | -1.20 | -3.95 to 1.63 |
| S1 | Solar Decide Flash (D10) - Microsoft-Decision-1 (D13) | 1,273 | -6.68 | -9.19 to -4.16 | 1,165 | -7.47 | -10.04 to -4.89 |
| S1 | Clef Omni (D11) - | 1,273 | 5.89 | 3.30 to 8.48 | 1,165 | 5.84 | 3.09 to 8.50 |

| Family | Model A - Model B | Complete-valid n | Complete-valid effect, pp | Complete-valid 95% CI, pp | Visual-filtered n | Visual-filtered effect, pp | Visual-filtered 95% CI, pp |
|---|---|---|---|---|---|---|---|
| | Drex v1.5 (D12) | | | | | | |
| S1 | Clef Omni (D11) - Microsoft-Decision-1 (D13) | 1,273 | -0.16 | -2.52 to 2.28 | 1,165 | -0.43 | -2.92 to 2.06 |
| S1 | Drex v1.5 (D12) - Microsoft-Decision-1 (D13) | 1,273 | -6.05 | -8.48 to -3.69 | 1,165 | -6.27 | -8.76 to -3.78 |
| S2 | GPT-6 Luna High (G03) - GPT-6 Luna Low (G02) | 1,273 | 1.02 | 0.16 to 1.89 | 1,165 | 0.94 | 0.09 to 1.80 |
| S2 | DeepSeek V4.1 Flash High (G07) - DeepSeek V4.1 Flash Low (G06) | 1,273 | -0.94 | -1.73 to -0.24 | 1,165 | -0.86 | -1.63 to -0.09 |
| S2 | Qwen3.8-27B Xhigh (G10) - Qwen3.8-27B Low (G09) | 1,272 | 1.34 | 0.16 to 2.52 | 1,165 | 1.37 | 0.17 to 2.66 |